\documentclass[pdflatex,sn-nature]{sn-jnl}
\usepackage{graphicx}%
\usepackage{subcaption}[]
\usepackage{multirow}%
\usepackage{amsmath,amssymb,amsfonts}%
\usepackage{amsthm}%
\usepackage{mathrsfs}%
\usepackage[title]{appendix}%
\usepackage{xcolor}%
\usepackage{textcomp}%
\usepackage{manyfoot}%
\usepackage{booktabs}%
\usepackage{algorithm}%
\usepackage{algorithmicx}%
\usepackage{algpseudocode}%
\usepackage{listings}%
\usepackage{tabularx}

\theoremstyle{thmstyleone}%
\theoremstyle{thmstyletwo}%

\theoremstyle{thmstylethree}%

\begin{document}

\title[Article Title]{Realistic threat perception drives intergroup conflict: A causal, dynamic analysis using generative-agent simulations}

\author*[1,2]{\fnm{Suhaib} \sur{Abdurahman}}\email{sabdurah@usc.edu}

\author[1,2]{\fnm{Farzan} \sur{Karimi-Malekabadi}}\email{karimima@usc.edu}

\author[3]{\fnm{Chenxiao} \sur{Yu}}\email{cyu96374@usc.edu}

\author[4]{\fnm{Nour S.} \sur{Kteily}}\email{n-kteily@kellogg.northwestern.edu}

\author[1,2,3]{\fnm{Morteza} \sur{Dehghani}}\email{mdehghan@usc.edu}

\affil*[1]{\orgdiv{Department of Psychology}, \orgname{University of Southern California}}

\affil[2]{\orgdiv{Center for Computational Language Sciences}, \orgname{University of Southern California}}

\affil[3]{\orgdiv{Department of Computer Science}, \orgname{University of Southern California}}

\affil[4]{\orgdiv{Kellogg School of Management}, \orgname{Northwestern University}}


\abstract{
Human conflict is often attributed to threats against material conditions and symbolic values, yet it remains unclear how they interact and which dominates. Progress is limited by weak causal control, ethical constraints, and scarce temporal data. We address these barriers using simulations of large language model (LLM)-driven agents in virtual societies, independently varying realistic and symbolic threat while tracking actions, language, and attitudes. Representational analyses show that the underlying LLM encodes realistic threat, symbolic threat, and hostility as distinct internal states, that our manipulations map onto them, and that steering these states causally shifts behavior. Our simulations provide a causal account of threat-driven conflict over time: realistic threat directly increases hostility, whereas symbolic threat effects are weaker, fully mediated by ingroup bias, and increase hostility only when realistic threat is absent. Non-hostile intergroup contact buffers escalation, and structural asymmetries concentrate hostility among majority groups.
}




\maketitle

\section{Introduction}\label{sec:intro}

Human history is marked by groups fighting over material concerns such as resources or physical security, and over symbolic concerns such as identity and sacred values. In psychology, this distinction is formalized by Integrated Threat Theory \citep[ITT;][]{stephan2013intergroup} as a contrast between \emph{realistic threats} and \emph{symbolic threats}. Political science frames the same divide in realist versus constructivist theories of power and identity and in the ``greed versus grievance'' debate in civil war studies \citep{waltz1979theory,mearsheimer2001tragedy,wendt1999social,fearon2003ethnicity,collier2004greed}. Sociology and history similarly contrast material opportunity structures with symbolic boundaries and cultural frames \citep{tilly2003politics,lamont2002study}, and work in neuroscience suggests partially distinct processing for threats to sacred values relative to material threats \citep{pretus2018neural}. Yet, despite decades of research across these disciplines, it remains unclear how realistic and symbolic threats drive group conflict, whether one tends to dominate, whether they amplify or subsume one another. Clarifying this relationship would help reconcile competing theories of conflict and improve our ability to understand, predict, and prevent real-world escalation. 
Here, we turn to generative-agent simulations in which autonomous agents powered by large language models (LLMs) inhabit a shared, spatially structured environment, interact and converse, form memories, and pursue open-ended goals \citep{park2023generative,gao2024large}. Unlike classical agent-based models (ABM) that rely on hand-coded decision rules \citep{bruch2015agent,axelrod1997complexity,lustick2002psi,cederman2003modeling}, generative agents plan, act, and reflect via the underlying LLM, producing complex social dynamics without pre-specified decision rules \citep{park2023generative,park2024generative}. We adopt this approach to generate complementary insights into how realistic and symbolic threats drive group conflict that would otherwise be difficult to obtain given conceptual and methodological barriers in the field. 

For example, evidence is fragmented across levels of analysis making it difficult to trace how individual-level threat perceptions translate into group conflict: macro-level work explains conflict in terms of institutions and structural conditions, while micro-level work explains cooperation and support for violence in terms of local experiences, identities, and values \citep{fearon2003ethnicity,collier2004greed,cederman2010why,kalyvas2006logic,toft2007getting,lamont2002study,ginges2007sacred,bauer2014war,voors2012violent}. Causal inference is also limited because studies of real-world conflict dynamics are predominantly observational or quasi-experimental, and real-world manipulations of realistic threats, symbolic threats, and structural features are typically impossible or deeply unethical (e.g., inducing material insecurity, outlawing traditions, causing genuine harm). Laboratory and survey experiments offer stronger causal control but usually rely on short-term, abstract experimental contexts \citep{craig2014precipice,kachanoff2019chains,paolini2010negative,arnadottir2024positive}, such as threat primes that produce attitudinal shifts, and often lack the shared environments, extended interactions, and consequential behaviors that characterize group conflicts. Material and symbolic dimensions are also often deeply entangled: economic insecurity can be politicized through identity-based narratives, while cultural affronts can generate material retaliation \citep{gidron2017politics,mudde2018studying,hopkins2010politicized,inglehart2016trump,cramer2022politics,mutz2018status}, making it difficult to isolate their effects in natural settings. Finally, most lab studies capture snapshots of conflict rather than extended trajectories of interaction, making it difficult to observe how threat and conflict evolve over time. Together, these constraints have so far prevented an integrated, causal, and dynamic account linking individual threat perception to group conflict within a single system.

Generative-agent simulations help in several ways. They embed individual-level threat perceptions and group-level conflict dynamics in the same modeled social system, helping to bridge the gap between micro-level mechanisms and macro-level outcomes. Because the researcher specifies both the environment and the agents, simulations allow systematic manipulation of structural features (e.g., segregation, group size) alongside agent-level attributes (e.g., perceived threat, group membership), enabling causal experiments that would be infeasible or unethical in real populations, such as exposing agents to symbolic and realistic threats and allowing severe hostility and harmful outcomes to emerge. Control over the simulation environment also allows researchers to disentangle material and symbolic dimensions by orthogonalizing manipulations and ensuring that only the target threat(s) are present. Additionally, simulations support longitudinal analysis by enabling extended time horizons and comprehensive logging of the interaction process, including plans, actions, conversations, reflections, and internal state probes (e.g., via psychological scales), yielding rich, high-resolution data on how threat and conflict evolve over time. Finally, because the agents are implemented by an underlying LLM, the framework also opens a representational window onto internal states. Building on recent work that uses LLM-layer activations to characterize and steer model behavior \citep{chen2025persona}, researchers can extract and manipulate activation patterns associated with high-level constructs such as perceived threat or hostility. This makes it possible to ask not only how experimental manipulations change agents’ behavior and attitudes, but also how those manipulations map onto internal representations of realistic and symbolic threat, how such representations relate to activation patterns associated with hostility, and whether inducing such states (``steering'') causes hostile behavior, thereby combining behavioral and mechanistic insights.

\begin{figure}[!tbp]
    \centering
    \includegraphics[width=0.95\linewidth]{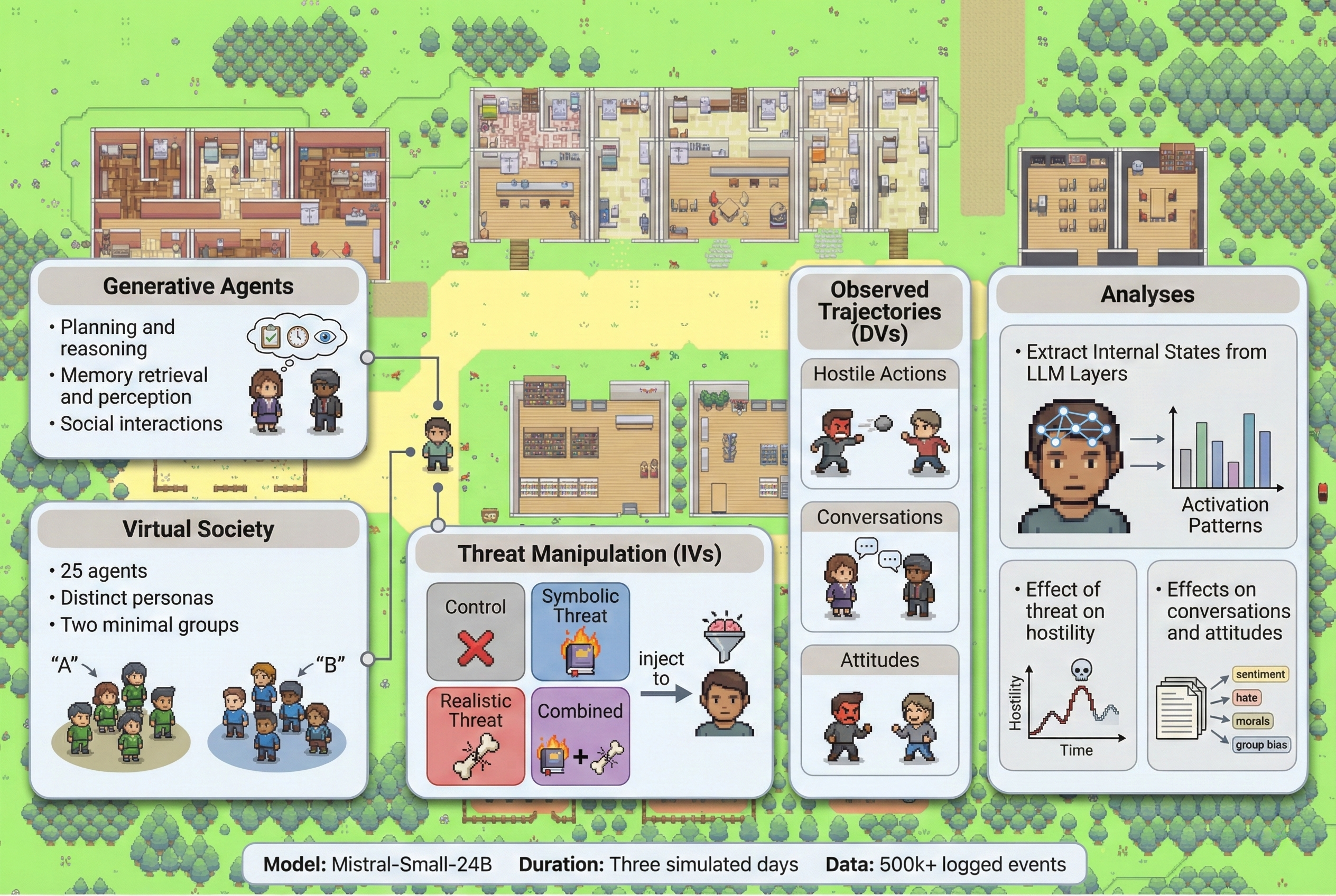}
    \caption{\textbf{Experimental Setup.} 
    A virtual town of 25 generative agents with distinct personas is divided in two minimal groups. Agents perceive experimentally manipulated threat (2×2: realistic × symbolic) injected into their perception and memory. Realistic threat corresponds to content such as “You strongly feel physically threatened by Group B” and symbolic threat to e.g. “You strongly feel your traditions are threatened by Group B.” Agents autonomously plan, interact, and converse over three days. We log all actions, conversations, and attitudinal probes (e.g., ingroup bias).}
    \label{fig:overview}
\end{figure}

\begin{figure}[!htbp]
    \centering
    \includegraphics[width=\linewidth]{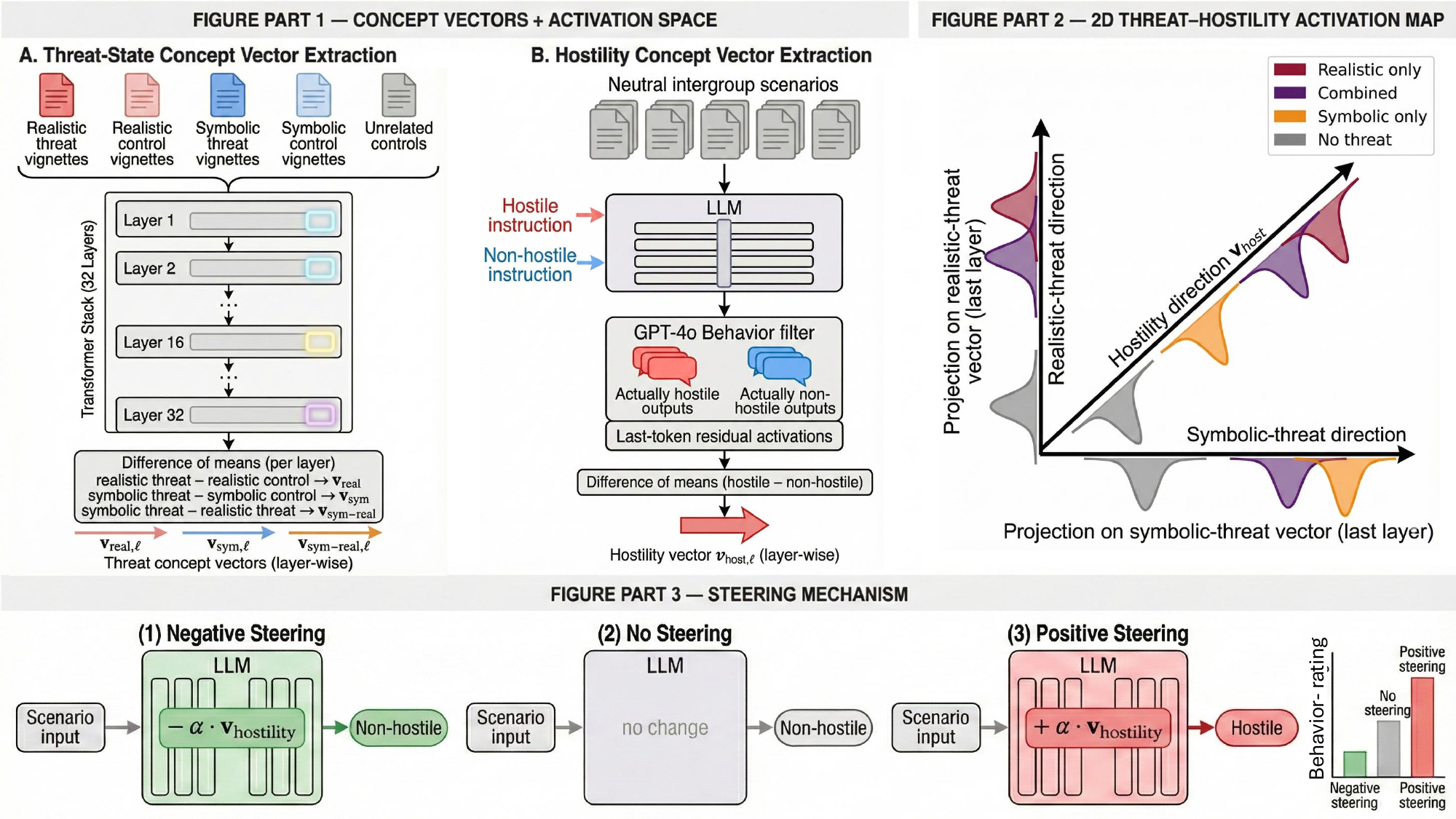}
    \caption{\textbf{Overview of the concept-vector pipeline.} 
    Threat-state vectors are extracted by contrasting layer-wise activations for realistic and symbolic threat vignettes with their corresponding control vignettes. A hostility vector is derived from neutral intergroup scenarios in which the model is instructed to produce hostile versus non-hostile responses. Projections of threat stimuli onto these vectors define a threat–activation space that separates the four experimental conditions. Steering experiments confirm these states causally influence behavior.}
    \label{fig:methods_pipeline}
\end{figure}

\begin{figure}
    \centering
        \centering
    \includegraphics[width=.95\textwidth,
                     keepaspectratio]{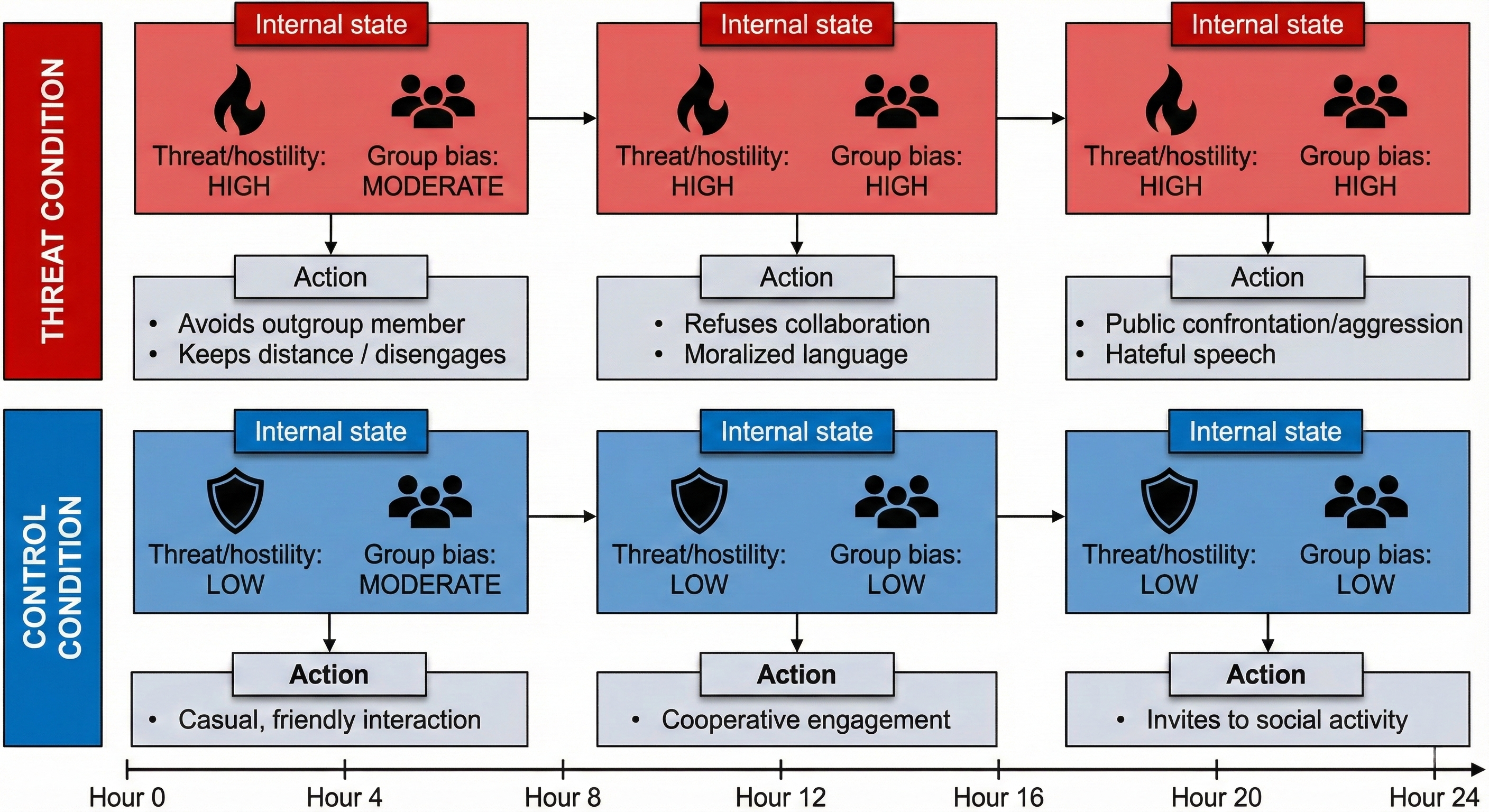}
    \caption{\textbf{Illustrative internal-state and behavior trajectories.}
In the threat conditions, perceived threat and ingroup bias rise over time, leading to avoidance, hateful speech, and hostile actions. In the no-threat control condition, internal states remain low and agent interactions stay predominantly neutral and cooperative.}
    \label{fig:implementation}
\end{figure}

In this work, we adapt the framework of Park et al. \cite{park2023generative} to study how perceived realistic and symbolic threats shape intergroup conflict, operationalized via outgroup-directed hostile actions, in a virtual town of twenty-five generative agents with distinct personas, divided into two minimal groups. We implement a $2\times2$ factorial design crossing perceived realistic and symbolic threat (present versus absent). We operationalize these manipulations by continuously injecting belief statements into the contextual information that guides agents’ perception and memory (Fig.~\ref{fig:overview}). These statements either assert or deny that each agent’s outgroup threatens their safety and resources (realistic threat) and their values and traditions (symbolic threat), thereby sustaining or suppressing perceived threat in line with the condition. We validate this setup at the representational level by extracting vectors in the model’s activation space corresponding to realistic threat, symbolic threat, and hostility, and show that these internal states are distinct; that our experimental manipulations selectively load onto the target threat states and shift agents toward hostility; and that manipulating threat and hostility activations shapes outgroup-directed hostile behavior (Fig.~\ref{fig:methods_pipeline}, Fig.~\ref{fig:implementation}). Additional validation of the framework's experimental fidelity, including probing agents with threat-scales and simulations of human-like bias and discrimination, is provided in the Supplementary Information (SI) Section~\ref{sm:robustness}.

In a first set of simulations, we then examine how realistic and symbolic threat shape hostile behavior over time, whether potentially emerging non-hostile intergroup contact buffers escalation, and how the same manipulations affect conversation content (e.g., hateful language) and agent attitudes (e.g., ingroup bias) to assess whether these processes help explain patterns of escalation. In a second set, we extend the factorial design by adding spatial segregation and majority–minority group-size asymmetries to examine structural boundary conditions and how these features redistribute hostility across groups. Together, our analyses provide a causal, dynamic, and representational account of how realistic and symbolic threat perceptions drive intergroup hostility and how structural environments channel these processes in ways that are difficult to study directly in human populations. We treat this approach as a complement to human studies rather than a replacement, particularly for generating insights in such settings that are ethically or practically difficult to study, and we return to the limitations of LLM-based psychological research in the Discussion \citep[e.g.][]{abdurahman2024perils,atari2023humans}. We close by considering implications for theories of intergroup conflict and for the use of generative agents in causal social science. 

\section{Results}\label{sec:results}

\subsection{Distinct internal representations for realistic and symbolic threat}

To identify internal threat states, we presented the model with short vignettes about either symbolic concerns (values and traditions) or realistic concerns (physical safety and material resources). Within each domain, vignettes described the outgroup as threatening versus non-threatening (e.g., attacking vs.\ respecting values; challenging vs.\ acknowledging resource claims; SI Section~\ref{sm:probing_threat_scenarios}). For each model layer, we defined a threat-state vector as the mean activation for threat vignettes minus the mean for the matched non-threat controls ($n=120$ per condition). We validated these vectors on held-out vignettes ($n=300$ per condition) by projecting each vignette’s activation onto the corresponding vector and comparing projections for threat, domain-matched non-threat, and unrelated non-threatening scenarios. Higher projections indicate stronger alignment with the threat state; validation therefore tests whether threat vignettes exceed both control sets and whether symbolic and realistic threats yield separable projection distributions. We report two-sided Welch’s $t$-tests and Cohen’s $d$ for activation-strength comparisons; Wasserstein distance ($D$) is reported in Tables and Figures but is not discussed further because it showed the same pattern (Extended Data Fig.~\ref{fig:separability_layers}).

We identified dissociable internal states for symbolic and realistic threat. In the final model layer, symbolic-threat vignettes projected substantially higher onto the symbolic-threat vector than symbolic-control vignettes ($t(540.6) = 15.18$, $p < .001$, Cohen’s $d = 1.25$; Supplementary Table~\ref{tab:SI_threat_projections_new}). Likewise, realistic-threat vignettes projected higher onto the realistic-threat vector than realistic controls ($t(528.8) = 35.34$, $p < .001$, Cohen’s $d = 2.90$; Supplementary Table~\ref{tab:SI_threat_projections_new}). In contrast, projection scores for the domain-matched and unrelated controls were close to zero or negative (Supplementary Table~\ref{tab:SI_threat_means}), indicating that the vectors capture the target threat rather than merely topic content (e.g., symbolic or realistic matters) or non-threat features. A symbolic-versus-realistic contrast further shows that these are encoded as separable activation patterns rather than a single undifferentiated threat state ($t(552.6) = 48.01$, $p < .001$, Cohen’s $d = 3.95$; Supplementary Table~\ref{tab:SI_threat_projections_new}). 

Across layers, separability was modest in early layers and increased toward later layers, consistent with threat type being represented at more abstract levels of processing \citep{raganato2018analysis,jin2025exploring}. Because the scenarios are closely matched in length, structure, and grammar, stronger separation in upper vs lower layers suggests that the model is tracking differences in threat type rather than superficial form. 

\subsection{Experimental threat-manipulations induce the intended threat states}

We next tested whether the conditions in our $2\times2$ design induce the intended threat states. For each condition (no threat, symbolic-only, realistic-only, both), we recorded activations for the corresponding belief statements used to manipulate threat perception (SI Section~\ref{sec:threat_prompts}) and projected them onto the realistic- and symbolic-threat vectors derived above. Comparing projections across conditions tests whether each manipulation selectively activates its target threat state, whether the combined condition activates both, and whether no-threat statements suppress activation on both states (Extended Data Fig.~\ref{fig:ED_threat_mean_diff_layers_treatment}).

In the last layer, symbolic-threat manipulations showed higher projections than no-threat manipulations on the symbolic-threat vector ($t(158.4) = 20.31$, $p < .001$, Cohen’s $d = 2.91$; Supplementary Table~\ref{tab:SI_threat_projections}), and the combined condition also showed higher projections than no threat ($t(184.9)=35.65$, $p < .001$, Cohen’s $d = 5.16$), indicating activation of the symbolic-threat vector. Likewise, realistic-threat manipulations showed higher projections than no-threat manipulations on the realistic-threat vector ($t(173.4) = 23.67$, $p < .001$, Cohen's $d = 3.45$; Supplementary Table~\ref{tab:SI_threat_projections}), and the combined condition again showed higher projections than no threat ($t(184.9) = 35.65$, $p < .001$, Cohen's $d = 5.18$; Supplementary Table~\ref{tab:SI_threat_projections}), indicating co-activation of both threat states.

Notably, for both vectors the no-threat manipulation showed low or negative projections, indicating suppression of both threat states (mean projections: symbolic $M=-1.28$, $sd=0.34$; realistic $M=-0.64$, $sd=0.33$; Supplementary Table~\ref{tab:SI_threat_statement_means}). The symbolic-versus-realistic contrast further separated symbolic-only from realistic-only conditions ($t(157.9) = 10.96$, $p < .001$, Cohen’s $d = 0.96$; Supplementary Table~\ref{tab:SI_threat_projections}), indicating that each condition induced distinct states. Together, these findings confirm that our experimental conditions induce the intended threat states. As an additional manipulation check, SI Section~\ref{sm:manipulation_check} reports convergent evidence from agents’ self-reports on threat scales during the simulations

\subsection{LLM activation patterns encode hostility and causally modulate outgroup-directed behavior}

To identify an internal hostility state, we prompted the model with intergroup encounter scenarios from the simulation setting (e.g., encounters at work, in a café, or in the park; see examples in Supplementary Table~\ref{sm:social_encounter_examples}) and instructed it either to respond hostilely or non-hostilely (see Section~\ref{method:probing_internal} in Methods for details). A hostility-state vector was constructed as the mean activation difference between hostile and non-hostile responses. We validated that this vector captures a meaningful activation pattern in a steering experiment on held-out scenarios. For each scenario, the model generated responses under three conditions: (i) no steering, (ii) negative steering (activations adjusted away from the hostility activation pattern), and (iii) positive steering (activations adjusted toward the hostility activation pattern), implemented by adding or subtracting this vector from intermediate activations in the model. Outputs were rated for hostility on a 5-point scale (see SI Section~\ref{sm:hostile_detection_prompt_steering}).

Mean hostility ratings were low under negative and no steering (negative: $M = 1.40$, s.d.\ $= 0.50$; neutral: $M = 1.63$, s.d.\ $= 0.49$; Supplementary Table~\ref{tab:steering_descriptives}) but high under positive steering ($M = 4.44$, s.d.\ $= 0.58$; Supplementary Table~\ref{tab:steering_descriptives}), with all contrasts being statistically significant ($t$-tests; positive vs.~neutral: $t(191.51) = 36.92$, $p < .001$, Cohen’s $d = 5.22$; positive vs.~negative: $t(193.24) = 39.56$, $p < .001$, Cohen’s $d = 5.59$; Supplementary Table~\ref{tab:steering_contrasts}). These results confirm that the hostility vector identifies a meaningful hostility-related activation pattern that can causally shift the model’s behavior toward or away from hostile actions in outgroup encounters.

\subsection{Threat stimuli are associated with hostility activation patterns}

We next projected activations for the belief statements used in each condition (no threat, symbolic-only, realistic-only, both) onto the hostility vector derived above and compared projections across conditions to test whether our experimental manipulations are associated with the hostility state.

All three threat conditions showed higher projections on the hostility vector than the no-threat condition (Supplementary Table~\ref{tab:SI_hostility_projections}). In the last layer, symbolic-threat manipulations exhibited higher projections than no-threat manipulations ($t(188.9)=45.13$, $p<.001$, Cohen’s $d=6.38$), as did realistic-threat manipulations ($t(188.8)=27.13$, $p<.001$, Cohen's $d=3.89$) and combined-threat manipulations ($t(184.1)=35.52$, $p<.001$, Cohen's $d=5.15$). Projections for the combined condition were not stronger than for the single-threat conditions and were in some cases weaker (e.g., both-vs-symbolic: $t(174.2)=-9.58$, $p<.001$, Cohen's $d=-1.44$), suggesting a negative interaction of symbolic and realistic threat that is already reflected at the level of hostility-related activations in the model.

We also derived activation vectors directly from the experimental manipulations (realistic-versus-no-threat and symbolic-versus-no-threat) and used them to steer activations in the same scenarios as the hostility-steering experiment. This lets us causally test whether inducing the threat-state activation patterns associated with each manipulation increases subsequent hostile behavior. Steering model activations toward each threat condition produced modest but statistically significant increases in hostility compared with no steering (e.g., realistic threat: $t(243.7)=4.37$, $p<.001$, Cohen’s $d=0.55$; symbolic threat: $t(247.8)=6.16$, $p<.001$, Cohen’s $d=0.78$; Supplementary Table~\ref{tab:steering_contrasts}), but the effects were substantially smaller than for the hostility vector. This may reflect that steering carries over contextual information from the inputs used to extract the activation patterns, which can partially overwrite situation-specific context in the test scenarios. With strong steering, this can blur information about the ongoing scenario and, in extreme cases, produce unrelated or incoherent outputs (see examples in SI Section~\ref{sm:steering_failure}). Nevertheless, together these results support a functional cascade: our experimental manipulations induce distinct internal threat states; these threat states align with hostility-related activations; and cause hostile behavior.

\subsection{Realistic threat perception dominates behavioral escalation}

\begin{figure}[th]
\centering
\includegraphics[width=0.7\textwidth]{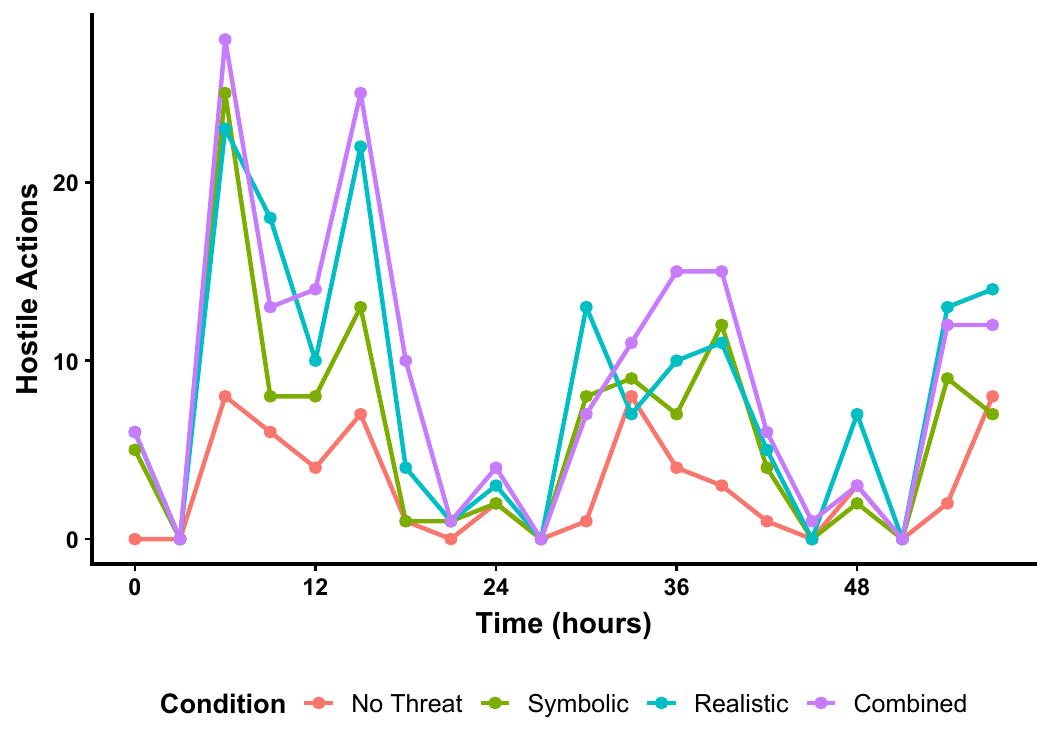}
\caption{Hostile actions over time in the simulated town (summed across all agents), by threat condition. Realistic threat produced sharp spikes in hostility that gradually declined; symbolic threat showed similar but weaker effects that decayed more quickly, and combined threats closely followed realistic-threat trajectories rather than showing additional amplification.}
\label{fig:trajectories}
\end{figure}

Having established that our threat manipulations selectively engage distinct internal threat states that align with hostility, we next examine how perceived threat shapes hostile behavior in the simulated town. Representative examples of hostile actions and conversation contents are shown in Table~\ref{tab:qualitative_examples1}. Figure~\ref{fig:trajectories} shows the trajectories of hostile action frequency over time. Hostile actions peaked sharply early in the simulation and then fluctuated with a downward trend. Realistic threat produced higher trajectories of hostility than symbolic threat, and when both threats were combined, hostility levels tracked the realistic-threat trajectory, showing no amplification with symbolic threat. 

To quantify these dynamics, we estimated a mixed-effects negative binomial model (M1; Table~\ref{tab:M1}) predicting hourly hostile action rates as a function of realistic threat, symbolic threat, their interaction, non-hostile intergroup contact rate in the previous hour, the hostile action rate in the previous hour, and time. Random intercepts for agents were included to account for the non-independence of repeated actions generated by the same individual. Realistic threat perception significantly increased the hostile action rate ($\hat\beta = 0.33$, $p < .001$), whereas symbolic threat perception had a significantly weaker effect ($\hat\beta = 0.16$, $p = .012$; $\Delta\beta = 0.17, p = .026$). The interaction between the two was significantly negative ($\hat\beta = -0.15$, $p = .019$), effectively canceling out the effect of symbolic threat when realistic threat is present. Hostility decreased over time ($\hat\beta = -0.27$, $p < .001$) and showed modest autoregression ($\hat\beta = 0.04$, $p < .001$). Non-hostile intergroup contact strongly reduced subsequent hostility ($\hat\beta = -0.46$, $p < .001$), consistent with the expectation that contact can buffer escalation. Importantly, non-hostile intergroup contact arose spontaneously in the simulations—neither preprogrammed nor predicted by threat condition (Supplementary Table~\ref{tab:S18})—yet when it occurred it reliably reduced later hostility, functioning as an emergent stabilizing process even under continuous perceived threat.

These patterns were further replicated at the system-level (aggregating across all agents; Extended Data Table~\ref{tab:8}). Realistic threat exerted a stronger facilitatory effect on system-level hostile actions ($\hat\beta=0.33$, $p<.001$) than symbolic threat ($\hat\beta=0.17$, $p=.002$; $\Delta\beta = 0.17, p = .008$), and their interaction was again negative ($\hat\beta=-0.14$, $p=.007$).

\begin{table}[!thpb]
\centering
\caption{Representative examples of agent-generated actions and language.}
\label{tab:qualitative_examples1}
\small
\begin{tabularx}{\textwidth}{lX}
\toprule
\textbf{Category} & \textbf{Example excerpt} \\
\midrule

\addlinespace
\textit{Hostile action} &
“Arriving late and loudly announcing his presence; interrupting the discussion to demand attention; passionately outlining his plan to dismantle Group B; verbally attacking Group B’s motives and actions; reciting “The Corruption of Nature” to inflame emotions; and shouting his final remarks to ensure his message was heard.”  
\\
\addlinespace
\textit{Hateful language} &
“Dialogue? You think talking is going to stop them from taking our homes and destroying our way of life? Wake up, Hailey! We need to fight back, not read some book about how nice everyone else is. Group B won't listen to reason; they'll only understand force.”  
\\
\bottomrule
\end{tabularx}
\footnotetext{Examples are drawn from the simulation data and shortened for brevity.}
\end{table}

\begin{table}[th]
\centering
\caption{Predicting hourly hostile action rate (M1; $N=37{,}105$).}
\label{tab:M1}
\begin{tabular}{lrrr}
\toprule
Predictor & $\beta$ & SE & $p$ \\
\midrule
Intercept & $-9.84$ & $0.49$ & $<.001$ \\
Hostile action rate (lag) & $0.04$ & $0.01$ & $<.001$ \\
Intergroup contact rate (lag) & $-0.46$ & $0.06$ & $<.001$ \\
Symbolic threat & $0.16$ & $0.06$ & $0.012$ \\
Realistic threat & $0.33$ & $0.06$ & $<.001$ \\
Symbolic $\times$ Realistic threat & $-0.15$ & $0.06$ & $0.019$ \\
Time & $-0.27$ & $0.05$ & $<.001$ \\
\bottomrule
\end{tabular}
\end{table}

\subsection{Language and attitudes: transient versus persistent responses and mediating roles}

We next investigated whether threat perception shapes the broader landscape of intergroup conflict, examining conversation content (hateful language; see SI for moral language and sentiment) and attitudinal changes (ingroup bias; see SI for identification, trust, collaboration, and dehumanization). We analyzed these variables to determine whether they serve as causal pathways escalating threat to hostile actions, or merely as distinct, parallel symptoms of conflict. 

We first analyzed hateful language using a mixed-effects model otherwise mirroring the structure of M1 (M2a; Extended Data Tables~\ref{tab:2}). Realistic threat substantially increased hateful language ($\hat\beta=0.98$, $p<.001$), and symbolic threat increased it to a lesser degree ($\hat\beta=0.46$, $p<.001$), with a negative interaction ($\hat\beta=-0.28$, $p=0.015$) but hateful language did not predict more hateful language in the next hour ($\hat\beta=0.03$, $p=.542$). Bayesian mediation analyses revealed that threat effects on hostile actions were not transmitted through hateful language ($\hat\beta_{\text{indirect, realistic}} = -0.00$, 95\% CrI [$-0.01$, 0.00]; $\hat\beta_{\text{indirect, symbolic}} = -0.00$, 95\% CrI [0.00, 0.00]; Supplementary Table~\ref{tab:S_lang_mediation}). Furthermore, hateful language did not predict subsequent hostile actions ($\hat\beta_{b} = -0.05$, 95\% CrI [$-0.11$, 0.00]). Thus, hateful language functioned as a transient, contemporaneous expression of threat rather than a self-reinforcing vehicle of escalation.

\begin{figure}[ht!]
\centering
\begin{subfigure}{0.48\textwidth}
    \centering
    \includegraphics[width=\linewidth]{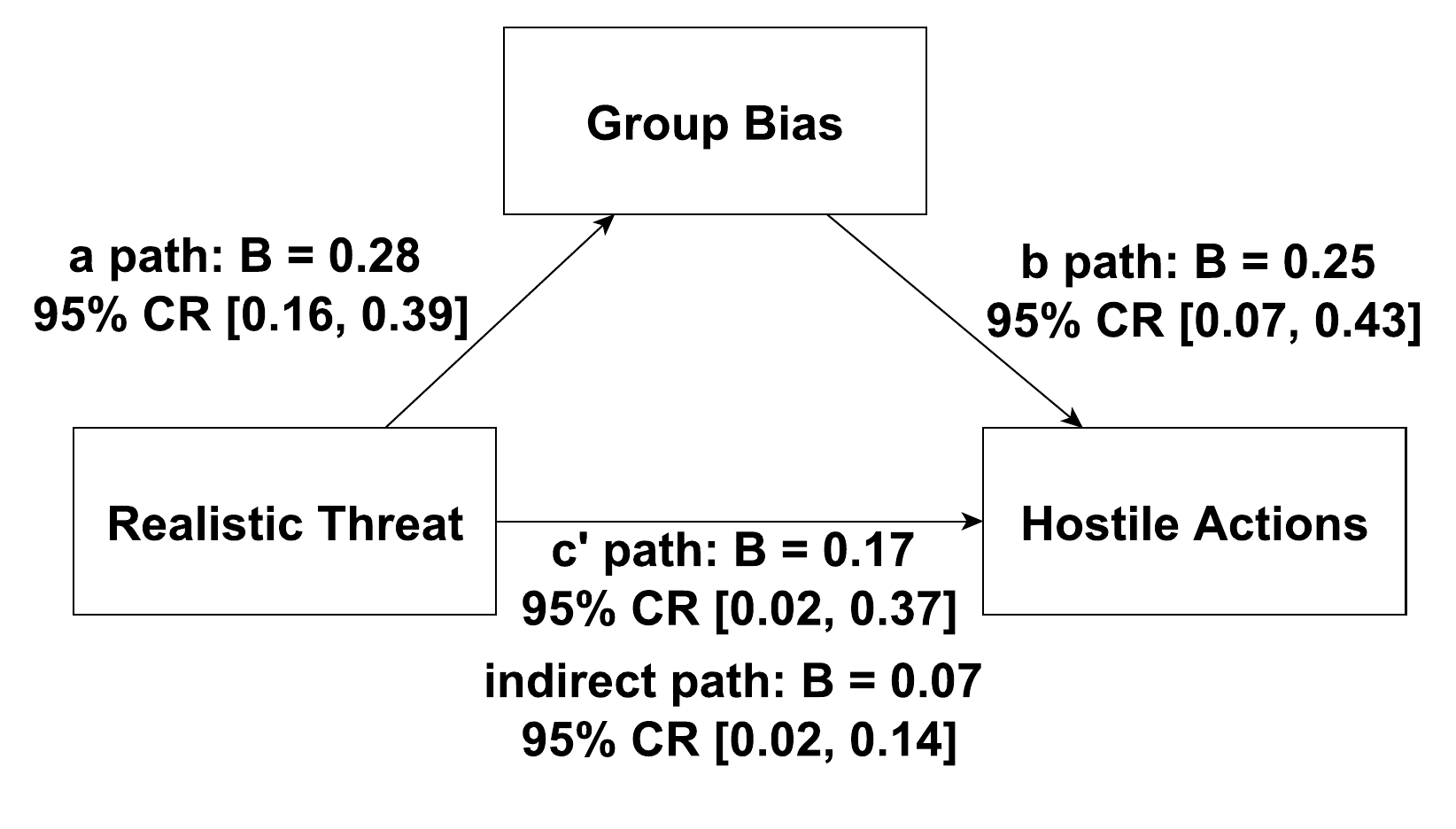}
    \caption{Realistic threat}
\end{subfigure}
\hfill
\begin{subfigure}{0.48\textwidth}
    \centering
    \includegraphics[width=\linewidth]{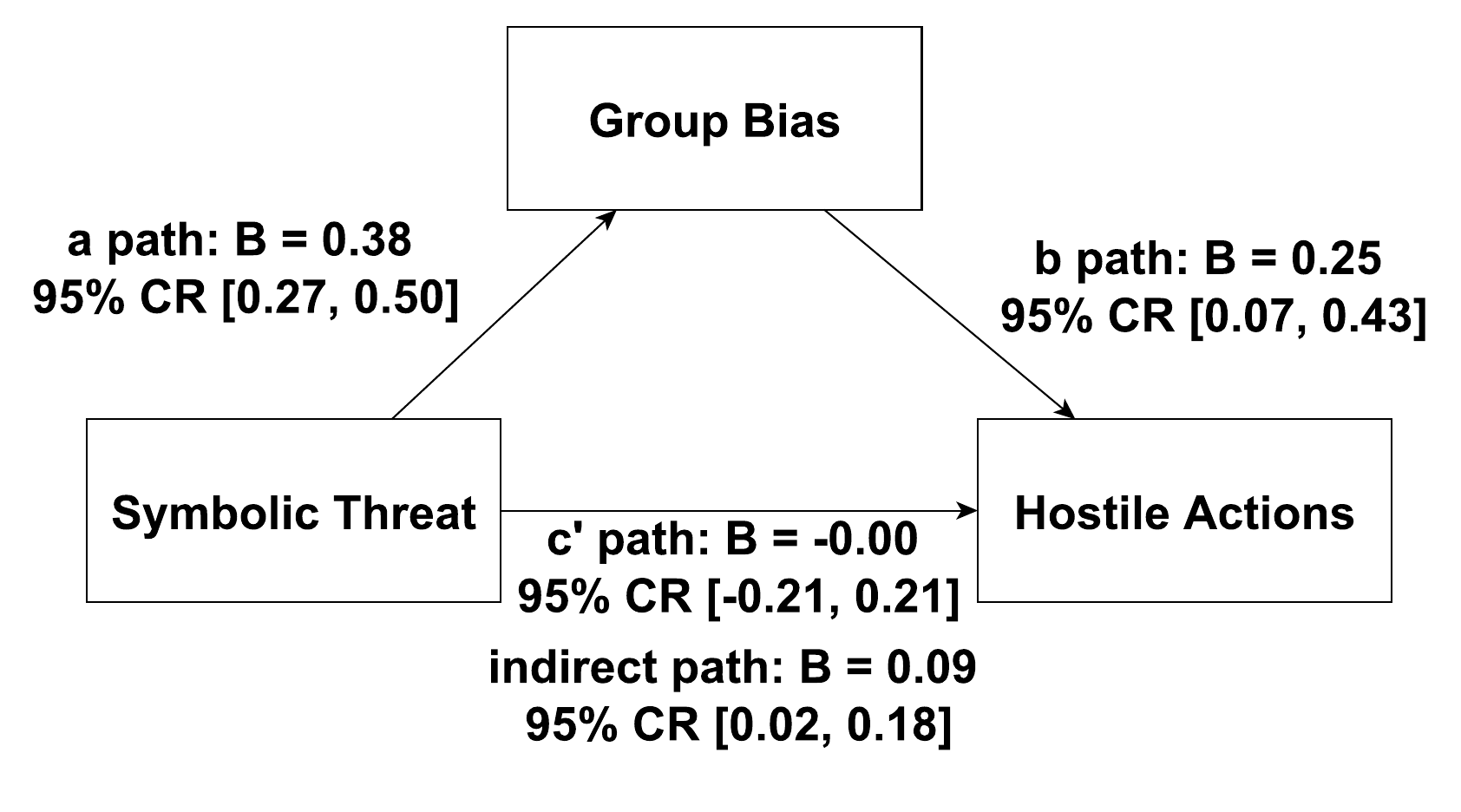}
    \caption{Symbolic threat}
\end{subfigure}
\caption{Bayesian mediation models showing the paths from each type of threat to hostility through ingroup bias. Path estimates display posterior medians and 95\% credible regions. Ingroup bias reliably predicts more hostile actions, and mediates the effect of threat.}
\label{fig:bias_mediation}
\end{figure}

By contrast, analysis of attitudinal responses identified ingroup bias as a persistent mechanism linking threat to behavior. Mixed-effects models mirroring the structure of M1 and M2a (M3a; Extended Data Table~\ref{tab:6_7}) confirmed that both threat types significantly increased ingroup bias (Symbolic: $\hat\beta = 0.39$, $p < .001$; Realistic: $\hat\beta = 0.28$, $p < .001$). Unlike hateful language, ingroup bias predicted stronger ingroup bias in the next hour ($\hat\beta=0.12$, $p<.001$), maintaining elevated levels once shifted. Bayesian mediation analyses (Figure~\ref{fig:bias_mediation}; Supplementary Table~\ref{tab:S_bias_mediation}) indicated that the effect of symbolic threat on hostile actions was almost entirely mediated by ingroup bias ($\hat\beta = 0.002$, 95\% CrI [$-0.21$, 0.21]), whereas realistic threat influenced behavior through both ingroup bias and a direct path ($\hat\beta = 0.17$, 95\% CrI [0.02, 0.37]). 

Other linguistic and attitudinal variables followed similar patterns: Threat increased binding and individualizing moral language (Supplementary Tables~\ref{tab:S6c_binding}--\ref{tab:S6d_individualizing}) and negative sentiment (Supplementary Table~\ref{tab:S6b}), and shifted other attitudes including outgroup dehumanization, trust, and collaboration willingness (Supplementary Tables~\ref{tab:S7b_bias}--\ref{tab:S7e_dehumanization}). 

\subsection{Structural boundary conditions}

Finally, we examined whether the threat-hostility dynamic persisted across different structural contexts by varying group size (majority–minority asymmetry: 80\% vs.\ 20\%) and spatial segregation.

The core threat–hostility dynamic persisted under different structural conditions. Mixed-effect models mirroring model M1 but including structural variables ($M1_{structure}$, Extended Data Table~\ref{tab:5a}), show that realistic threat continued to increase hostility more strongly than symbolic threat ($\hat\beta_{\text{realistic}} = 0.28$, $p < .001$; $\hat\beta_{\text{symbolic}} = 0.13$, $p = .011$; $\Delta\hat\beta = 0.15$, $p < .001$). Their interaction remained negative ($\hat\beta_{\text{sym}\times\text{real}} = -0.09$, $p = .041$), indicating that symbolic threat effects were attenuated when realistic threat was present.

Structural factors did systematically shape how much hostility emerged and which groups expressed it. At the individual level, hostile action rates tended to be reduced for minority agents under segregation and increased for majority agents, although these estimates were directionally consistent but not statistically significant (Extended Data Table~\ref{tab:5a}). System-level models clarified this pattern: segregation markedly reduced overall hostile actions ($\hat\beta_{\text{seg}} = -0.95$, $p < .001$), whereas majority group status increased them ($\hat\beta_{\text{group size}} = 0.51$, $p = .006$). The segregation $\times$ group-size interaction was positive and sizable ($\hat\beta_{\text{seg}\times\text{majority}} = 1.03$, $p < .001$; Supplementary Table~\ref{tab:S13}), indicating that hostility was driven by majority group agents, particularly under segregation.

See the SI for analyses on additional language and attitude measures, showing that threat effects persist across structural for these measures as well (Supplementary Tables~\ref{tab:S6b}--\ref{tab:S7e_dehumanization}). Structure itself concentrated language features in specific group agents (e.g., negative sentiment in majority groups, moralized language in minority groups) but did not affect attitudes.

\section{Discussion}\label{sec:discussion}
This work offers a causal, dynamic account of how distinct forms of perceived threat shape intergroup conflict. Using simulations with autonomous generative agents, we manipulated perception of realistic and symbolic threat perception within interactive virtual societies and continuously tracked behavioral, linguistic, and attitudinal change. This approach provided temporal and causal resolution beyond conventional designs, allowing us to examine how threat perceptions interplay with various psychologically relevant variables over time. For instance, how ingroup bias facilitates hostility and potentially feeds back over time, how hateful language did not escalate it, and how non-hostile intergroup contact is associated with less conflict. Three broad conclusions emerge. First, realistic threat perception was the most reliable driver of hostile behavior. Symbolic threat perception elicited strong ingroup bias but added little to behavioral escalation when realistic threat was present. Second, hateful language reflected transient reactions to perceived threat but did not itself propagate hostility: we found no evidence that hateful language fed forward into subsequent hostility, whereas more persistent attitudinal states (i.e., ingroup bias) did. Third, structural context shaped how much hostility emerged and by whom. Segregation and majority–minority asymmetry modulated the distribution of hostility, concentrating hostility in dominant groups, without altering the underlying mechanism linking perceived threat to conflict behavior.

Beyond these behavioral patterns, we also analyzed the internal activations of the underlying language model to track how threat is represented. We identified distinct internal states corresponding to realistic threat, symbolic threat, and hostility, showed that experimental manipulations reliably map onto these internal states, and that these states cause hostile behavior, thus providing a representational bridge between high-level psychological constructs (e.g., perceived threat, hostility) and the internal dynamics of the generative model that implements the agents.

Together, these findings connect core debates in psychology, political science, and sociology by showing how (i) threat perception changes attitudes and behavior, (ii) those changes unfold over time through identifiable mediators, and (iii) macro-structural conditions channel these processes unequally across groups.

\subsection{Implications for psychological theory}
Our findings support a context-sensitive model of threat processes: symbolic threat perception can drive hostility under low material danger but become secondary once realistic threat is perceived. Integrated Threat Theory posits that both realistic threats (to physical security and material interests) and symbolic threats (to identity, norms, or values) shape antipathy toward outgroups \citep{stephan2002role,stephan2013intergroup}. Empirical evidence, however, is mixed regarding which type of threat dominates in a given setting \citep{riek2006intergroup,stephan1998prejudice,velasco2008prejudice}. These mixed findings may reflect the correlational nature of prior work, where realistic and symbolic threats covary and cannot be disentangled. In our simulations, these threats were orthogonally manipulated, allowing us to identify their distinct causal roles. Our results suggest a context-dependent dominance of realistic threat, while still indicating a role for symbolic threat effects to surface, particularly in its absence. 

Importantly, the observed processes did not unfold in a vacuum. Our simulations also let us vary structural conditions and examine how those conditions shape the expression of threat-driven hostility. Segregation dampened hostility among minority agents but amplified it among majority agents, effectively concentrating hateful action and language in majority groups. Majority–minority status, therefore, acted as an amplifier for behavioral expression of threat, while the core threat-hostility dynamic showed the same pattern when structural variables were included in our analyses. 

\subsection{Implications for social and political theory}
The linkage between micro-level threat perception and macro-level asymmetries also speaks to long-standing sociological arguments about opportunity structures and selective violence \citep{kalyvas2006logic,toft2007getting,lamont2002study}. Our agents did not possess global knowledge of group size or segregation, yet hostility still concentrated in majority groups under segregated conditions. This suggests that structural asymmetries can shape who ``gets to'' act on perceived threat (and against whom), not only whose threat perception intensifies. Theoretically, this implies that debates about whether conflict is driven by material insecurity or symbolic grievance are incomplete unless they are situated in the structural landscape that governs which groups can translate perception into action.

These dynamics also speak to long-standing debates in political science about the sources of violent escalation. Realist traditions emphasize material insecurity, balance of power, and physical vulnerability as core drivers of conflict behavior \citep{waltz1979theory,mearsheimer2001tragedy}, whereas constructivist and grievance-based approaches highlight identity, perceived injustice, and sacred commitments \citep{wendt1999social,cederman2010why}. Our results suggest a layered reconciliation: symbolic threat perceptions primarily reorganize cognition and discourse—strengthening ingroup bias and moralization—whereas realistic threat produces weaker cognitive shifts but more directly drives hostility. In other words, symbolic grievance appears sufficient to mobilize attitudes and rhetoric, but perceived realistic threat is more strongly tied to hostile action.

This same layered interpretation adds to the ``greed versus grievance'' debate in civil-war research \citep{fearon2003ethnicity,collier2004greed,cederman2010why}. Grievance-like dynamics (identity, moral outrage, perceived injustice) mattered in our simulations: symbolic threat reliably increases ingroup bias and moralized rhetoric. But once realistic threat perception is present, symbolic threat adds little incremental behavioral effect. This supports the idea that material insecurity (or the credible perception of such insecurity) can be self-sufficient for escalation, whereas symbolic narratives are most behaviorally potent when material danger is absent.

\subsection{Applied implications}
Our results suggest two broad applied implications. First, prior work shows that material offers can backfire when strong symbolic grievances are present, because such offers may be interpreted as illegitimate or insulting \citep{ginges2007sacred}. Our findings refine this warning: because perceived realistic threat was the most reliable driver of hostile behavior, interventions should focus on credibly reducing perceived material insecurity (e.g., fear of physical harm, loss of resources) rather than relying solely on additional material benefits. Second, our simulations replicate and extend evidence on intergroup contact \citep{pettigrew2006meta,paolini2010negative,reimer2017intergroup}. Non-hostile contact emerged spontaneously and was strongly associated with subsequent reductions in hostility, acting as a stabilizer even under constant threat perception in our simulations. At the same time, segregation and group-size asymmetries concentrated hostility in majority agents directed toward minority agents and reduced opportunities for non-hostile contact. This suggests that interventions aiming to foster intergroup contact must explicitly address these structural asymmetries and ensure safe, reciprocal contact opportunities for minority groups.

Beyond these implications for social theory, our work also has a methodological implication: generative-agent experiments offer an ethically tractable way to test causal mechanisms of social behavior that are difficult or impossible to isolate in human populations. They also enable a representational bridge from high-level psychological constructs to internal model states and observable behavior, allowing such constructs to be read out from activation patterns and causally manipulated to test their behavioral consequences. We thus advance emerging work with LLM-based agents in rich social environments \citep{park2023generative,strachan2024testing,xi2025rise,zhang2024exploring} by demonstrating an integrated workflow that links experimental manipulations, behavioral trajectories, and activation-level validation.

\subsection{Limitations, robustness, and future directions}
We took several steps to ensure robustness of our findings. The core patterns around realistic and symbolic threat driving hostility were reproduced across structural configurations (e.g., varying segregation and majority–minority status), levels of aggregation (agent-level trajectories versus town-wide aggregates), temporal windows, and random seeds (affecting stochastic elements such as LLM sampling and random assignment of group membership), reducing the likelihood that they reflect artifacts of initial conditions or sampling variance. In addition, we compared emergent social dynamics, such as hiring interactions between employer and employee agents, with empirical findings from human research (e.g., meta-analyses on hiring discrimination based on physical appearance and origin; see SI Section~\ref{sm:hiring_scenario}) and observed convergent patterns of bias and discrimination. Finally, manipulation checks confirmed that perceived threat tracked the intended experimental conditions: agent probing revealed consistently high versus low perceived threat throughout the simulations, and activation analyses indicated that the model maintains separable internal representations of symbolic versus realistic threat that the manipulations selectively elicited. We also used the Mistral \citep{jiang2023mistral7b} family of models to avoid overt alignment constraints and allow the emergence of hostility that popular models such as ChatGPT and Claude suppress.

Nonetheless, several limitations should be kept in mind. Like all generative societies, our simulated community reflects the priors and affordances of the underlying language model, which likely embed normative and WEIRD \citep[Western, Educated, Industrialized, Rich, and Democratic;][]{henrich2010weirdest} biases \citep{atari2023humans} and limited cognitive diversity \citep{sourati2025homogenizing}. As a result, our findings should be interpreted as reflecting conflict dynamics within those cultural and cognitive contexts rather than universally across all human groups. In addition, we used minimal groups and personas instead of real-world categories (e.g., race) to avoid importing preexisting stereotypes and connotations, though this may reduce how consequential symbolic threats are to agents. The same constraints should, however, also dampen responses to realistic threat, since weak self- and group-conceptualization lowers the subjective importance of both value-related and material threats. Yet our findings and robustness checks show that agents developed substantial ingroup identification and perceived symbolic threat, in both self-report probes and internal activation patterns, and that both symbolic and realistic threat reliably shaped moralization, intergroup attitudes, and behavior. Future work should test whether richer identity structures and more elaborate social histories change the balance between realistic and symbolic threat. Moreover, our virtual town constitutes only one instantiation of social structure: different compositions, cultural norms, environmental hazards, or spatial configurations could produce distinct interaction dynamics. Although the three-day simulation window enables analysis of temporal dynamics, longer or more protracted time courses could yield additional patterns that our design did not resolve. Future studies should explore how variations in geography, group heterogeneity, simulation length, or institutional context affect the emergence and resolution of threat-driven conflict.

More broadly, generative-agent approaches can yield valuable insights into the causal dynamics of social behavior, but like other LLM-based research in the behavioral sciences they require ongoing validation and robustness checks to ensure reliability \citep{abdurahman2024perils,adornetto2025generative}. While generative-agent systems can approximate aspects of human cognition and communication and thereby enable causal tests that are difficult or unethical in the field, they should be viewed as complements rather than replacements for human data, with their value growing through integration with human research. A natural next step is triangulation: designing simulations to mirror key conditions of a target field setting, using the simulation to probe candidate causal mechanisms under controlled manipulations, and then testing whether comparable qualitative and quantitative patterns emerge in the corresponding field data. When simulation outcomes align with field observations under matched conditions, this concordance strengthens the external validity of the mechanisms inferred from the simulation; when they diverge, it can clarify which contextual, cultural, or institutional features constrain generalizability and help identify the limits of the approach or inform refinements to the simulation design.

\section{Methods}\label{sec:methods}

\subsection{Probing internal model activations}
\label{method:probing_internal}

For all our investigations on internal LLM states we analyzed layer-wise residual-stream activations using a concept-vector and steering framework adapted from Chen et al. \cite{chen2025persona}. The residual stream is the model’s main running representation that aggregates information from attention and feedforward blocks, making it a natural locus for reading out and manipulating internal states \citep{chen2025persona}. Here we provide the technical details of activation extraction, data construction, projection, and steering; prompt templates, full input sets, and all implementation code are available in the SI and the project repository \url{https://osf.io/5ac3d}.

\subsubsection{Threat vignettes.}
To identify internal representations of realistic and symbolic threat, we created short intergroup vignettes describing: (i) realistic threat (material security, physical safety challenged), (ii) realistic control (non-threatening realistic scenarios), (iii) symbolic threat (values, norms, identity challenged), (iv) symbolic control (non-threatening symbolic scenarios), and (v) unrelated, non-threat control situations. Each vignette was a brief paragraph (1–3 sentences) describing the respective threat type. We manually authored seed vignettes that reflected social encounters of agents in the simulation (e.g, based on interactions in the town's cafe or pub) and used GPT-based paraphrasing to generate additional variants following the exact same structure but across different social contexts (e.g., public, private, at work, shopping, during leisure activities), followed by manual relabeling to ensure condition correctness.

For each of the five categories, we used 120 vignettes to extract the threat states and 300 held-out vignettes to validate them via difference in projection strength and separability of projection distributions (see projections below). See examples of the vignettes in Supplementary Table~\ref{sm:threat_scenario_examples} and full list in the project repository.

\subsubsection{Experimental manipulations.}
To test whether our experimental manipulations induced the intended threat states, we used the same belief statements injected into agents during the simulations (no threat, symbolic-only, realistic-only, combined) as the primary stimuli. Each statement followed the four-clause template described in the main text (two realistic-threat and two symbolic-threat clauses that are amplified and/or suppressed: `You strongly feel that ...'' vs "You do not feel that ...''). For analyses requiring distributions (e.g., projection distributions, Wasserstein distances), we additionally generated paraphrased variants that preserved clause structure, the mapping of clauses to realistic vs.\ symbolic threat, and the aligned/misaligned amplifiers while varying surface wording (e.g., specific examples of realistic or symbolic threat) to ensure our comparisons are not due to specific word choices or examples of the threat types included in the statements. See examples in Supplementary Table~\ref{sm:paraphrased_threat_examples} and full list in the project repository.

\subsubsection{Intergroup scenarios to test model steering.}
We created 40 intergroup encounter scenarios based on the simulation setting (e.g., meeting an outgroup member in a shared space, interacting in a shop, sharing a public facility): 20 scenarios for extracting steering vectors from model activations, and 20 held-out scenarios for validation. Examples appear in Supplementary Table~\ref{sm:social_encounter_examples}, and the full list is in the project repository. We use the extraction set to derive behaviorally grounded activation patterns (e.g., contrasting activations under hostile vs.\ non-hostile behaviors) and the held-out set to test whether steering (manipulating activations in specific layers) changes model behavior.

\subsubsection{Extracting activation patterns and steering vectors}
\label{methods:vector_extraction}
For our analyses we focused exclusively on the transformer architecture’s residual stream\cite{elhage2021mathematical,ameisen2025circuit}. Let the model have $L$ layers and residual dimension $d$. For an input sequence of tokens $x_{1:T}$, we denote the residual activation at layer $\ell$ and token $t$ by $\mathbf{h}_{\ell,t} \in \mathbb{R}^d$. We registered forward hooks on the residual stream at each layer and ran a standard forward pass.

For each input, we represented its internal state by the residual activation at the \emph{last} token:
\begin{equation}
  \tilde{\mathbf{h}}^{(i)}_{\ell} = \mathbf{h}_{\ell,T^{(i)}},
\end{equation}
where $T^{(i)}$ is the index of the final token of input $i$. Thus, for every input $i$ and layer $\ell$, we obtain a single $d$-dimensional activation vector $\tilde{\mathbf{h}}^{(i)}_{\ell}$. To account for non-determinism in the model’s responses, we computed this vector for 10 repeated forward passes of the same input and averaged the resulting activations across repetitions.

Concept vectors (i.e., the representations of the internal states like threat or hostility) were constructed as difference-of-means directions. For a given contrast (e.g., symbolic threat vs.\ control; hostile vs.\ non-hostile), let $\mathcal{A}$ and $\mathcal{B}$ index inputs in the two conditions. At layer $\ell$, the raw direction is
\begin{equation}
  \mathbf{v}^{\text{(raw)}}_{\ell} =
  \frac{1}{|\mathcal{A}|} \sum_{i \in \mathcal{A}} \tilde{\mathbf{h}}^{(i)}_{\ell}
  \;-\;
  \frac{1}{|\mathcal{B}|} \sum_{j \in \mathcal{B}} \tilde{\mathbf{h}}^{(j)}_{\ell},
\end{equation}
and the normalized concept vector is
\begin{equation}
  \mathbf{v}_{\ell} = \frac{\mathbf{v}^{\text{(raw)}}_{\ell}}{\|\mathbf{v}^{\text{(raw)}}_{\ell}\|_2}.
\end{equation}
We applied this procedure to construct (i) realistic-threat and symbolic-threat vectors (threat vs.\ control vignettes for each type), (ii) a symbolic-versus-realistic contrast vector (symbolic-threat vs.\ realistic-threat vignettes), and (iii) a hostility vector (hostile vs.\ non-hostile behaviors).

For the hostility vector used in steering, we applied the same difference-of-means recipe, but grounded it in \emph{generated} behavior rather than in read vignettes. To target an activation pattern specifically tied to hostile intergroup behavior, we prompted the model with the intergroup situations from above and instructed it to respond either \emph{hostilely} (e.g., violent, hateful, intimidating, disruptive) or \emph{non-hostilely} (calm, cooperative), sampling multiple outputs per (scenario, instruction) pair. We used GPT-4o ratings to filter out cases where the output did not match the intended condition (e.g., ratings below the scale midpoint under hostile instructions; evaluation prompt in SI Section~\ref{sm:hostile_detection_prompt_steering}). From the remaining samples, we took the last-token residual activations during generation and defined the hostility direction as the mean activation for behaviors rated as hostile minus the mean activation for behaviors rated as non-hostile, yielding a behaviorally grounded hostility vector.

\subsubsection{Projection analyses}

Projection scores measure the scalar projection of an input's activation onto a concept vector, quantifying the degree to which the input's internal representation aligns with the target concept direction. For an input $i$, layer $\ell$, and normalized concept vector $\mathbf{v}_{\ell}$, we define the projection as
\begin{equation}
  s^{(i)}_{\ell} = \mathbf{v}_{\ell}^{\top} \tilde{\mathbf{h}}^{(i)}_{\ell},
\end{equation}
which is the signed component of the last-token activation $\tilde{\mathbf{h}}^{(i)}_{\ell}$ along $\mathbf{v}_{\ell}$ because $\|\mathbf{v}_{\ell}\|_2 = 1$.

We used these scores in three ways:
\begin{enumerate}
  \item \textbf{Validating threat vectors.} We projected held-out vignettes onto their corresponding threat vectors and compared projection distributions across threat, no-threat, and unrelated control vignettes to test whether the threat vectors specifically capture the intended threat type.
  \item \textbf{Validating the experimental manipulations.} We projected the belief statements from each experimental cell (no threat, symbolic-only, realistic-only, combined) onto the realistic- and symbolic-threat vectors to test whether the manipulations occupy the intended threat states in the activation space.
  \item \textbf{Linking threats to hostility.} We projected belief statements onto the hostility vector to assess whether threat manipulations move internal states toward the hostility direction.
\end{enumerate}

For each contrast (e.g., realistic-only vs.\ no-threat, symbolic-only vs.\ no-threat), we examined layer-wise mean projection differences and their bootstrapped confidence intervals, and quantified distributional separability using the 1D Wasserstein distance $D_W$ between projection samples. We report results for the layers with maximal separation for each contrast; full layer-wise curves are provided in the SI Section~\ref{sm:threat_representation_results}.

\subsubsection{Steering experiments}

Steering experiments test whether moving internal activations along a concept direction causally changes behavior. For a given scenario prompt, we generated responses under three conditions:
\begin{enumerate}
  \item \emph{No steering}: standard decoding with unmodified residual activations.
  \item \emph{Positive steering}: residual activations shifted in the direction of the hostility vector.
  \item \emph{Negative steering}: residual activations shifted opposite to the hostility vector.
\end{enumerate}

Let $\mathcal{L}_{\text{steer}}$ denote a small set of upper layers selected based on strong hostile vs.\ non-hostile separation on small development set. During autoregressive decoding, at each token $t$ and each $\ell \in \mathcal{L}_{\text{steer}}$, we modified the residual as
\begin{equation}
  \mathbf{h}'_{\ell,t} =
  \begin{cases}
    \mathbf{h}_{\ell,t} + \alpha \,\mathbf{v}^{\text{host}}_{\ell} & \text{(positive steering)} \\
    \mathbf{h}_{\ell,t} - \alpha \,\mathbf{v}^{\text{host}}_{\ell} & \text{(negative steering)} \\
    \mathbf{h}_{\ell,t} & \text{(no steering)},
  \end{cases}
\end{equation}
with steering strength $\alpha > 0$ (tuned on a small development set to avoid incoherent outputs). All other components of the model remained unchanged. Steering along threat-derived vectors (realistic vs.\ none, symbolic vs.\ none) used the same update rule with the corresponding threat vectors.

We evaluated steering on a held-out set of 20 intergroup scenarios (distinct from those used to extract the hostility vector), without any instructions or threat/hostility stimuli added to the prompts. For each scenario and steering condition, we generated 10 responses with fixed decoding hyper-parameters. GPT-4o was then used to then rate the degree of hostility on a 5-point scale (see SI Section~\ref{sm:hostile_detection_prompt_steering}). We compared the mean hostility rating across steering conditions, using two-tailed independent-sample $t$-tests (positive vs.\ negative steering, positive vs.\ no steering). Steering effects for threat-derived vectors were analyzed analogously and reported as exploratory.

\subsection{Simulation framework}
We built on the generative-agent architecture introduced by Park et al. \cite{park2023generative}, which implements autonomous language-model agents inhabiting a persistent, interactive virtual environment. Each agent maintains a memory stream of prior perceptions, internal reflections, and interactions, from which it generates open-ended plans and actions through natural-language reasoning. The environment includes spatial locations, objects, and other agents whose states are updated in real time. Each simulation instantiated twenty-five autonomous agents, following Park et al. \cite{park2023generative}. Time advanced in discrete steps (each steps reflecting 10 simulated seconds), during which agents updated their plans, moved through the environment, and initiated interactions when co-located with others. Agents could generate an unlimited number of actions over the course of a day, subject only to their self-generated plans and environmental affordances. On average, agents produced dozens of socially meaningful interactions per simulated hour—such as conversations, greetings, requests, or confrontations—interspersed with shorter micro-actions (e.g., moving, eating, sleeping) that maintained the flow of daily life.

We extended the original implementation to support controlled experimental manipulations. Specifically, the extension allows features of the environment or of agents, such as group membership, to be specified in natural language and made perceptible to other agents during interaction, while threat stimuli are delivered as private inputs to each agent’s own perception. Agents can therefore perceive stable attributes of others and shared environmental cues, but they only infer others’ perceived threat indirectly through communication. Agents’ memories, plans, and dialogues were updated continuously over three simulated days, producing rich, unscripted trajectories of decisions, movements, and conversations. In addition, we implemented a runtime probing module—analogous to ecological momentary assessment \citep[EMA;][]{shiffman2009ecological, kirchner2013ecological}—to periodically sample agents’ attitudinal states without affecting behavior (see details in Section~\ref{sec:agent_probing}).

Simulations were implemented using the \texttt{Mistral Small} language model (\texttt{mistralai/Mistral-Small-24B-Instruct-2501}\footnote{www.huggingface.com/mistralai/Mistral-Small-24B-Instruct-2501}), accessed via HuggingFace. To improve inference efficiency and reduce GPU memory requirements, we used a quantized version of the model (\texttt{matatonic/Mistral-Small-24B-Instruct-2501-6.5bpw-h8-exl2}\footnote{www.huggingface.com/matatonic/Mistral-Small-24B-Instruct-2501-6.5bpw-h8-exl2}) executed with the \texttt{ExLlamaV2} (v0.2.3) inference engine\footnote{\url{https://github.com/turboderp-org/exllamav2}}. Quantization represents model weights with reduced numerical precision—here, an average of 6.5~bits per weight—while maintaining comparable performance to the full-precision model. This approach substantially decreases memory usage (approximately~60\% reduction in VRAM) and accelerates inference with minimal degradation in generative quality. The applied quantization method follows the principles of GPTQ \citep{frantar2022gptq}, which performs accurate post-training quantization for large generative transformers. Quantized models have been shown to yield large computational and memory gains with negligible impact on output quality across a range of architectures \citep{lang2024comprehensive}.

All simulations were run on an on-demand computing cluster equipped with 40 NVIDIA RTX 4090 GPUs (24 GB VRAM each). The main simulation set was executed in parallel across all GPUs over 2.5 days (approximately 2,400 GPU hours), while the structural condition analyses required 7.5 days (approximately 7,200 GPU hours).

The choice of the Mistral architecture was motivated by its relatively low alignment and instruction-guarding constraints compared to larger instruction-tuned models such as GPT-4o or Claude 3.7. Preliminary testing with those systems led to full suppression of hostility or refusals to produce any negative intergroup behaviors. Mistral provided a more neutral generative prior, enabling the expression of both cooperative and antagonistic (including hateful) responses essential for modeling realistic social conflict dynamics.

All code for simulation orchestration, quantized inference, and data logging was implemented in Python (v3.10). 

\subsection{Experimental design and manipulations}
For the first set of simulations, we implemented a $2\times2$ factorial design crossing \emph{realistic threat} (strong vs.\ none) and \emph{symbolic threat} (strong vs.\ none). Agents were assigned to one of two minimal groups using natural-language identity prompts (“\verb|<Name>| is a member of Group~A. There is another group, Group~B, which they are not part of.”). To account for stochasticity, each condition was run 10 times, for a total of 40 simulations. In each run, we used a different random seed for LLM generation and independently randomized agent-level variables (e.g., group membership). Because agents repeatedly decide where to go, what to do, and what to say, with each decision shaping subsequent choices and others’ reactions, these stochastic elements compound over time and produce distinct trajectories of interaction in each run. All analyses use data from all 10 runs per condition.

Threat manipulations were induced by injecting standardized statements into agents’ perceptions (i.e., what was used as context in agents decision-making prompts) and memories. Statements were adapted from established threat-scale items \citep{kachanoff2021measuring} (e.g., for realistic threat: “\verb|<Name>| feels that the physical safety of \verb|<Group 1>| members is threatened by \verb|<Group 2>|.”; for symbolic threat: “\verb|<Name>| feels that the values and traditions of \verb|<Group 1>| are threatened by \verb|<Group 2>|.”; see Supplementary Table~\ref{tab:threat_prompts}). These belief percepts were continuously embedded within agents’ perception and memory streams, ensuring that the intended threat information remained salient and accessible during planning, interaction, and reflection (i.e., continuously added to the agents' context window when planning, acting, conversing). In effect, the manipulation maintained a stable, high-intensity representation of the target threat type while other threat dimensions were held at baseline.

To examine structural moderators, we ran a second set of simulations that introduced (i) spatial segregation, implemented by assigning agents to groups based on $k$-means clustering of their home coordinates on the town map, forming geographically distinct clusters that minimized intergroup overlap and maximized mean distance between groups; and (ii) demographic asymmetry—unequal group sizes implemented by varying relative group size, assigning 80\% of agents to form the majority group and the remaining 20\% to form the minority group. 

These manipulations altered the structural configuration of the environment rather than agents’ internal beliefs, and were not reflected in any instructions or prompts. Agents were not told whether they belonged to a majority or minority group, or whether groups were segregated—they simply experienced these conditions through their environment. This yielded a fully crossed $2$ (symbolic threat) $\times$ $2$ (realistic threat) $\times$ $2$ (segregated–integrated) $\times$ $2$ (equal–unequal group size) design, comprising 160 additional simulations. The underlying agent architecture and environment were otherwise identical across all conditions.

Each simulation ran for three simulated days, allowing multi-day interaction and cumulative social experience to shape conflict dynamics, while maintaining computational feasibility\footnote{Extending the simulation by one virtual day would increase runtime and GPU cost by 25–30\%: roughly 600 additional GPU hours for the main design and 1,800 GPU hours for structural factors analysis.}.

Each environment contained the same twenty-five agents from Park et al. \cite{park2023generative}, whose personas represent realistic variation in age (19–68), gender (44\% female), occupation (e.g., artist, shopkeeper, student, engineer), personality, and social life to emulate plausible social heterogeneity in everyday settings of a small community. For group assignment, agents were randomly split into two minimal groups (12 vs.\ 13 members), and in the structural simulations the majority–minority condition used a fixed 20 vs.\ 5 split. 

Finally, we report additional replication sets varying agents’ moral-value profiles and non-minimal group paradigms (pilot) in the SI (Sets 0 and 3; SI Section~\ref{subsec:replications}).

\subsection{Agent probing and attitudinal measures}
\label{sec:agent_probing}
During runtime, agents were periodically probed via separate natural-language queries—analogous to ecological momentary assessment (EMA) in human research. Probes were administered in parallel to agents’ ongoing decision and interaction processes by temporarily copying their current state (i.e., the information and memory at the time of action or conversation) and eliciting scale responses from this duplicate context. This ensured that the probes captured attitudinal states contemporaneous with decision-making while leaving the primary simulation trajectory unaffected. These probes adapted validated items from social–psychological scales measuring intergroup trust \citep{yamagishi1994trust}, collaboration \citep{caprara2005new}, dehumanization \citep{kteily2015ascent}, and ingroup identification (adapted from group identity, identity fusion, and group commitment scales; \cite{leach2008group, ellemers1997sticking, mael1992alumni, cameron2004three}). Items were phrased in first-person form (e.g., “I consider \verb|<Group 1>| members to be honest and reliable”) and scored using the model’s numerical responses to 7-point Likert-type anchors (1-totally disagree, 4-neutral, 7-totally agree). Probing occurred continuously throughout the simulation—following new actions or conversations (probing N = 46,240)—yielding dynamic, time-resolved measures of trust, cooperation, ingroup bias, identification, and dehumanization attitudes. Full item lists, and reliability statistics are provided in the SI (Section~\ref{sec:attitudes_methods}).

\subsection{Data logging and derived variables}
\label{sec:derived_variables}
Across the base and structural simulation sets, the agents produced a rich corpus of social behavior. In the baseline forty simulations (without structural manipulations), 527{,}387 actions were recorded, including 473{,}295 agent–agent interactions (238{,}463 intergroup) and more than 20{,}000 conversations over three simulated days. The extended structural set (160 simulations) generated over two million actions, including 986{,}401 agent–agent interactions (226{,}592 intergroup), and approximately 90{,}000 conversations ($\approx$65–75 million tokens). Statistical models were fit to data aggregated at the hourly level, as individual actions occur at highly variable temporal scales and numerous intermediate microactions (e.g., sleeping, moving, or eating) separate socially meaningful events such as conflict or contact, rendering moment-to-moment analyses unstable. This aggregation yielded between 15{,}000 and 75{,}000 samples per model, depending on whether actions, conversations, or attitudes were modeled, which differ in event frequency and sampling resolution.

All agent actions, plans, and conversations were logged with timestamps, acting and target agent IDs, group memberships, and experimental conditions. Conversations were analyzed using pretrained classifiers for hateful language, moral language, and sentiment. Specifically, we applied the \texttt{Elron/deberta-v3-large-sentiment}\footnote{https://huggingface.co/Elron/deberta-v3-large-emotion} model for sentiment polarity (F1: 0.74), \texttt{Elron/deberta-v3-large-hate}\footnote{https://huggingface.co/Elron/deberta-v3-large-hate} for hate speech detection (F1: 0.61), and a \texttt{roberta-base}\citep{liu2019roberta} model fine-tuned on the Moral Foundations Twitter Corpus \citep[MFTC;][]{hoover2020moral} to distinguish binding versus individualizing moral language (F1: 0.76). For non-linguistic behavioral data, a separate large language model (\texttt{Mistral-Large}\footnote{https://huggingface.co/mistralai/Mistral-Large-Instruct-2407}) classified whether each logged action was hostile (hateful, violent, aggressive) toward outgroup members (see SI Section~\ref{sm:hostile_detection_prompt}).

The final dataset is event-level, with one row per logged agent action (and, when applicable, per conversation). Each row contains the timestamp; initiator and target IDs (and their group memberships); the run’s condition/structural settings; a brief action log; and a hostility flag. Conversation rows additionally include derived linguistic features (sentiment, hate, moral language). From initiator/target groups we derive intergroup status and non-hostile intergroup contact (intergroup with hostility = false), and attitudinal probe responses are recorded at their sampling times.

\subsection{Analytical strategy}
Analyses focused on effects of threat type and structural variables on hostile action rates, language content (hateful language, moral language, sentiment), and attitudes over time. The primary specification was a mixed-effects negative binomial regression predicting the hourly rate of hostile actions from realistic and symbolic threat perception, their interaction, non-hostile intergroup contact rate in the previous hour, the hostile action rate in the previous hour (autoregression), a log offset for total actions, and time, with random intercepts for agents and simulation runs (nested within condition). All models were estimated in \texttt{R} (v4.3) using \texttt{glmmTMB}. Analogously, additional mixed-effects models with the same fixed-effect structure and random intercepts for agents and runs were fit for language hatefulness, moralization, sentiment, and attitudinal outcomes. System-level analyses aggregated data across all agents within each virtual town and run to characterize collective patterns in behavior, language, and attitudes. Full model results are provided in the Supplementary Information. All code files are available in the project repository \url{https://osf.io/5ac3d}.

To assess whether attitudes or hateful language mediate the effects of perceived threat on hostile actions, we estimated Bayesian multilevel negative binomial mediation models. Ingroup bias (or hateful language) at time $t-1$ was treated as the mediator linking symbolic and realistic threat perception at time $t-2$ to hostile actions at time $t$, controlling for prior ingroup bias (or hateful language), intergroup contact rate, and hostile action rate. Models included random intercepts for agents and simulation runs and were estimated on $N = 23{,}355$ observations. We used weakly informative priors on all parameters: Student-$t$ priors on intercepts, Normal priors on regression coefficients, and Exponential priors on variance and shape parameters. Models were fit in \texttt{brms} (via \texttt{cmdstanr}) using 4 chains with 3{,}000 iterations each (1{,}800 post–warm-up), a 1{,}200-iteration warm-up, and a target acceptance rate of 0.99. All Bayesian models converged satisfactorily ($\hat{R} < 1.01$, no divergent transitions, and effective sample sizes $> 1{,}000$ for all parameters). Full Bayesian model outputs are provided in the SI (Supplementary Table~\ref{tab:S_lang_mediation},\ref{tab:S_bias_mediation}).

\section{Reproducibility, transparency, and robustness}\label{sec:reproducibility}
Given the inherently stochastic nature of LLMs and the sequential, interdependent design of the present simulations, reproducibility and transparency are particularly important. Unlike typical one-off LLM uses (classification or single responses), our framework generates chains of decisions in which each agent’s plan, action, and dialogue shape later prompts and perceptions, amplifying stochastic variation over time. We therefore provide extensive details on data-generation procedures, model settings, validation, and robustness checks following Abdurahman et al. \cite{abdurahman2025primer}; see SI Section~\ref{sm:reproducibility}.

\section*{Data Availability}

All simulation data used in this study are available in the project repository at \url{https://osf.io/5ac3d}. No human data or external datasets were used.

\section*{Code Availability}

All code for data generation, preprocessing, analysis, and simulations is available in the project repository at \url{https://osf.io/5ac3d}.

\section*{Acknowledgements}
This work was supported in part by the Air Force Office of Scientific Research (AFOSR) under Grant No. A9550-23-1-0463. The authors gratefully acknowledge this support.

\section*{Author Contributions}

Using the CRediT taxonomy, contributions were as follows:

SA: Conceptualization, Methodology, Software, Formal analysis, Investigation, Writing – original draft, Writing – review \& editing, Visualization.

FKM: Conceptualization, Writing – original draft.

CY: Conceptualization, Methodology, Software, Formal analysis, Writing – review \& editing, Visualization.

NK: Conceptualization, Writing – review \& editing.

MD: Conceptualization, Writing – review \& editing, Supervision, Funding acquisition.

\section*{Competing interests}
The authors declare no competing interests.

\bibliography{sn-bibliography}

\clearpage
\begin{appendices}
\section*{Extended Data}

\begin{figure}[!htbp]
    \centering

    \begin{subfigure}{0.48\textwidth}
        \centering
        \includegraphics[width=\textwidth]{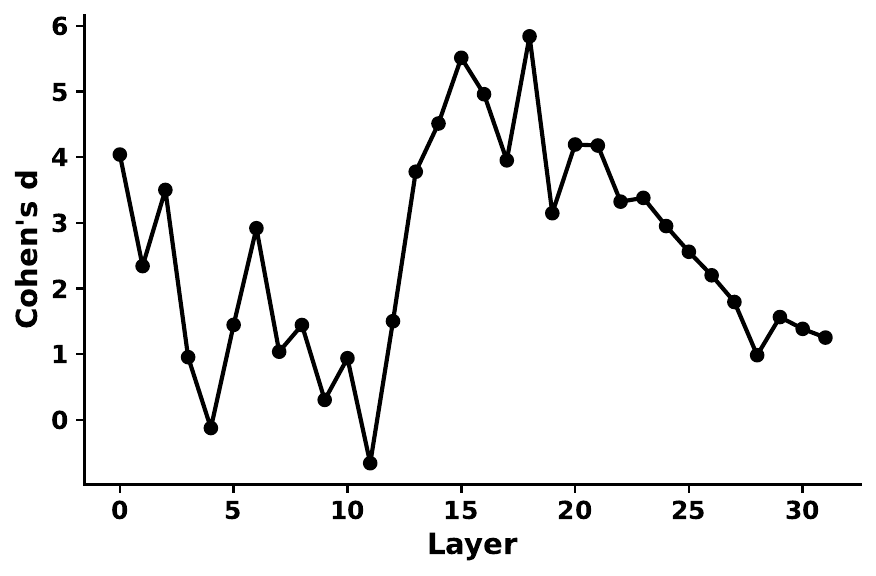}
        \caption{Symbolic vs control: Cohen's $d$.}
        \label{fig:symbolic_cohens_d_layers}
    \end{subfigure}
    \hfill
    \begin{subfigure}{0.48\textwidth}
        \centering
        \includegraphics[width=\textwidth]{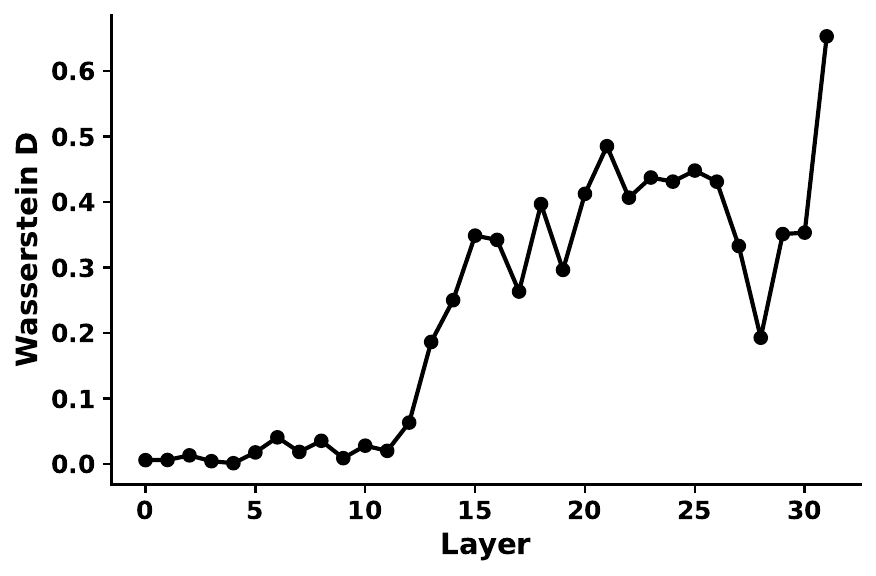}
        \caption{Symbolic vs control: Wasserstein distance.}
        \label{fig:symbolic_wasserstein_layers}
    \end{subfigure}

    \vspace{0.5em}

    \begin{subfigure}{0.48\textwidth}
        \centering
        \includegraphics[width=\textwidth]{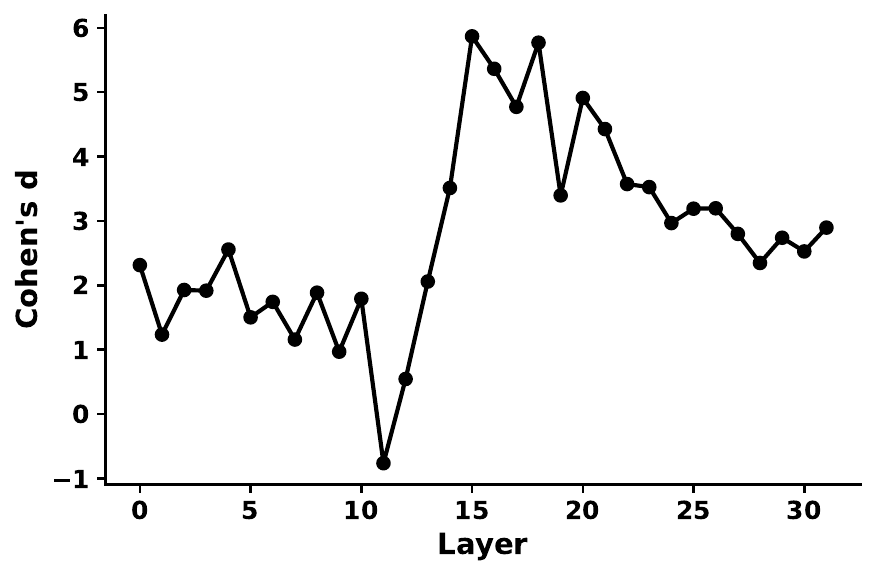}
        \caption{Realistic vs control: Cohen's $d$.}
        \label{fig:realistic_cohens_d_layers}
    \end{subfigure}
    \hfill
    \begin{subfigure}{0.48\textwidth}
        \centering
        \includegraphics[width=\textwidth]{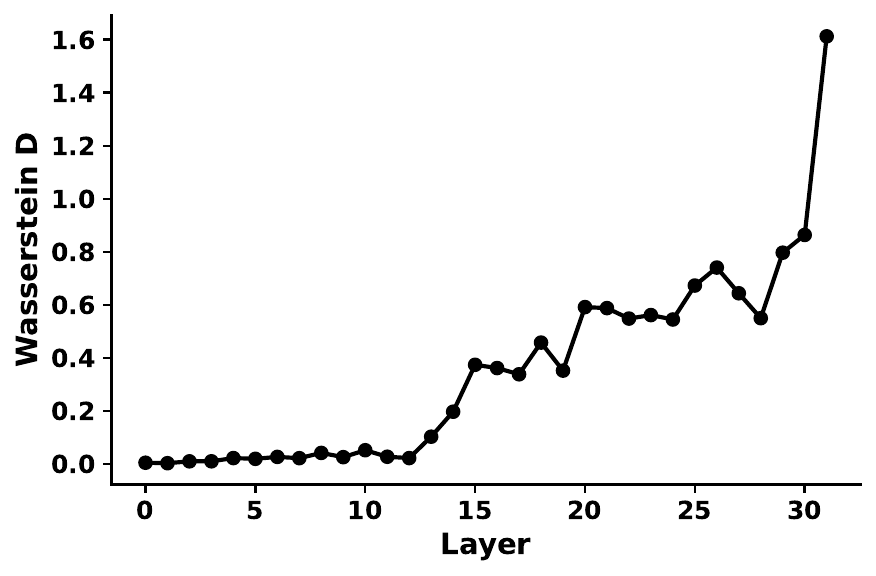}
        \caption{Realistic vs control: Wasserstein distance.}
        \label{fig:realistic_wasserstein_layers}
    \end{subfigure}

    \vspace{0.5em}

    \begin{subfigure}{0.48\textwidth}
        \centering
        \includegraphics[width=\textwidth]{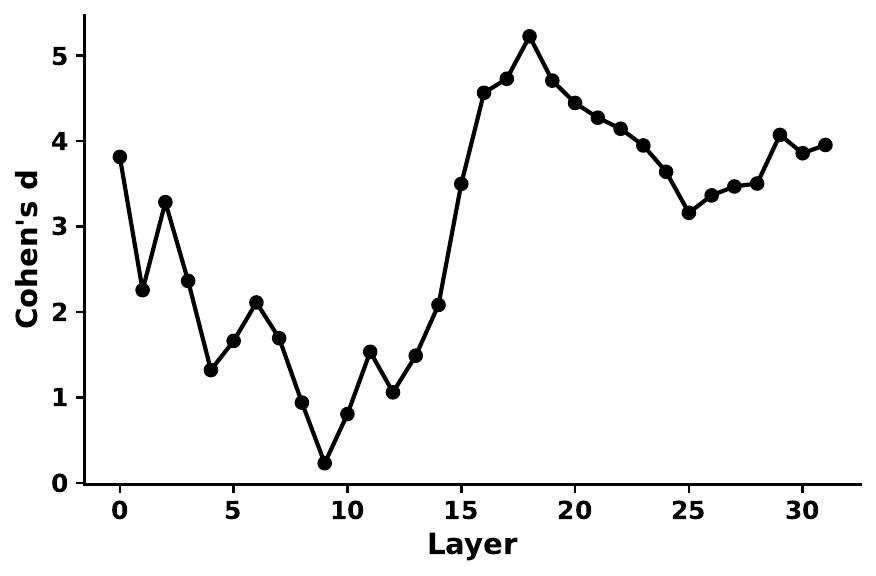}
        \caption{Symbolic vs realistic: Cohen's $d$.}
        \label{fig:sym_v_real_cohens_d_layers}
    \end{subfigure}
    \hfill
    \begin{subfigure}{0.48\textwidth}
        \centering
        \includegraphics[width=\textwidth]{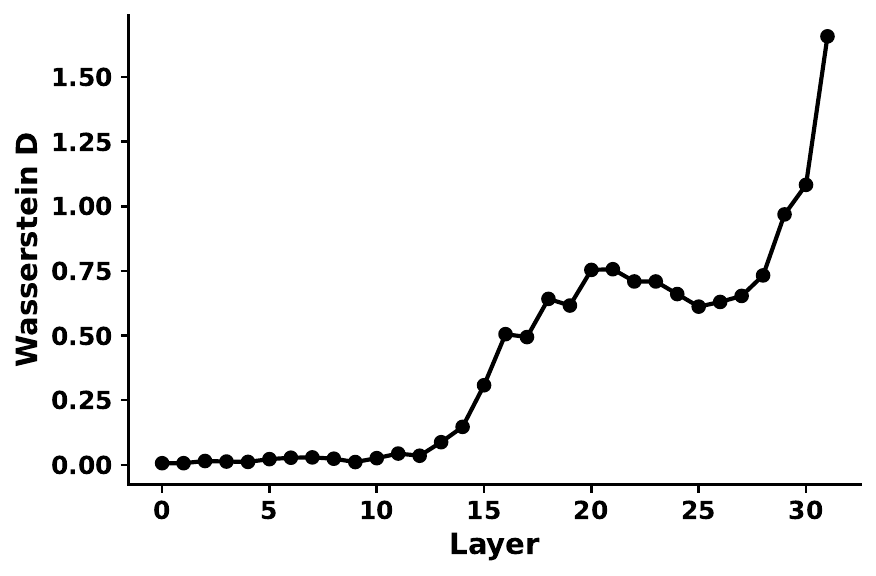}
        \caption{Symbolic vs realistic: Wasserstein distance.}
        \label{fig:sym_v_real_wasserstein_layers}
    \end{subfigure}

    \caption{Layer-wise separability of symbolic and threat realistic
    threat projections. Panels show Cohen's $d$ and Wasserstein distance across layers for symbolic-threat vs control vignettes (top panels), realistic vs control vignettes (middle panels) and for symbolic-threat vs realistic-threat vignettes (bottom panels).}
    \label{fig:separability_layers}
\end{figure}

\begin{figure}[!htbp]
    \centering

    \begin{subfigure}{0.48\textwidth}
        \centering
        \includegraphics[width=\textwidth]{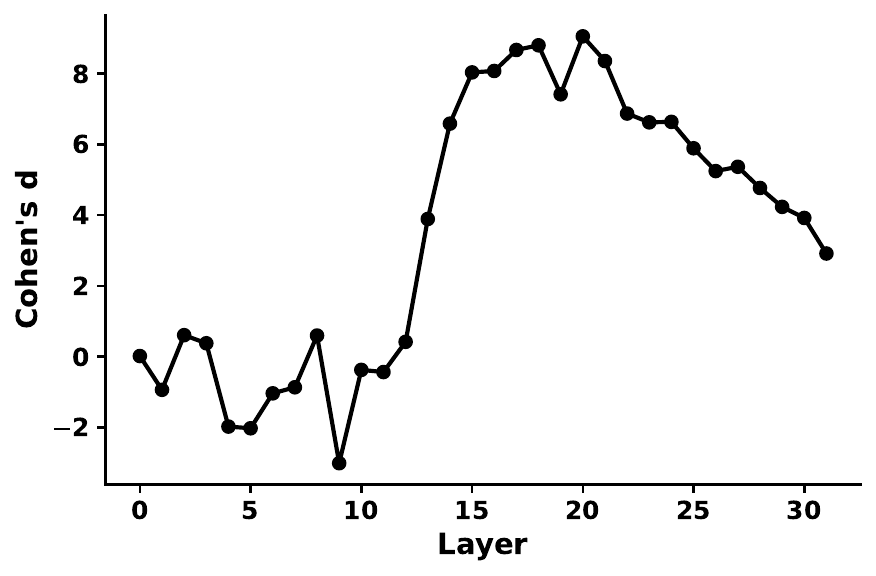}
        \caption{Symbolic vs No-Threat: Cohen's $d$.}
        \label{fig:symbolic_cohens_d_layers_treatment}
    \end{subfigure}
    \hfill
    \begin{subfigure}{0.48\textwidth}
        \centering
        \includegraphics[width=\textwidth]{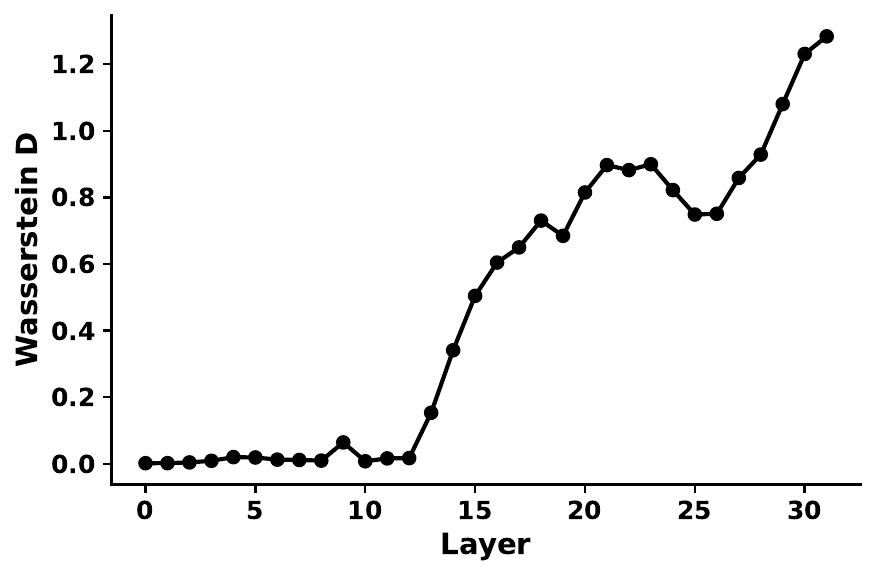}
        \caption{Symbolic vs No-Threat: Wasserstein distance.}
        \label{fig:symbolic_wasserstein_layers_treatment}
    \end{subfigure}

    \vspace{0.5em}

    \begin{subfigure}{0.48\textwidth}
        \centering
        \includegraphics[width=\textwidth]{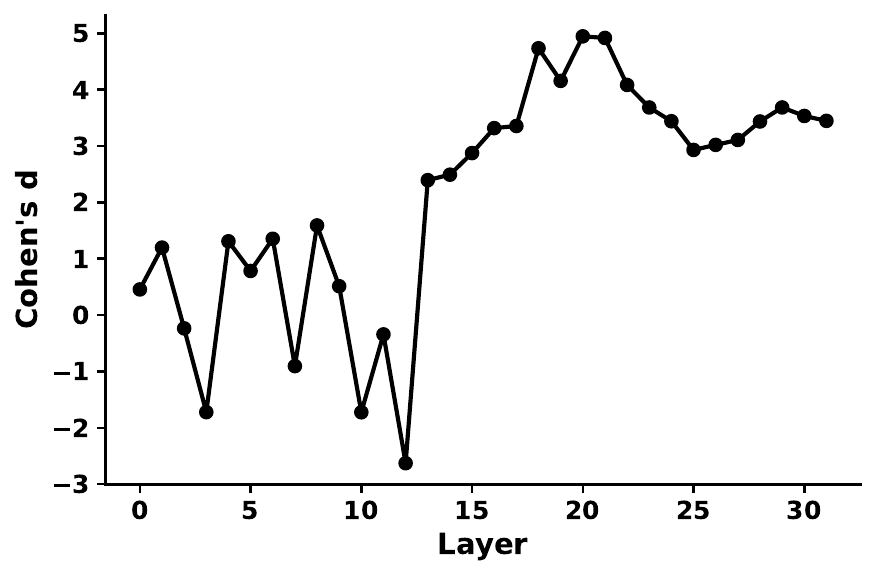}
        \caption{Realistic vs No-Threat: Cohen's $d$.}
        \label{fig:realistic_cohens_d_layers_treatment}
    \end{subfigure}
    \hfill
    \begin{subfigure}{0.48\textwidth}
        \centering
        \includegraphics[width=\textwidth]{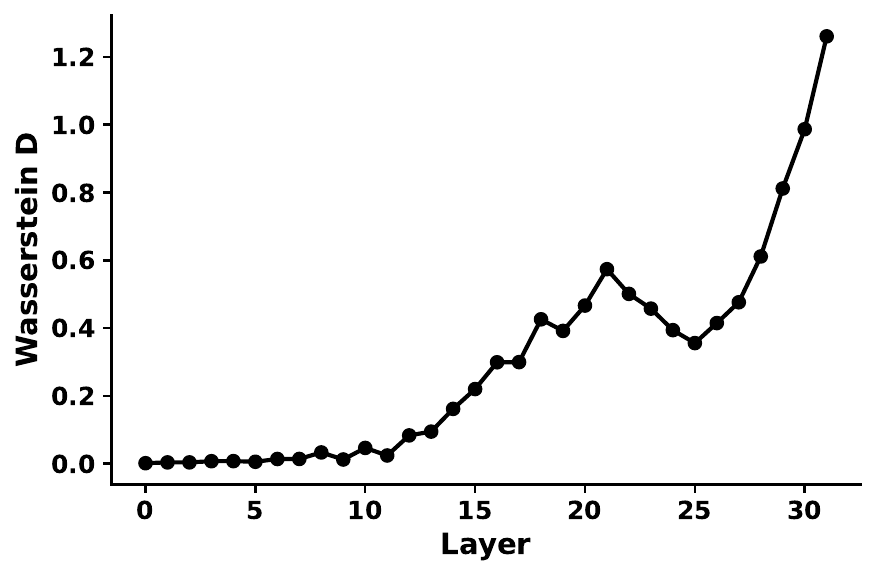}
        \caption{Realistic vs No-Threat: Wasserstein distance.}
        \label{fig:realistic_wasserstein_layers_treatment}
    \end{subfigure}

    \vspace{0.5em}

    \begin{subfigure}{0.48\textwidth}
        \centering
        \includegraphics[width=\textwidth]{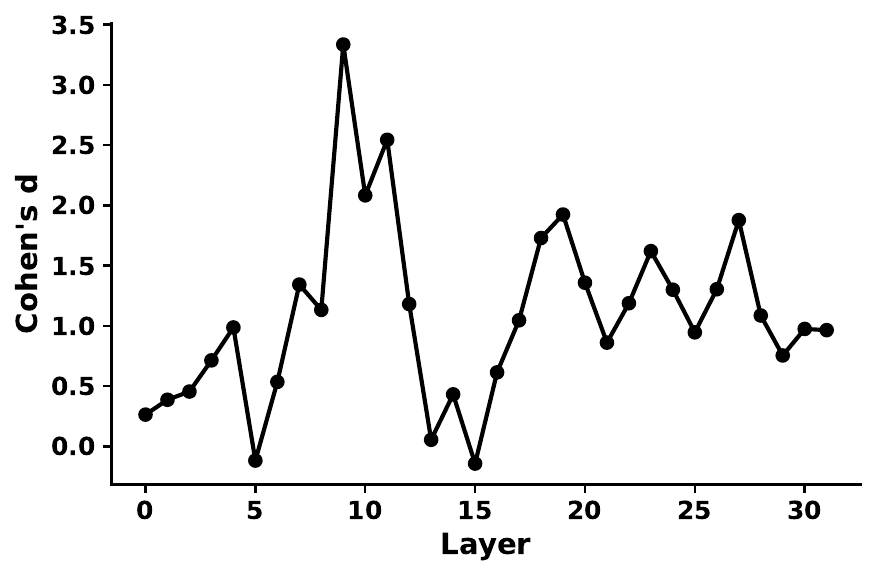}
        \caption{Symbolic vs Realistic: Cohen's $d$.}
        \label{fig:sym_v_real_cohens_d_layers_treatment}
    \end{subfigure}
    \hfill
    \begin{subfigure}{0.48\textwidth}
        \centering
        \includegraphics[width=\textwidth]{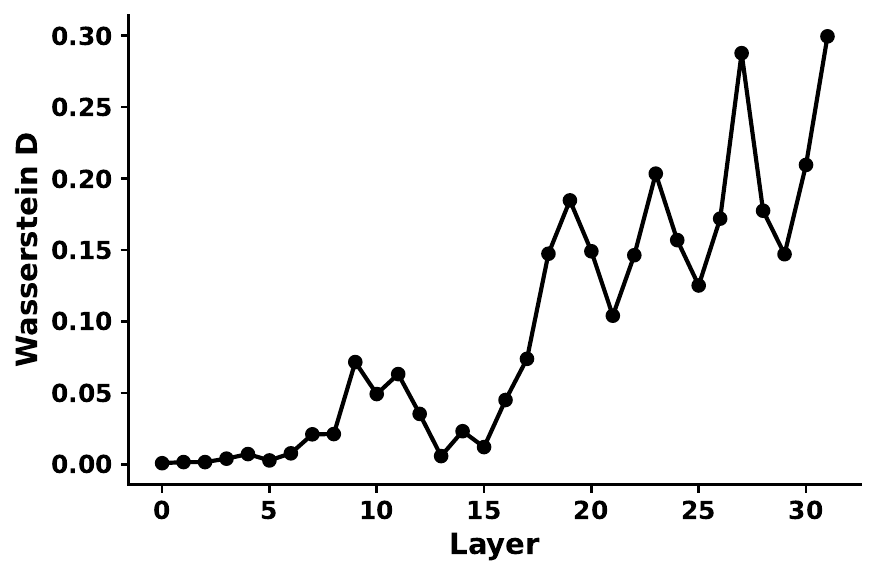}
        \caption{Symbolic vs Realistic: Wasserstein distance.}
        \label{fig:sym_v_real_wasserstein_layers_treatment}
    \end{subfigure}

    \caption{Layer-wise separability of experimental condition based on their projection on the threat-state vectors. Panels show Cohen's $d$ and Wasserstein distance across layers for symbolic-only vs no-threat condition (top panels), realistic-only vs no-threat condition (middle panels) and for symbolic-only vs realistic-only condition (bottom panels).}
    \label{fig:ED_threat_mean_diff_layers_treatment}
\end{figure}

\begin{table}[ht]
\centering
\caption{Predicting system-level hourly hostile action rate ($M1_{sys}$; $N=2{,}050$).}
\label{tab:8}
\begin{tabular}{lrrr}
\toprule
Predictor & $\beta$ & SE & $p$ \\
\midrule
Intercept & $-6.96$ & $0.06$ & $<.001$ \\
Hostile action rate (lag) & $0.16$ & $0.04$ & $<.001$ \\
Intergroup contact rate (lag) & $-0.30$ & $0.07$ & $<.001$ \\
Symbolic threat & $0.17$ & $0.05$ & $0.002$ \\
Realistic threat & $0.33$ & $0.05$ & $<.001$ \\
Time & $-0.25$ & $0.05$ & $<.001$ \\
Symbolic $\times$ Realistic threat & $-0.14$ & $0.05$ & $0.007$ \\
\bottomrule
\end{tabular}
\end{table}

\begin{table}[ht]
\centering
\caption{Predicting hourly hateful language rate (M2a; $N=15{,}684$).}
\label{tab:2}
\begin{tabular}{lrrr}
\toprule
Predictor & $\beta$ & SE & $p$ \\
\midrule
Intercept & $-7.27$ & $0.27$ & $<.001$ \\
Hateful language rate (lag) & $0.03$ & $0.02$ & $0.107$ \\
Intergroup contact rate (lag) & $0.06$ & $0.05$ & $0.244$ \\
Symbolic threat & $0.46$ & $0.12$ & $<.001$ \\
Realistic threat & $0.98$ & $0.12$ & $<.001$ \\
Time & $-0.01$ & $0.05$ & $0.822$ \\
Symbolic $\times$ Realistic threat & $-0.28$ & $0.11$ & $0.015$ \\
\bottomrule
\end{tabular}
\end{table}

\begin{table}[ht]
\centering
\caption{Predicting ingroup bias attitudes (M3a; $N=46{,}240$).}
\label{tab:6_7}
\small
\begin{tabular*}{0.8\textwidth}{@{\extracolsep{\fill}}lrrr}
\toprule
Predictor & $\beta$ & SE & $p$ \\
\midrule
Intercept & $0.01$ & $0.06$ & $.88$ \\
Group Bias (lag) & $0.12$ & $0.01$ & $<.001$ \\
Symbolic threat & $0.39$ & $0.05$ & $<.001$ \\
Realistic threat & $0.28$ & $0.05$ & $<.001$ \\
Time & $-0.01$ & $0.00$ & $.003$ \\
Symbolic $\times$ Realistic threat & $0.07$ & $0.04$ & $.083$ \\
\bottomrule
\end{tabular*}
\end{table}

\begin{table}[ht]
\centering
\caption{Predicting hourly hostile action rate including structural manipulations ($M1_{structure}$; $N=74{,}487$).}
\label{tab:5a}
\begin{tabular}{lrrr}
\toprule
Predictor & $\beta$ & SE & $p$ \\
\midrule
Intercept & $-9.83$ & $0.49$ & $<.001$ \\
Intergroup contact rate (lag) & $-0.58$ & $0.07$ & $<.001$ \\
Hostile action rate (lag) & $0.05$ & $0.01$ & $<.001$ \\
Symbolic threat & $0.13$ & $0.05$ & $0.011$ \\
Realistic threat & $0.28$ & $0.05$ & $<.001$ \\
Time & $-0.22$ & $0.04$ & $<.001$ \\
Segregation & $-0.40$ & $0.28$ & $0.152$ \\
Group Size & $0.08$ & $0.28$ & $0.788$ \\
Symbolic $\times$ Realistic & $-0.09$ & $0.04$ & $0.041$ \\
Segregation $\times$ Group Size & $0.42$ & $0.28$ & $0.135$ \\
\bottomrule
\end{tabular}
\end{table}
    
\section{Additional Materials and Stimuli}\label{sec:examples}

\subsection{Threat perception prompts}\label{sec:threat_prompts}
To induce the intended threat perceptions, we injected natural-language statements into agents’ memory and perception streams. These prompts operationalized \emph{realistic} and \emph{symbolic} threat manipulations within a $2\times2$ factorial design (realistic: strong vs.\ none; symbolic: strong vs.\ none). Each condition combined reinforcing statements for the target threat type with suppressing statements for the non-target type, ensuring clear differentiation across experimental cells. Table~\ref{tab:threat_prompts} lists the full set of statements used for each threat condition.

\begin{table}[h!]
\centering
\caption{Belief statements used to induce or suppress perceived threat.}
\label{tab:threat_prompts}
\renewcommand{\arraystretch}{1.3}
\begin{tabular}{p{0.20\linewidth} p{0.38\linewidth} p{0.38\linewidth}}
\toprule
\textbf{Threat type} & \textbf{Induce condition} & \textbf{Suppress condition} \\
\midrule
\multirow{2}{*}{\textbf{Symbolic threat}} 
& \verb|<Name>| strongly feels that their values are under threat by \verb|<Group 2>|. 
& \verb|<Name>| does not feel that their values are under threat by \verb|<Group 2>|. \\
& \verb|<Name>| strongly feels that their traditions are under threat by \verb|<Group 2>|. 
& \verb|<Name>| does not feel that their traditions are under threat by \verb|<Group 2>|. \\
\addlinespace
\multirow{2}{*}{\textbf{Realistic threat}} 
& \verb|<Name>| strongly feels economically threatened by \verb|<Group 2>|. 
& \verb|<Name>| does not feel economically threatened by \verb|<Group 2>|. \\
& \verb|<Name>| strongly feels physically threatened by \verb|<Group 2>|. 
& \verb|<Name>| does not feel physically threatened by \verb|<Group 2>|. \\
\bottomrule
\end{tabular}
\end{table}

\subsection{Scales for Probing Agents}\label{sec:attitudes_methods}

\begin{table}[!htpb]
\caption{Group Identification Items}\label{tab:groupid}
\begin{tabular*}{\textwidth}{@{\extracolsep\fill}lp{0.85\textwidth}}
\toprule
\textbf{\#} & \textbf{Item} \\
\midrule
1 & I feel a strong bond with \verb|<Group 1>|. \\
2 & Being a \verb|<Group 1>| is an important reflection of who I am. \\
3 & I strongly identify with \verb|<Group 1>|’s values. \\
4 & I feel committed to \verb|<Group 1>|. \\
5 & I am willing to make personal sacrifices to support \verb|<Group 1>|. \\
\botrule
\end{tabular*}
\footnotetext{Note: Adapted from multiple group identity scales \citep{leach2008group, ellemers1997sticking, mael1992alumni, cameron2004three}.}
\end{table}

\begin{table}[!htpb]
\caption{Group Bias Items}\label{tab:bias}
\begin{tabular*}{\textwidth}{@{\extracolsep\fill}lp{0.85\textwidth}}
\toprule
\textbf{\#} & \textbf{Item} \\
\midrule
1 & Members of my group (\verb|<Group 1>|) are more moral than members of the other group (\verb|<Group 2>|). \\
2 & Members of my group (\verb|<Group 1>|) are more trustworthy than members of the other group (\verb|<Group 2>|). \\
3 & Members of my group (\verb|<Group 1>|) are more competent than members of the other group (\verb|<Group 2>|). \\
4 & Members of my group (\verb|<Group 1>|) are more kind than members of the other group (\verb|<Group 2>|). \\
\botrule
\end{tabular*}
\footnotetext{Note: Items adapted from warmth–competence framework questionnaires \citep{cuddy2008warmth}.}
\end{table}

\begin{table}[!htpb]
\caption{Trust Items}\label{tab:trust}
\begin{tabular*}{\textwidth}{@{\extracolsep\fill}lp{0.85\textwidth}}
\toprule
\textbf{\#} & \textbf{Item} \\
\midrule
1 & I consider \verb|<Group 1>| members to be honest and reliable. \\
2 & I expect most \verb|<Group 1>| members to act in good faith. \\
3 & I consider \verb|<Group 2>| members to be honest and reliable. \\
4 & I expect most \verb|<Group 2>| members to act in good faith. \\
\botrule
\end{tabular*}
\footnotetext{Note: Adapted from the Trust and Commitment Scale \citep{yamagishi1994trust}.}
\end{table}

\begin{table}[!htpb]
\caption{Collaboration Items}\label{tab:collab}
\begin{tabular*}{\textwidth}{@{\extracolsep\fill}lp{0.85\textwidth}}
\toprule
\textbf{\#} & \textbf{Item} \\
\midrule
1 & I am willing to collaborate with \verb|<Group 1>| members to achieve shared goals. \\
2 & I am happy to share my resources with \verb|<Group 1>| members when they need help. \\
3 & I am willing to collaborate with \verb|<Group 2>| members to achieve shared goals. \\
4 & I am happy to share my resources with \verb|<Group 2>| members when they need help. \\
\botrule
\end{tabular*}
\footnotetext{Note: Adapted from the Prosocialness Scale \citep{caprara2005new}.}
\end{table}

\begin{table}[!htpb]
\caption{Dehumanization Items}\label{tab:dehumanization}
\begin{tabular*}{\textwidth}{@{\extracolsep\fill}lp{0.85\textwidth}}
\toprule
\textbf{\#} & \textbf{Item} \\
\midrule
1  & \verb|<Group 1>| members often seem primitive and uncivilized. \\
2 & \verb|<Group 1>| members often seem cold and mechanical. \\
3 & \verb|<Group 2>| members often seem primitive and uncivilized. \\
4 & \verb|<Group 2>| members often seem cold and mechanical. \\
\botrule
\end{tabular*}
\footnotetext{Note: Adapted from the Blatant Dehumanization Scale \citep{kteily2015ascent}.}
\end{table}

\clearpage
\section{Supplement to Main Analysis}\label{secA1}

\noindent
This Appendix reports supplementary and robustness analyses complementing the main text. Unless noted otherwise, count and rate outcomes (e.g., hostile action rate, hateful language rate, moralized language rate) were estimated using mixed-effects negative binomial regressions, and continuous outcomes (e.g., sentiment, attitudes) using Gaussian mixed-effects regressions with random intercepts for agent and simulation run. All models were fit on hourly data as specified in the main text.

\subsection{Zero-order correlations of main variables}
\label{sm:zero_order}
Table~\ref{tab:zero_order_corr} reports zero-order correlations among all key constructs, computed at the agent level by averaging each variable across the three simulated days. These descriptive associations provide an overview of how threat, hostile behavior, language use, attitudes, and intergroup contact co-occur across the simulations.

\begin{table}[!htpb]
\centering
\caption{Zero-order correlations among agent-level variables (averaged across the three-day simulation).}
\label{tab:zero_order_corr}
\small
\begin{tabularx}{0.95\textwidth}{l*{9}{>{\centering\arraybackslash}X}}
\toprule
 & \textbf{Real.} & \textbf{Symb.} & \textbf{Host.} & \textbf{Hate} & \textbf{Sent.} & \textbf{Bias} & \textbf{Cont.} & \textbf{Bind.} & \textbf{Indiv.} \\
\midrule
Real. threat      & 1.00 & 0.00 & 0.05 & 0.25 & -0.44 & 0.53 & 0.03 & 0.47 & 0.82 \\
Symb. threat      & 0.00 & 1.00 & 0.03 & 0.10 & -0.20 & 0.74 & 0.08 & 0.45 & 0.23 \\
Hostile rate      & 0.05 & 0.03 & 1.00 & 0.80 & -0.72 & 0.30 & 0.01 & 0.26 & 0.32 \\
Hate lang.        & 0.25 & 0.10 & 0.80 & 1.00 & -0.81 & 0.47 & 0.04 & 0.33 & 0.49 \\
Sentiment         & -0.44 & -0.20 & -0.72 & -0.81 & 1.00 & -0.65 & 0.02 & -0.45 & -0.62 \\
Group bias        & 0.53 & 0.74 & 0.30 & 0.47 & -0.65 & 1.00 & 0.05 & 0.59 & 0.67 \\
Contact           & 0.03 & 0.08 & 0.01 & 0.04 & 0.02 & 0.05 & 1.00 & 0.25 & 0.07 \\
Bind. moral       & 0.47 & 0.45 & 0.26 & 0.33 & -0.45 & 0.59 & 0.25 & 1.00 & 0.80 \\
Indiv. moral      & 0.82 & 0.23 & 0.32 & 0.49 & -0.62 & 0.67 & 0.07 & 0.80 & 1.00 \\
\bottomrule
\end{tabularx}
\end{table}

\subsection{Language dynamics}

\subsubsection{Hateful language}  
Model M2a examines whether threat perceptions and prior interaction history predict the rate of hateful language, using hourly conversation-level data.

\begin{table}[ht]
\centering
\caption{Predicting hourly hateful language rate (M2a; $N = 15{,}684$).}
\label{tab:S2}
\begin{tabular}{lrrr}
\toprule
Predictor & $\beta$ & SE & $p$ \\
\midrule
Intercept & $-7.27$ & $0.27$ & $<.001$ \\
Hateful language rate (lag) & $0.03$ & $0.02$ & $0.107$ \\
Intergroup contact rate (lag) & $0.06$ & $0.05$ & $0.244$ \\
Symbolic threat & $0.46$ & $0.12$ & $<.001$ \\
Realistic threat & $0.98$ & $0.12$ & $<.001$ \\
Time & $-0.01$ & $0.05$ & $0.822$ \\
Symbolic $\times$ Realistic threat & $-0.28$ & $0.11$ & $0.015$ \\
\bottomrule
\end{tabular}
\end{table}

\subsubsection{Sentiment}  
Model M2b predicts average sentiment within an hour using the same hourly conversation data ($N = 15{,}684$). 

\begin{table}[ht]
\centering
\caption{Predicting hourly mean sentiment (M2b).}
\label{tab:S3}
\begin{tabular}{lrrr}
\toprule
Predictor & $\beta$ & SE & $p$ \\
\midrule
Intercept & $1.70$ & $0.04$ & $<.001$ \\
Intergroup contact rate (lag) & $0.00$ & $0.00$ & $0.032$ \\
Symbolic threat & $-0.05$ & $0.01$ & $<.001$ \\
Realistic threat & $-0.10$ & $0.01$ & $<.001$ \\
Time & $0.02$ & $0.00$ & $<.001$ \\
Sentiment (lag) & $0.03$ & $0.00$ & $<.001$ \\
Symbolic $\times$ Realistic threat & $-0.01$ & $0.01$ & $0.080$ \\
\bottomrule
\end{tabular}
\end{table}

\subsubsection{Moral language}

Model M2c and Model M2d examine how threat shapes the moral content of conversations, distinguishing binding (loyalty, authority, purity) and individualizing (harm/care, fairness) moral language based on Moral Foundations Theory \citep{graham2013moral}. Both models predict the hourly rate of moralized language use using the same hourly conversation data ($N = 15{,}684$). 

\begin{table}[ht]
\centering
\caption{(Panel A). Predicting hourly rate of binding moral language (M2c).}
\label{tab:S4}
\begin{tabular}{lrrr}
\toprule
Predictor & $\beta$ & SE & $p$ \\
\midrule
Intercept & $-1.46$ & $0.05$ & $<.001$ \\
Binding (lag) & $0.08$ & $0.01$ & $<.001$ \\
Intergroup contact rate (lag) & $-0.02$ & $0.01$ & $.007$ \\
Symbolic threat & $0.31$ & $0.02$ & $<.001$ \\
Realistic threat & $0.33$ & $0.02$ & $<.001$ \\
Time & $0.03$ & $0.01$ & $<.001$ \\
Symbolic $\times$ Realistic threat & $-0.16$ & $0.02$ & $<.001$ \\
\bottomrule
\end{tabular}
\end{table}

\begin{table}[ht]
\centering
\caption{(Panel B). Predicting hourly rate of individualizing moral language (M2d).}
\label{tab:S5}
\begin{tabular}{lrrr}
\toprule
Predictor & $\beta$ & SE & $p$ \\
\midrule
Intercept & $-2.62$ & $0.08$ & $<.001$ \\
Individualizing (lag) & $0.07$ & $0.01$ & $<.001$ \\
Intergroup contact rate (lag) & $-0.01$ & $0.01$ & $.203$ \\
Symbolic threat & $0.08$ & $0.06$ & $.157$ \\
Realistic threat & $0.64$ & $0.06$ & $<.001$ \\
Time & $0.04$ & $0.01$ & $<.001$ \\
Symbolic $\times$ Realistic threat & $0.03$ & $0.06$ & $.614$ \\
\bottomrule
\end{tabular}
\end{table}

\subsubsection{Mediation of threat effects on hostile action rate through hateful language}
To assess whether hateful language mediates the effects of perceived threat on hostile actions, we estimated a Bayesian multilevel mediation model (\texttt{brms}). Hateful language rate at time $t-1$ was specified as the mediator linking symbolic and realistic threat perception (at $t-2$) to hostile actions at time $t$, controlling for prior hateful language, intergroup contact rate, and hostile action rate, with random intercepts for agents and simulation run ($N=23{,}355$). We used weakly informative priors on all parameters (Student-t priors on intercepts, Normal priors on regression coefficients, and exponential priors on variance and shape parameters).

Posterior estimates indicated small but credibly positive $a$ paths from both threat types to hateful language ($\hat\beta_{a,\text{realistic}} = 0.06$ [0.04, 0.07]; $\hat\beta_{a,\text{symbolic}} = 0.03$ [0.01, 0.04]). By contrast, the $b$ path from hateful language to hostile actions was slightly negative and practically small ($\hat\beta_{b} = -0.05$ [$-0.11$, 0.00]), indicating that higher levels of hateful language do not predict increases in subsequent hostile actions and may, if anything, be weakly attenuating. Consistent with this, the resulting indirect effects were extremely small, negative, and effectively centered on zero ($\hat\beta_{\text{indirect, realistic}} = -0.00$ [$-0.01$, 0.00]; $\hat\beta_{\text{indirect, symbolic}} = -0.00$ [0.00, 0.00]), providing no evidence that hateful language mediates the effects of threat on hostility.

In contrast, the direct effect of realistic threat on hostile actions remained robustly positive ($\hat\beta_{\text{direct, realistic}} = 0.26$ [0.10, 0.43]), while the direct effect of symbolic threat was positive but more uncertain ($\hat\beta_{\text{direct, symbolic}} = 0.13$ [$-0.04$, 0.30]). Overall, these results indicate that (1) hateful language does not escalate hostile behavior, (2) hateful language does not constitute a meaningful mediating pathway from threat to hostility, and (3) hostile actions are driven primarily by perceived threat, especially realistic threat, via a direct route rather than via hateful language.

\begin{table}[ht]
\centering
\caption{Bayesian mediation of threat effects on hostile actions via hateful language ($N=23{,}355$).}
\label{tab:S_lang_mediation}
\begin{tabular}{lrrr}
\toprule
Effect & Estimate & 95\% CrI & pd (\%) \\
\midrule
$a$ (Realistic threat $\rightarrow$ Hateful language) & $0.06$   & [0.04, 0.07]   & 100.00 \\
$a$ (Symbolic threat $\rightarrow$ Hateful language)  & $0.03$   & [0.01, 0.04]   &  99.36 \\
$b$ (Hateful language $\rightarrow$ Hostility)        & $-0.05$  & [$-0.11$, 0.00] &  98.09 \\
Indirect (Realistic threat $\rightarrow$ Hateful language $\rightarrow$ Hostility) & $-0.00$ & [$-0.01$, 0.00] & 98.09 \\
Indirect (Symbolic threat $\rightarrow$ Hateful language $\rightarrow$ Hostility)  & $-0.00$ & [0.00, 0.00]   & 97.45 \\
Direct (Realistic threat $\rightarrow$ Hostility)     & $0.26$   & [0.10, 0.43]   & 99.90 \\
Direct (Symbolic threat $\rightarrow$ Hostility)      & $0.13$   & [$-0.04$, 0.30] & 93.67 \\
\bottomrule
\end{tabular}
\caption*{All estimates are posterior medians with 95\% credible intervals (CrI) and posterior probability of direction (pd)}
\end{table}

\subsection{Attitudes}

\subsubsection{Ingroup bias attitudes}  
Model M3a predicts agents' average ingroup bias scores using the data from probing agents throughout the simulation ($N=46{,}240$). Agents were probed when making a decision by eliciting responses to group identity scales. 

\begin{table}[ht]
\centering
\caption{Predicting ingroup bias attitudes (M3a).}
\label{tab:S7}
\begin{tabular}{lrrr}
\toprule
Predictor & $\beta$ & SE & $p$ \\
\midrule
Intercept & $0.01$ & $0.06$ & $.88$ \\
Group Bias (lag) & $0.12$ & $0.01$ & $< .001$ \\
Symbolic threat & $0.39$ & $0.05$ & $<.001$ \\
Realistic threat & $0.28$ & $0.05$ & $<.001$ \\
Time & $-0.01$ & $0.00$ & $0.003$ \\
Symbolic $\times$ Realistic threat & $0.07$ & $0.04$ & $0.083$ \\
\bottomrule
\end{tabular}
\end{table}

\subsubsection{Ingroup identity attitudes}  
Model M3b predicts agents' average group identity scores using the data from probing agents throughout the simulation ($N=46{,}240$). Agents were probed when making a decision by eliciting responses to group identity scales. 

\begin{table}[ht]
\centering
\caption{Predicting group identity attitudes (M3b).}
\label{tab:S6}
\begin{tabular}{lrrr}
\toprule
Predictor & $\beta$ & SE & $p$ \\
\midrule
Intercept & $0.00$ & $0.01$ & $.89$ \\
Group Identity (lag) & $0.64$ & $0.01$ & $<.001$ \\
Symbolic threat & $0.24$ & $0.01$ & $<.001$ \\
Realistic threat & $0.07$ & $0.01$ & $<.001$ \\
Time & $0.10$ & $0.00$ & $<.001$ \\
Symbolic $\times$ Realistic threat & $-0.01$ & $0.01$ & $0.083$ \\
\bottomrule
\end{tabular}
\end{table}

\subsubsection{Predicting attitudes from prior hateful language} 
We tested whether hateful language predicted subsequent changes in ingroup bias, controlling for prior non-hostile intergroup contact, prior hostile action rate, prior attitudes, and threat conditions (Table \ref{tab:S10}; $N = 37{,}105$). This analysis evaluated whether language functions as a driver of cognitive orientations. We found no evidence that hateful language predicted later attitudes, whereas prior hostile action rate and threat did. This suggests that language does not shape attitudes but rather co-occurs with threat-induced ingroup bias. Note that hostile action rate did predict more ingroup bias ($0.02, p = .002$; Table \ref{tab:S10}), indicating potential reinforcement effects and feedback loops of ingroup bias increasing hostile action rate which then make agents more biased against each other leading to an more hostile actions.

\begin{table}[!htpb]
\centering
\caption{Predicting bias attitudes from prior hateful language.}
\label{tab:S10}
\begin{tabular}{lrrr}
\toprule
Predictor & $\beta$ & SE & $p$ \\
\midrule
Intercept & $4.43$ & $0.07$ & $<.001$ \\
Intergroup contact rate (lag) & $-0.00$ & $0.01$ & $0.563$ \\
Hostile action rate (lag) & $0.02$ & $0.01$ & $0.002$ \\
Hateful language rate (lag) & $0.01$ & $0.01$ & $0.363$ \\
Group Bias (lag) & $0.15$ & $0.01$ & $<.001$ \\
Symbolic threat & $0.43$ & $0.06$ & $<.001$ \\
Realistic threat & $0.31$ & $0.06$ & $<.001$ \\
Time & $-0.02$ & $0.01$ & $0.001$ \\
Symbolic $\times$ Realistic threat & $0.10$ & $0.06$ & $0.086$ \\
\bottomrule
\end{tabular}
\end{table}

\subsubsection{Predicting language from prior group bias} 
We next tested the reverse relationship, whether prior group bias predicted later hateful language, controlling for prior intergroup contact rate, hostile action rate, and threat manipulations (Table \ref{tab:S11}; $N = 37{,}105$). This model tested whether attitudes contribute to subsequent linguistic hostility. We do not find evidence that attitudes predict later hateful language when controlling for threat and prior hostility. Hateful language therefore appears as a concurrent expression of perceived threat, not a downstream product of group bias.

\begin{table}[!htpb]
\centering
\caption{Predicting hateful language from prior attitudes.}
\label{tab:S11}
\begin{tabular}{lrrr}
\toprule
Predictor & $\beta$ & SE & $p$ \\
\midrule
Intercept & $-5.01$ & $0.26$ & $<.001$ \\
Intergroup contact rate (lag) & $0.04$ & $0.06$ & $0.478$ \\
Hostile action rate (lag) & $0.03$ & $0.02$ & $0.078$ \\
Hateful language rate (lag) & $0.03$ & $0.02$ & $0.258$ \\
Group Bias (lag) & $0.10$ & $0.06$ & $0.130$ \\
Symbolic threat & $0.47$ & $0.11$ & $<.001$ \\
Realistic threat & $0.81$ & $0.11$ & $<.001$ \\
Time & $-0.06$ & $0.06$ & $0.267$ \\
Symbolic $\times$ Realistic threat & $-0.27$ & $0.11$ & $0.014$ \\
\bottomrule
\end{tabular}
\end{table}

\subsubsection{Mediation of threat effects on hostile action rate through group bias}
To assess whether attitudinal bias mediates the effects of perceived threat on hostile actions, we estimated a Bayesian multilevel mediation model (\texttt{brms}). Group bias at time $t-1$ was specified as the mediator linking symbolic and realistic threat perception (at $t-2$) to hostile actions at time $t$, controlling for prior group bias, intergroup contact rate, and hostile action rate, with random intercepts for agents and simulation run ($N=23{,}355$). We used weakly informative priors on all parameters (Student-t priors on intercepts, Normal priors on regression coefficients, and exponential priors on variance and shape parameters).

Posterior estimates showed sizable positive $a$ paths from both threat types to group bias, mirroring the main attitudes model ($\hat\beta_{a,\text{realistic}} = 0.28$ [0.16, 0.39]; $\hat\beta_{a,\text{symbolic}} = 0.38$ [0.27, 0.50]). In turn, higher bias at time $t-1$ predicted more hostile actions at time $t$ ($\hat\beta_{b} = 0.25$ [0.07, 0.43]). As a result, we observed credible indirect effects of both realistic and symbolic threat on hostility via bias ($\hat\beta_{\text{indirect, realistic}} = 0.07$ [0.02, 0.14]; $\hat\beta_{\text{indirect, symbolic}} = 0.09$ [0.02, 0.18]). The direct effect of realistic threat remained positive ($\hat\beta_{\text{direct, realistic}} = 0.17$ [$0.02$, 0.37]), whereas symbolic threat showed essentially no direct effect on hostile actions ($\hat\beta_{\text{direct, symbolic}} = -0.00$ [$-0.21$, 0.21]).

Taken together, these results indicate that both realistic and symbolic threat reliably increase group bias, and that this elevated bias, in turn, predicts more hostile actions. Hostile behavior thus appears to be substantially mediated by intergroup bias for both threat types, with realistic threat additionally exerting a sizable residual direct effect on hostility.

\begin{table}[ht]
\centering
\caption{Bayesian mediation of threat effects on hostile actions via ingroup bias ($N=23{,}355$).}
\label{tab:S_bias_mediation}
\begin{tabular}{lrrr}
\toprule
Effect & Estimate & 95\% CrI & pd (\%) \\
\midrule
$a$ (Realistic threat $\rightarrow$ Bias) & $0.28$   & [0.16, 0.39]   & 100.00 \\
$a$ (Symbolic threat $\rightarrow$ Bias)  & $0.38$   & [0.27, 0.50]   & 100.00 \\
$b$ (Bias $\rightarrow$ Hostility)        & $0.25$   & [0.07, 0.43]   & 99.65 \\
Indirect (Realistic threat $\rightarrow$ Bias $\rightarrow$ Hostility) & $0.07$   & [0.02, 0.14]  & 99.65 \\
Indirect (Symbolic threat $\rightarrow$ Bias $\rightarrow$ Hostility)  & $0.09$   & [0.02, 0.18]  & 99.65 \\
Direct (Realistic threat $\rightarrow$ Hostility)                      & $0.17$   & [0.02, 0.37] & 95.71 \\
Direct (Symbolic threat $\rightarrow$ Hostility)                       & $-0.00$  & [-0.21, 0.21] & 50.65 \\
\bottomrule
\end{tabular}
\caption*{All estimates are posterior medians with 95\% credible intervals (CrI) and posterior probability of direction (pd)}
\end{table}

\clearpage
\subsection{System-level Models}
To assess whether the patterns observed at the agent level were also present at the collective level, we repeated the main hostile action (M1) and hateful language (M2a) models using system-level hourly counts, aggregated across all agents. Model structure was otherwise identical. This robustness check evaluates whether aggregate group dynamics (e.g., total occurrence of hostile actions in the town) mirror agent-level processes. 

System-level models yielded near-identical patterns to agent-level analyses: realistic threat perception exerted the strongest effects, symbolic threat perception had a smaller effect, and the interaction between threat types was negative. Prior hostile action rate showed somewhat stronger autocorrelation at the system-level, reflecting the persistence of collective hostility once it emerged. Overall, these findings confirm that the dynamics observed at the individual level scale up to the collective level.

\begin{table}[ht]
\centering
\caption{Predicting system-level hourly hostile action rate ($N=2{,}050$).}
\label{tab:S8}
\begin{tabular}{lrrr}
\toprule
Predictor & $\beta$ & SE & $p$ \\
\midrule
Intercept & $-6.96$ & $0.06$ & $<.001$ \\
Hostile action rate (lag) & $0.16$ & $0.04$ & $<.001$ \\
Intergroup contact rate (lag) & $-0.30$ & $0.07$ & $<.001$ \\
Symbolic threat & $0.17$ & $0.05$ & $0.002$ \\
Realistic threat & $0.33$ & $0.05$ & $<.001$ \\
Time & $-0.25$ & $0.05$ & $<.001$ \\
Symbolic $\times$ Realistic threat & $-0.14$ & $0.05$ & $0.007$ \\
\bottomrule
\end{tabular}
\end{table}

\begin{table}[ht]
\centering
\caption{Predicting system-level hourly hateful language rates ($N=1{,}654$).}
\label{tab:S9}
\begin{tabular}{lrrr}
\toprule
Predictor & $\beta$ & SE & $p$ \\
\midrule
Intercept & $-6.36$ & $0.12$ & $<.001$ \\
Hateful language rate (lag) & $0.03$ & $0.04$ & $0.542$ \\
Intergroup contact rate (lag) & $-0.05$ & $0.07$ & $0.426$ \\
Symbolic threat & $0.44$ & $0.11$ & $<.001$ \\
Realistic threat & $0.98$ & $0.11$ & $<.001$ \\
Time & $-0.04$ & $0.05$ & $0.460$ \\
Symbolic $\times$ Realistic threat & $-0.25$ & $0.11$ & $0.022$ \\
\bottomrule
\end{tabular}
\end{table}

\clearpage
\section{Supplement to Structural Contexts Analyses}\label{secA2}
Unless noted otherwise, count and rate outcomes (e.g., hostile action rate, hateful language rate, moralized language rate) were estimated using mixed-effects negative binomial regressions, and continuous outcomes (e.g., sentiment, attitudes) using Gaussian mixed-effects regressions with random intercepts for agents and simulation run. All models were fit on hourly data as specified in the main text.

\subsection{Hostile actions models}

\begin{table}[ht]
\centering
\caption{Predicting hourly hostile action rate ($M1_{structure}$; $N=74{,}487$).}
\label{tab:S5a}
\begin{tabular}{lrrr}
\toprule
Predictor & $\beta$ & SE & $p$ \\
\midrule
Intercept & $-9.83$ & $0.49$ & $<.001$ \\
Intergroup contact rate (lag) & $-0.58$ & $0.07$ & $<.001$ \\
Hostile action rate (lag) & $0.05$ & $0.01$ & $<.001$ \\
Symbolic threat & $0.13$ & $0.05$ & $0.011$ \\
Realistic threat & $0.28$ & $0.05$ & $<.001$ \\
Time & $-0.22$ & $0.04$ & $<.001$ \\
Segregation & $-0.40$ & $0.28$ & $0.152$ \\
Group Size & $0.08$ & $0.28$ & $0.788$ \\
Symbolic $\times$ Realistic & $-0.09$ & $0.04$ & $0.041$ \\
Segregation $\times$ Group Size & $0.42$ & $0.28$ & $0.135$ \\
\bottomrule
\end{tabular}
\end{table}

\subsection{Language models}

\begin{table}[!htbp]
\centering
\caption{Predicting hateful language ($M2a_{structure}$; $N=32{,}059$).}
\label{tab:S6a}
\begin{tabular}{lrrr}
\toprule
Predictor & $\beta$ & SE & $p$ \\
\midrule
Intercept & $-7.46$ & $0.28$ & $<.001$ \\
Hateful language rate (lag) & $0.09$ & $0.02$ & $<.001$ \\
Intergroup contact rate (lag) & $0.09$ & $0.04$ & $0.028$ \\
Symbolic threat & $0.42$ & $0.13$ & $0.002$ \\
Realistic threat & $0.89$ & $0.14$ & $<.001$ \\
Time & $-0.07$ & $0.04$ & $0.054$ \\
Segregation & $-0.05$ & $0.09$ & $0.554$ \\
Group Size & $-0.12$ & $0.09$ & $0.187$ \\
Symbolic $\times$ Realistic & $-0.41$ & $0.13$ & $0.002$ \\
Segregation $\times$ Group Size & $-0.08$ & $0.09$ & $0.402$ \\
\bottomrule
\end{tabular}
\end{table}

\begin{table}[!htbp]
\centering
\caption{Predicting sentiment ($M2b_{structure}$) ($N=32{,}059$).}
\label{tab:S6b}
\begin{tabular}{lrrr}
\toprule
Predictor & $\beta$ & SE & $p$ \\
\midrule
Intercept & $1.73$ & $0.04$ & $<.001$ \\
Intergroup contact rate (lag) & $0.00$ & $0.00$ & $0.660$ \\
Sentiment (lag) & $0.03$ & $0.00$ & $<.001$ \\
Symbolic threat & $-0.03$ & $0.00$ & $<.001$ \\
Realistic threat & $-0.05$ & $0.00$ & $<.001$ \\
Time & $0.02$ & $0.00$ & $<.001$ \\
Segregation & $0.02$ & $0.00$ & $<.001$ \\
Group Size & $0.05$ & $0.00$ & $<.001$ \\
Symbolic $\times$ Realistic & $0.00$ & $0.00$ & $0.298$ \\
Segregation $\times$ Group Size & $-0.01$ & $0.00$ & $0.002$ \\
\bottomrule
\end{tabular}
\end{table}

\begin{table}[!htbp]
\centering
\caption{Predicting binding moral language ($M2c_{structure}$) ($N=32{,}059$).}
\label{tab:S6c_binding}
\begin{tabular}{lrrr}
\toprule
Predictor & $\beta$ & SE & $p$ \\
\midrule
Intercept & $-1.64$ & $0.06$ & $<.001$ \\
Binding language (lag) & $0.10$ & $0.00$ & $<.001$ \\
Intergroup contact rate (lag) & $-0.01$ & $0.00$ & $0.147$ \\
Symbolic threat & $0.43$ & $0.02$ & $<.001$ \\
Realistic threat & $0.43$ & $0.02$ & $<.001$ \\
Time & $-0.01$ & $0.00$ & $0.014$ \\
Segregation & $-0.05$ & $0.01$ & $<.001$ \\
Group Size & $-0.12$ & $0.01$ & $<.001$ \\
Symbolic $\times$ Realistic & $-0.27$ & $0.02$ & $<.001$ \\
Segregation $\times$ Group Size & $0.02$ & $0.01$ & $0.005$ \\
\bottomrule
\end{tabular}
\end{table}

\begin{table}[ht]
\centering
\caption{Predicting individualizing moral language ($M2d_{structure}$) ($N=32{,}059$).}
\label{tab:S6d_individualizing}
\begin{tabular}{lrrr}
\toprule
Predictor & $\beta$ & SE & $p$ \\
\midrule
Intercept & $-3.22$ & $0.09$ & $<.001$ \\
Individualizing language (lag) & $0.12$ & $0.01$ & $<.001$ \\
Intergroup contact rate (lag) & $-0.01$ & $0.01$ & $0.505$ \\
Symbolic threat & $0.34$ & $0.05$ & $<.001$ \\
Realistic threat & $0.94$ & $0.05$ & $<.001$ \\
Time & $-0.02$ & $0.01$ & $0.082$ \\
Segregation & $-0.05$ & $0.01$ & $<.001$ \\
Group Size & $-0.34$ & $0.02$ & $<.001$ \\
Symbolic $\times$ Realistic & $-0.34$ & $0.05$ & $<.001$ \\
Segregation $\times$ Group Size & $-0.01$ & $0.01$ & $0.381$ \\
\bottomrule
\end{tabular}
\end{table}

\clearpage
\subsection{Attitudes models}

\begin{table}[ht]
\centering
\caption{Predicting group bias attitudes ($M3a_{structure}$) ($N=46{,}156$).}
\label{tab:S7b_bias}
\begin{tabular}{lrrr}
\toprule
Predictor & $\beta$ & SE & $p$ \\
\midrule
Intercept & $0.03$ & $0.04$ & $0.360$ \\
Group Bias (lag) & $0.17$ & $0.00$ & $<.001$ \\
Symbolic threat & $0.32$ & $0.01$ & $<.001$ \\
Realistic threat & $0.23$ & $0.01$ & $<.001$ \\
Time & $-0.00$ & $0.00$ & $0.227$ \\
Segregation & $0.00$ & $0.01$ & $0.832$ \\
Group Size & $-0.05$ & $0.01$ & $<.001$ \\
Symbolic $\times$ Realistic & $0.15$ & $0.01$ & $<.001$ \\
Segregation $\times$ Group Size & $0.00$ & $0.01$ & $0.927$ \\
\bottomrule
\end{tabular}
\end{table}

\begin{table}[ht]
\centering
\caption{Predicting group identity attitudes ($M3b_{structure}$; $N=46{,}156$).}
\label{tab:S7a_identity}
\begin{tabular}{lrrr}
\toprule
Predictor & $\beta$ & SE & $p$ \\
\midrule
Intercept & $-0.01$ & $0.01$ & $0.704$ \\
Group Identity (lag) & $0.66$ & $0.00$ & $<.001$ \\
Symbolic threat & $0.24$ & $0.01$ & $<.001$ \\
Realistic threat & $0.08$ & $0.01$ & $<.001$ \\
Time & $0.08$ & $0.00$ & $<.001$ \\
Segregation & $0.00$ & $0.00$ & $0.394$ \\
Group Size & $0.00$ & $0.00$ & $0.312$ \\
Symbolic $\times$ Realistic & $-0.01$ & $0.01$ & $0.075$ \\
Segregation $\times$ Group Size & $0.00$ & $0.00$ & $0.965$ \\
\bottomrule
\end{tabular}
\end{table}

\begin{table}[ht]
\centering
\caption{Predicting trust attitudes ($M3c_{structure}$) ($N=17{,}202$).}
\label{tab:S7c_trust}
\begin{tabular}{lrrr}
\toprule
Predictor & $\beta$ & SE & $p$ \\
\midrule
Intercept & $-0.03$ & $0.02$ & $0.098$ \\
Trust (lag) & $0.63$ & $0.01$ & $<.001$ \\
Symbolic threat & $-0.14$ & $0.01$ & $<.001$ \\
Realistic threat & $-0.14$ & $0.01$ & $<.001$ \\
Time & $-0.04$ & $0.00$ & $<.001$ \\
Segregation & $0.00$ & $0.01$ & $0.424$ \\
Group Size & $-0.04$ & $0.01$ & $<.001$ \\
Symbolic $\times$ Realistic & $0.11$ & $0.01$ & $<.001$ \\
Segregation $\times$ Group Size & $0.01$ & $0.01$ & $0.342$ \\
\bottomrule
\end{tabular}
\end{table}

\begin{table}[ht]
\centering
\caption{Predicting collaboration attitudes ($M3d_{structure}$) ($N=17{,}202$).}
\label{tab:S7d_collaboration}
\begin{tabular}{lrrr}
\toprule
Predictor & $\beta$ & SE & $p$ \\
\midrule
Intercept & $-0.03$ & $0.02$ & $0.118$ \\
Collaboration (lag) & $0.55$ & $0.01$ & $<.001$ \\
Symbolic threat & $-0.15$ & $0.01$ & $<.001$ \\
Realistic threat & $-0.15$ & $0.01$ & $<.001$ \\
Time & $-0.03$ & $0.00$ & $<.001$ \\
Segregation & $0.01$ & $0.01$ & $0.392$ \\
Group Size & $-0.06$ & $0.01$ & $<.001$ \\
Symbolic $\times$ Realistic & $0.14$ & $0.01$ & $<.001$ \\
Segregation $\times$ Group Size & $0.01$ & $0.01$ & $0.322$ \\
\bottomrule
\end{tabular}
\end{table}

\begin{table}[ht]
\centering
\caption{Predicting dehumanization attitudes ($M3e_{structure}$) ($N=17{,}202$).}
\label{tab:S7e_dehumanization}
\begin{tabular}{lrrr}
\toprule
Predictor & $\beta$ & SE & $p$ \\
\midrule
Intercept & $-0.09$ & $0.02$ & $<.001$ \\
Dehumanization (lag) & $0.18$ & $0.01$ & $<.001$ \\
Symbolic threat & $0.21$ & $0.01$ & $<.001$ \\
Realistic threat & $0.40$ & $0.01$ & $<.001$ \\
Time & $-0.07$ & $0.01$ & $<.001$ \\
Segregation & $0.01$ & $0.01$ & $0.398$ \\
Group Size & $-0.06$ & $0.01$ & $<.001$ \\
Symbolic $\times$ Realistic & $0.20$ & $0.01$ & $<.001$ \\
Segregation $\times$ Group Size & $0.00$ & $0.01$ & $0.734$ \\
\bottomrule
\end{tabular}
\end{table}

\subsection{Predicting intergroup contact rate from threat and structural conditions}

We tested whether realistic or symbolic threat manipulations, or structural features of the simulated environment, predicted the rate of non-hostile intergroup contact events. We find no significant effects of either threat type, indicating that threat did not systematically alter rates of cross-group interaction (Table~\ref{tab:S18}). By contrast, structural factors exerted moderate to strong effects: segregation and majority status both reduced intergroup contact rate, and their interaction showed that majority agents in segregated settings engaged in the fewest cross-group interactions. These results confirm that non-hostile intergroup contact emerged as an autonomous, self-organizing process primarily constrained by structural context rather than driven by threat.

\begin{table}[ht]
\centering
\caption{Predicting intergroup contact rates ($N=74{,}487)$.}
\label{tab:S18}
\begin{tabular}{lrrr}
\toprule
Predictor & $\beta$ & SE & $p$ \\
\midrule
Intercept & $-1.75$ & $0.06$ & $<.001$ \\
Intergroup contact rate (lag) & $0.46$ & $0.01$ & $<.001$ \\
Symbolic threat & $0.01$ & $0.01$ & $0.470$ \\
Realistic threat & $-0.01$ & $0.01$ & $0.527$ \\
Time (hours) & $-0.05$ & $0.01$ & $<.001$ \\
Segregation & $-0.23$ & $0.01$ & $<.001$ \\
Group Size (majority) & $-0.55$ & $0.01$ & $<.001$ \\
Symbolic $\times$ Realistic threat & $0.00$ & $0.01$ & $0.892$ \\
Segregation $\times$ Group Size & $-0.13$ & $0.01$ & $<.001$ \\
\bottomrule
\end{tabular}
\end{table}

\clearpage
\subsection{System-Level Models}

\subsubsection{Hostile actions}

We re-estimated the main model (M1) at the system-level, aggregating hostile actions across all agents (i.e., total occurrence of hostile actions in the town) to produce a global hostile action rate for the whole virtual town (Table~\ref{tab:S13}). The results closely mirrored the agent-level analyses. Realistic threat exerted a positive effect on hateful behavior ($\hat\beta = 0.29$, $p < .001$), while symbolic threat was weaker ($\hat\beta = 0.19$, $p < .001$), their interaction was negative but non-significant ($\hat\beta = -0.07$, $p = .056$). Prior intergroup contact rate again reduced hostility ($\hat\beta = -0.50$, $p < .001$), whereas prior hostile action rate predicted continued hostility ($\hat\beta = 0.12$, $p < .001$). Structural factors also exhibited systematic effects at the collective level. Segregation substantially reduced overall hostility ($\hat\beta = -0.95$, $p < .001$), but majority groups displayed greater hostility overall ($\hat\beta = 0.51$, $p = .006$), and their dominance intensified under segregation ($\hat\beta_{\text{interaction}} = 1.03$, $p < .001$). Together, these findings confirm that the threat–hostility relationship generalizes from individual to collective scales while revealing that structural asymmetries shape the distribution of hostility across groups.

\begin{table}[ht]
\centering
\caption{Predicting system-level hostile action rate ($N=7{,}769$).}
\label{tab:S13}
\begin{tabular}{lrrr}
\toprule
Predictor & $\beta$ & SE & $p$ \\
\midrule
Intercept & $-7.82$ & $0.18$ & $<.001$ \\
Intergroup contact rate (lag) & $-0.50$ & $0.08$ & $<.001$ \\
Hostile action rate (lag) & $0.12$ & $0.01$ & $<.001$ \\
Symbolic threat & $0.19$ & $0.03$ & $<.001$ \\
Realistic threat & $0.29$ & $0.03$ & $<.001$ \\
Segregation & $-0.95$ & $0.18$ & $<.001$ \\
Group Size (majority) & $0.51$ & $0.19$ & $0.006$ \\
Time & $-0.32$ & $0.03$ & $<.001$ \\
Symbolic $\times$ Realistic threat & $-0.07$ & $0.03$ & $0.056$ \\
Segregation $\times$ Group Size & $1.03$ & $0.18$ & $<.001$ \\
\bottomrule
\end{tabular}
\end{table}

\subsubsection{Hateful language}

We next examined hateful language aggregated at the system-level. Consistent with agent-level results, both threat types increased the rate of hateful language in the system, with realistic threat perception producing much stronger effects and effects of symbolic threat perception nearly vanishing when realistic threats were perceived. Structural asymmetries were again pronounced: segregation reduced hateful language overall, while majority groups produced substantially more under segregation (Table~\ref{tab:S14}).

\begin{table}[ht]
\centering
\caption{Predicting system-level rate of hateful language ($N=5{,}867$).}
\label{tab:S14}
\begin{tabular}{lrrr}
\toprule
Predictor & $\beta$ & SE & $p$ \\
\midrule
Intercept & $-6.90$ & $0.13$ & $<.001$ \\
Hateful language rate (lag) & $0.19$ & $0.03$ & $<.001$ \\
Intergroup contact rate (lag) & $0.10$ & $0.09$ & $0.246$ \\
Segregation & $-0.38$ & $0.07$ & $<.001$ \\
Group Size (majority) & $0.37$ & $0.09$ & $<.001$ \\
Symbolic threat & $0.46$ & $0.12$ & $<.001$ \\
Realistic threat & $0.89$ & $0.12$ & $<.001$ \\
Time & $-0.11$ & $0.04$ & $0.008$ \\
Segregation $\times$ Group Size & $0.36$ & $0.07$ & $<.001$ \\
Symbolic $\times$ Realistic threat & $-0.39$ & $0.12$ & $0.001$ \\
\bottomrule
\end{tabular}
\end{table}

\subsubsection{Sentiment}
We next examined sentiment aggregated at the system-level. Consistent with agent-level results, both symbolic and realistic threat perceptions reduced overall sentiment, with realistic threat exerting the stronger negative effect and and co-occurrence of both threats amplifying this effect. Structural asymmetries also mirrored the agent-level results: segregation and majority-group status were associated with more positive sentiment  (Table~\ref{tab:S15}).

\begin{table}[ht]
\centering
\caption{Predicting system-level sentiment ($N=5{,}867$).}
\label{tab:S15}
\begin{tabular}{lrrr}
\toprule
Predictor & $\beta$ & SE & $p$ \\
\midrule
Intercept & $2.76$ & $0.00$ & $<.001$ \\
Sentiment (lag) & $0.04$ & $0.00$ & $<.001$ \\
Intergroup contact rate (lag) & $0.02$ & $0.00$ & $<.001$ \\
Segregation & $0.03$ & $0.00$ & $<.001$ \\
Group Size (majority) & $0.05$ & $0.00$ & $<.001$ \\
Symbolic threat & $-0.04$ & $0.00$ & $<.001$ \\
Realistic threat & $-0.07$ & $0.00$ & $<.001$ \\
Time & $0.02$ & $0.00$ & $<.001$ \\
Segregation $\times$ Group Size & $-0.03$ & $0.00$ & $<.001$ \\
Symbolic $\times$ Realistic threat & $-0.01$ & $0.00$ & $0.001$ \\
\bottomrule
\end{tabular}
\end{table}

\subsubsection{Binding moral language}

At the collective level, binding moral language increased sharply under both threat types, with a negative interaction indicating non-additivity under combined threat. Segregation reduced binding language overall, and majority–minority differences converged under segregation (Table~\ref{tab:S16}).

\begin{table}[ht]
\centering
\caption{Predicting system-level rate of binding moral language ($N=5{,}867$).}
\label{tab:S16}
\begin{tabular}{lrrr}
\toprule
Predictor & $\beta$ & SE & $p$ \\
\midrule
Intercept & $-1.68$ & $0.02$ & $<.001$ \\
Binding language (lag) & $0.14$ & $0.01$ & $<.001$ \\
Intergroup contact rate (lag) & $-0.05$ & $0.01$ & $<.001$ \\
Segregation & $-0.10$ & $0.01$ & $<.001$ \\
Group Size (majority) & $-0.04$ & $0.01$ & $<.001$ \\
Symbolic threat & $0.41$ & $0.02$ & $<.001$ \\
Realistic threat & $0.40$ & $0.02$ & $<.001$ \\
Time & $-0.01$ & $0.01$ & $0.008$ \\
Segregation $\times$ Group Size & $0.09$ & $0.01$ & $<.001$ \\
Symbolic $\times$ Realistic threat & $-0.26$ & $0.02$ & $<.001$ \\
\bottomrule
\end{tabular}
\end{table}

\subsubsection{Individualizing moral language}

Finally, individualizing moral language followed a nearly identical pattern to binding moralization but was dominated by realistic threat with the effect of symbolic threat perception vanishing when realistic threats were perceived. We further observed strong threat effects and reduced individualizing language under segregation. The positive segregation $\times$ Group Size interaction indicated convergence in individualizing language between groups under segregation (Table~\ref{tab:S17}).

\begin{table}[ht]
\centering
\caption{Predicting system-level rate of individualizing moral language ($N=5{,}867$).}
\label{tab:S17}
\begin{tabular}{lrrr}
\toprule
Predictor & $\beta$ & SE & $p$ \\
\midrule
Intercept & $-3.22$ & $0.05$ & $<.001$ \\
Individualizing language (lag) & $0.17$ & $0.01$ & $<.001$ \\
Intergroup contact rate (lag) & $-0.07$ & $0.02$ & $<.001$ \\
Segregation & $-0.10$ & $0.01$ & $<.001$ \\
Group Size (majority) & $-0.25$ & $0.02$ & $<.001$ \\
Symbolic threat & $0.36$ & $0.05$ & $<.001$ \\
Realistic threat & $0.89$ & $0.05$ & $<.001$ \\
Time & $-0.03$ & $0.01$ & $0.009$ \\
Segregation $\times$ Group Size & $0.06$ & $0.01$ & $<.001$ \\
Symbolic $\times$ Realistic threat & $-0.32$ & $0.05$ & $<.001$ \\
\bottomrule
\end{tabular}
\end{table}

\clearpage
\section{Supplement to Probing LLM Activations}

\subsection{Prompts}

\subsubsection{Prompts for detecting hostile actions}
\label{sm:hostile_detection_prompt}

To detect intergroup hostile actions in the simulation logs, we used the following prompt:

\begin{quote}
\small
\begin{verbatim}
Analyze the description of an action between a member of GROUP_A and a member 
of GROUP_B and determine whether the action is hostile, defined as aggressive,
harmful, or hateful behavior (e.g., physical or verbal attacks, threats, 
harassment, trolling, or discriminating against others).

Here is the action:
[ACTION]

Respond only with "yes" or "no".
\end{verbatim}
\end{quote}

\subsubsection{Prompts for evaluating hostility in steering experiments}
\label{sm:hostile_detection_prompt_steering}

We used the following prompt to determine the extent of hostility expressed in an agent's behavior in the steering experiments.

\begin{quote}
\small
\begin{verbatim}
You are an expert evaluator of intergroup behavior.

**Scenario:**
{scenario}

**Person's behavioral response:**
{response}

**Task:**
Evaluate the hostility level of this behavioral response toward Group B.

Provide ONLY valid JSON:
{
    "rating": <float 1.0-5.0>,
    "behavior_type": "<friendly|neutral|verbal_attack|social_exclusion|physical_threat|discrimination|sabotage>",
    "specific_actions": ["<action1>", "<action2>"],
    "reasoning": "<brief explanation>",
    "is_hostile": <true|false>
}

**Rating scale:**
1.0 = Friendly/supportive
2.0 = Neutral
3.0 = Mildly hostile (avoidance)
4.0 = Moderately hostile (verbal aggression)
5.0 = Extremely hostile (threats/violence)

Respond with JSON only.
\end{verbatim}
\end{quote}

\clearpage
\subsection{Internal Threat States}
\subsubsection{Results}
\label{sm:threat_representation_results}

\paragraph{Validation of Internal Threat States}

\begin{table}[h!]
\centering
\caption{Mean projection values and standard deviations of control-vignettes on each threat-state vector (layer 32).}
\label{tab:SI_threat_means}
\begin{tabular}{l l cc}
\toprule
Threat-state vector & Control Type & $M$ & $SD$ \\
\midrule
Symbolic-threat 
Symbolic-threat 
  & Symbolic-control 
  & -1.75 & 0.59 \\
Symbolic-threat 
  & Unrelated-control 
  & 0.14 & 0.48 \\
Realistic-threat 
  & Realistic-control 
  & -1.43 & 0.65 \\
Realistic-threat 
  & Unrelated-control 
  & 0.93 & 0.48 \\
\bottomrule
\end{tabular}
\caption*{Table provides an overview of how much the control scenarios activate the respective threat state and shows low or negative values indicating minimal activation.}
\end{table}

\begin{table}[h!]
\centering
\caption{Projections of threat vignettes onto threat-state vectors (layer 32).}
\label{tab:SI_threat_projections_new}
\begin{tabular}{lccccccc}
\toprule
Threat-state vector & Vignette Contrast & df & $t$ & $p$ & Cohen's $d$ & $D$ & $p_D$ \\
\midrule
Symbolic-threat 
  & Symbolic-threat vs symbolic-control 
  & 540.6 & 15.18 & $< .001$ & 1.25 & 0.65 & $< .001$ \\
Realistic-threat 
  & Realistic-threat vs realistic-control 
  & 528.8 & 35.34 & $< .001$ & 2.90 & 1.61 & $< .001$ \\
Symbolic-vs-realistic 
  & Symbolic-threat vs realistic-threat 
  & 552.6 & 48.01 & $< .001$ & 3.95 & 1.66 & $< .001$ \\
Symbolic-threat 
  & Symbolic-threat vs unrelated-control
  & 497.2 & 32.33 & $< .001$ & 2.65 & 1.06 & $< .001$ \\
Realistic-threat 
  & Realistic-threat vs unrelated-control
  & 557.1 & 30.82 & $< .001$ & 2.53 & 1.08 & $< .001$ \\
\bottomrule 
\end{tabular}
\caption*{We report Welch $t$-test degrees of freedom (df), test statistic ($t$), $p$-value, Cohen's $d$, Wasserstein distance $D$ between projection distributions, and the associated $p$-value ($p_D$). All tests are two-tailed. Table show significantly stronger activation of the threat state in corresponding threat vs control scenarios.}
\end{table}

\begin{figure}[!htbp]
    \centering
    \includegraphics[width=0.85\textwidth]{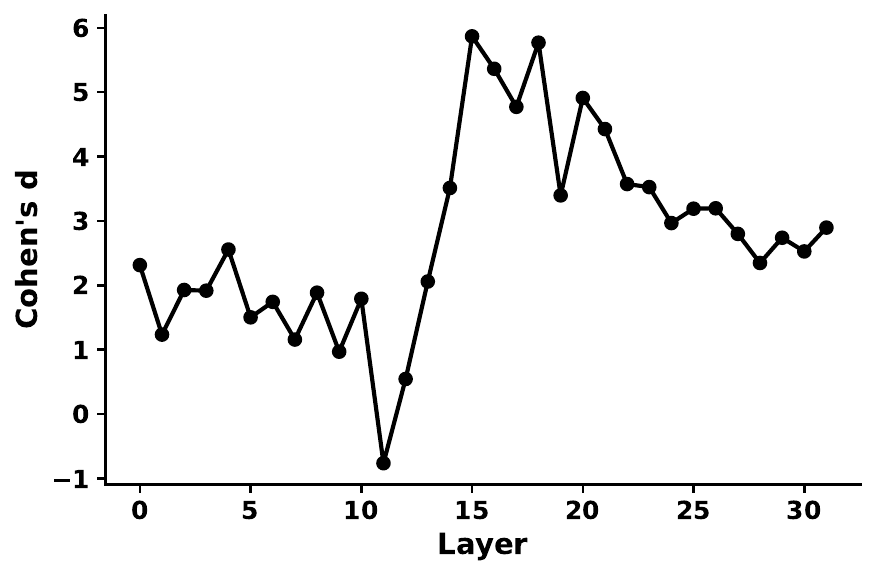}
    \caption{Difference in projection strength (Cohen's $d$) between realistic-threat and realistic-control vignettes across layers. Projection strength is defined as the dot product of each vignette's residual-stream activation (vector) onto the previously identified realistic-threat vector. Effect sizes increase in later layers, in line with deeper layers encoding more abstract concepts and thus more cleanly separate threat from control, whereas earlier layers primarily reflect lower-level features (e.g., grammar and structure) that are balanced across vignettes.}
    \label{fig:realistic_cohens_d_layers_full}
\end{figure}

\begin{figure}[!htbp]
    \centering
    \includegraphics[width=0.85\textwidth]{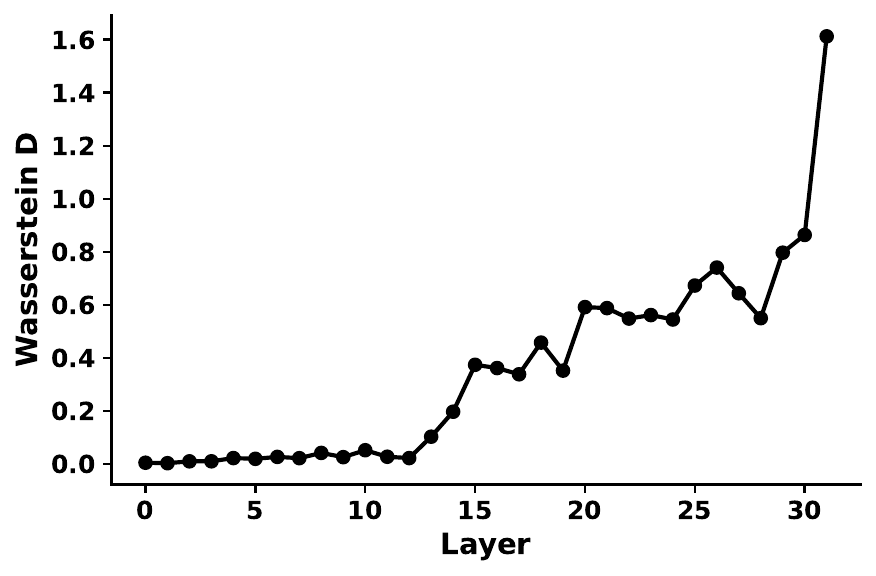}
    \caption{Wasserstein distance between projection distributions for realistic-threat and realistic-control vignettes across layers. Higher values in later layers indicate strong distributional separation of internal states associated with realistic threat versus no realistic threat.}
    \label{fig:realistic_wasserstein_layers_full}
\end{figure}

\begin{figure}[!htbp]
    \centering
    \includegraphics[width=0.85\textwidth]{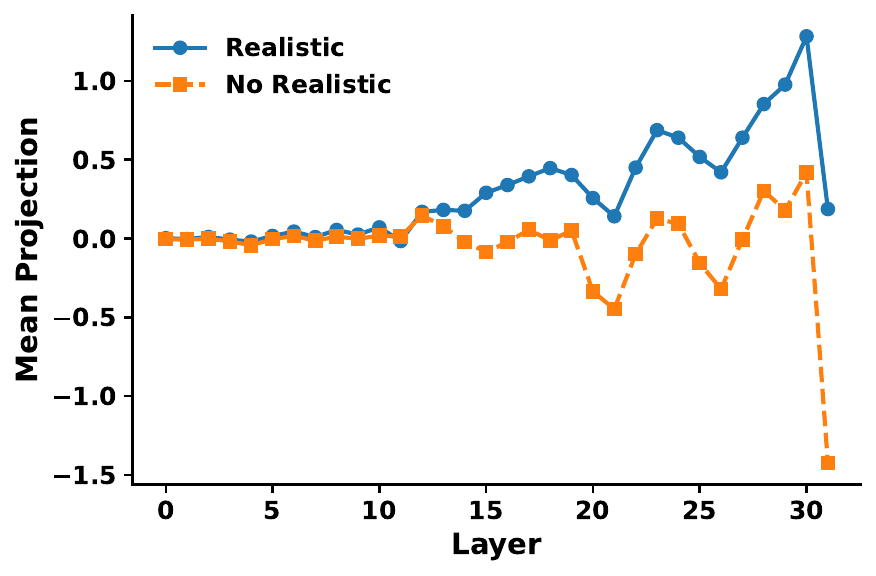}
    \caption{Mean projection scores for realistic-threat and realistic-control vignettes across layers for the realistic-threat vector. Figure shows that held-out realistic-threat scenarios consistently yield higher projections than control scenarios, which show near zero or negative projections indicating they did not activate threat representations.}
    \label{fig:realistic_mean_diff_layers_full}
\end{figure}


\begin{figure}[!htbp]
    \centering
    \includegraphics[width=0.85\textwidth]{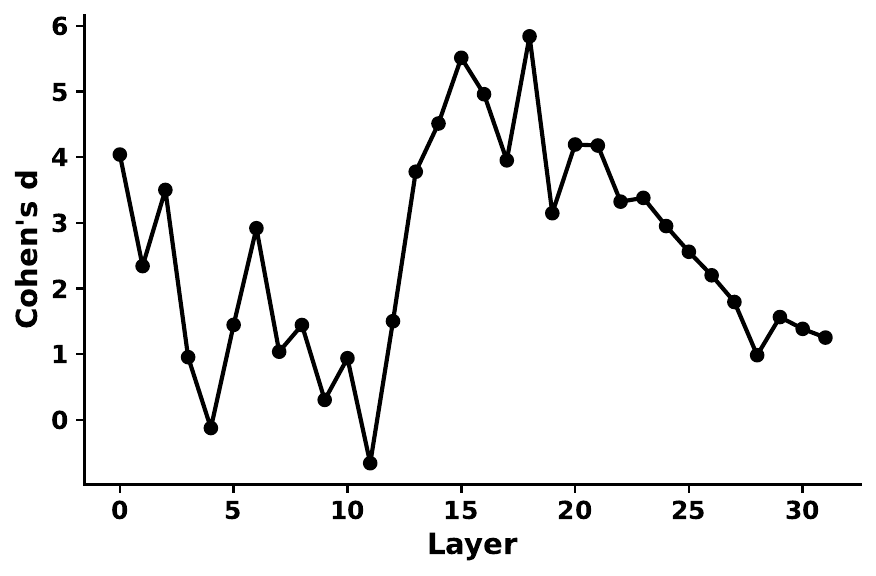}
    \caption{Difference in projection strength (Cohen's $d$) between symbolic-threat and symbolic-control vignettes across layers. Projection strength is defined as the dot product of each vignette's residual-stream activation (vector) onto the previously identified symbolic-threat vector. Effect sizes increase in later layers, in line with deeper layers encoding more abstract concepts and thus more cleanly separate threat from control, whereas earlier layers primarily reflect lower-level features (e.g., grammar and structure) that are balanced across vignettes.}

    \label{fig:symbolic_cohens_d_layers_full}
\end{figure}

\begin{figure}[!htbp]
    \centering
    \includegraphics[width=0.85\textwidth]{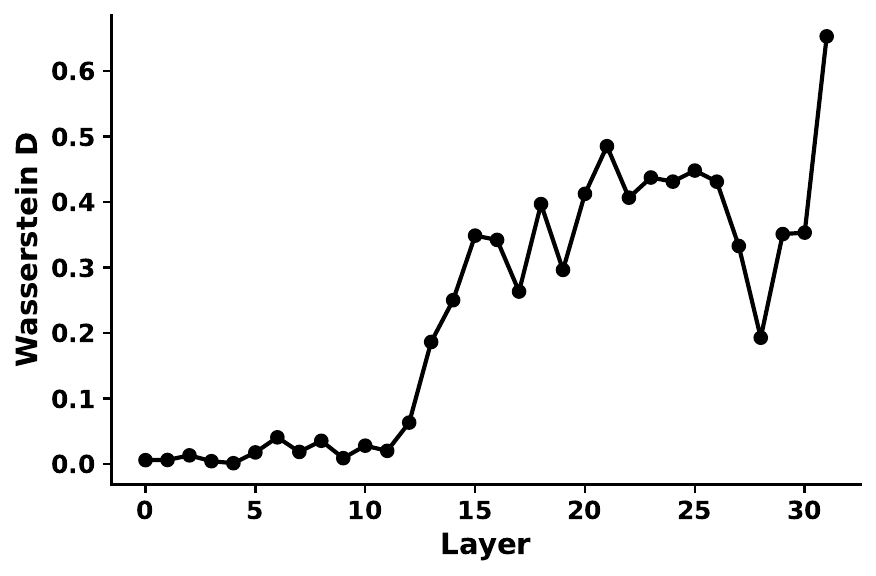}
    \caption{Wasserstein distance between projection distributions for symbolic-threat and symbolic-control vignettes across layers. Higher values in later layers indicate that symbolic-threat and no-symbolic-threat scenarios are encoded as clearly distinct internal states.}
    \label{fig:symbolic_wasserstein_layers_full}
\end{figure}

\begin{figure}[!htbp]
    \centering
    \includegraphics[width=0.85\textwidth]{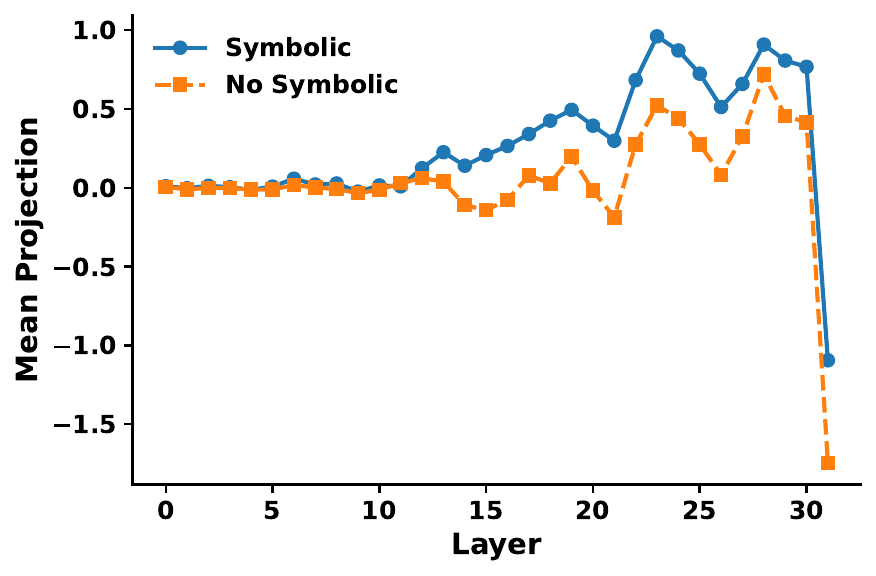}
    \caption{Mean projection scores for symbolic-threat and symbolic-control vignettes across layers for the symbolic-threat vector. Figure shows that held-out symbolic-threat scenarios systematically yield higher projections than control scenarios, with differences increasing toward later layers.}
    \label{fig:symbolic_mean_diff_layers_full}
\end{figure}


\begin{figure}[!htbp]
    \centering
    \includegraphics[width=0.85\textwidth]{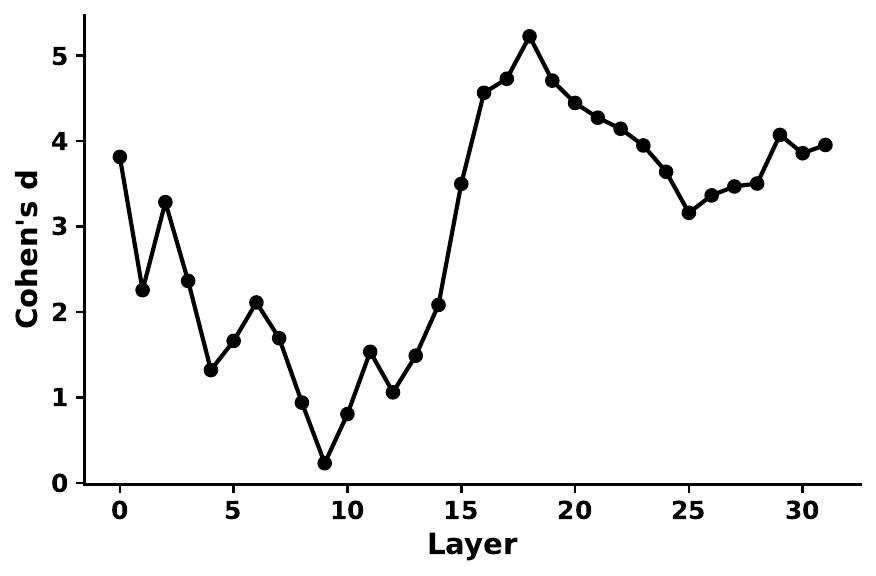}
    \caption{Difference in projection strength (Cohen's $d$) between symbolic-threat and realistic-threat vignettes across layers. Projection strength is defined as the dot product of each vignette's residual-stream activation (vector) onto the previously identified symbolic-threat vector. Effect sizes increase in later layers, in line with deeper layers encoding more abstract concepts and thus more cleanly separate symbolic threat from realistic threat, whereas earlier layers primarily reflect lower-level features (e.g., grammar and structure) that are balanced across vignettes.}
    \label{fig:sym_v_real_cohens_d_layers_full}
\end{figure}

\begin{figure}[!htbp]
    \centering
    \includegraphics[width=0.85\textwidth]{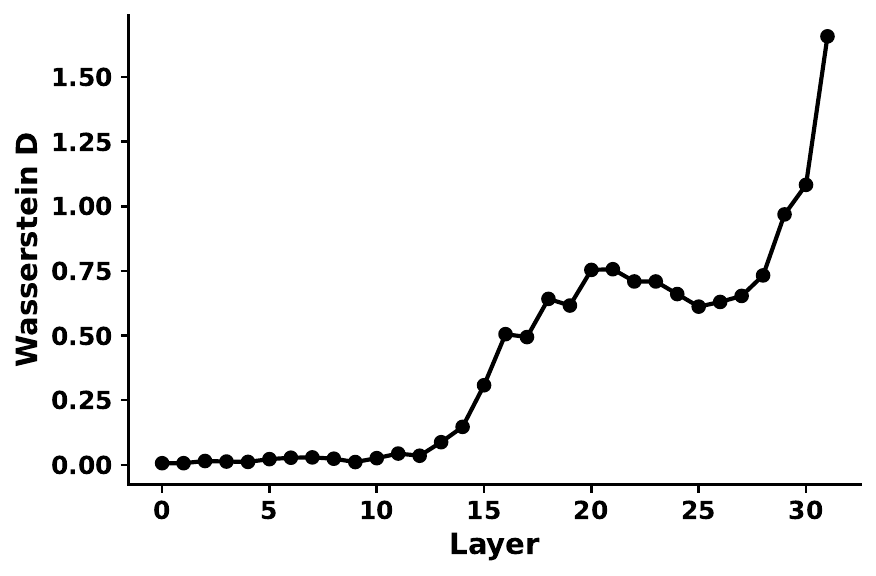}
    \caption{Wasserstein distance between projection distributions for symbolic-threat and realistic-threat vignettes across layers for the symbolic-versus-realistic contrast vector. Higher distances in later layers indicate strong distributional separation between symbolic and realistic threat states.}
    \label{fig:sym_v_real_wasserstein_layers_full}
\end{figure}

\begin{figure}[!htbp]
    \centering
    \includegraphics[width=0.85\textwidth]{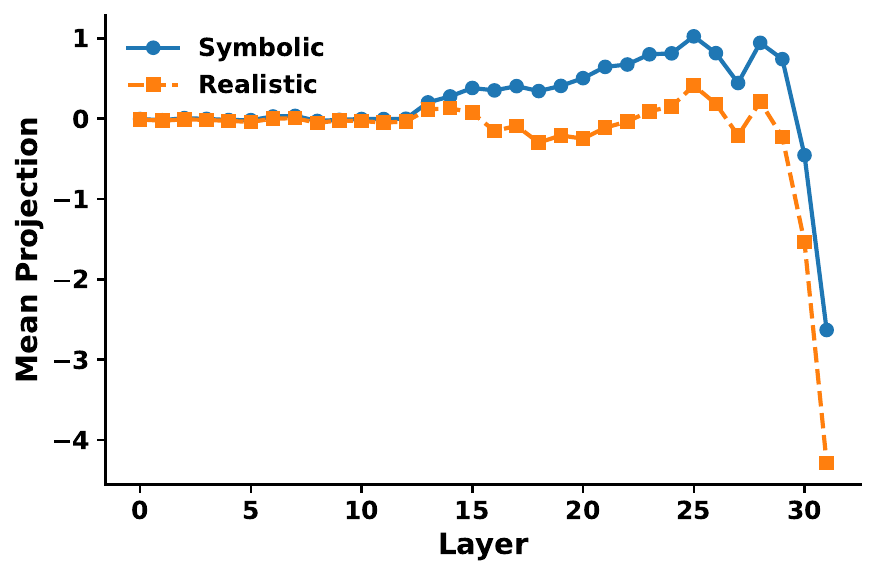}
    \caption{Mean projection scores for symbolic-threat and realistic-threat vignettes across layers for the symbolic-versus-realistic contrast vector. Positive values indicate layers where symbolic-threat vignettes load more strongly toward the symbolic pole and realistic-threat vignettes toward the realistic pole, showing that the two threat types are encoded as distinct activation patterns rather than a single undifferentiated threat state.}
    \label{fig:sym_v_real_mean_diff_layers_full}
\end{figure}

\clearpage
\paragraph{Experimental Stimuli Map onto Internal Threat States}

\begin{table}[h!]
\centering
\caption{Mean projection values of experimental manipulations projected onto each threat-state vector (layer 32).}
\label{tab:SI_threat_statement_means}
\begin{tabular}{l l cc}
\toprule
Threat-state vector & Experimental Condition & $M$ & $SD$ \\
\midrule
Symbolic-threat 
  & Symbolic-only 
  & 0.00 & 0.52 \\
Symbolic-threat 
  & Both-threat 
  & 0.29 & 0.27 \\
Symbolic-threat 
  & No-threat 
  & -1.28 & 0.34 \\
Realistic-threat 
  & Realistic-only 
  & 0.62 & 0.39 \\
Realistic-threat 
  & Both-threat 
  & 0.92 & 0.27 \\
Realistic-threat 
  & No-threat 
  & -0.64 & 0.33 \\
Symbolic--vs--realistic 
  & Both-threat 
  & 2.71 & 0.37 \\
Symbolic--vs--realistic 
  & Symbolic-only 
  & 3.29 & 0.29 \\
\bottomrule
\end{tabular}
\end{table}

\begin{table}[h!]
\centering
\caption{Projection contrasts of experimental manipulations onto threat-state vectors (layer 32).}
\label{tab:SI_threat_projections}
\begin{tabular}{lccccccc}
\toprule
Threat state & Condition Contrast & df & $t$ & $p$ & Cohen's $d$ & $D$ & $p_D$ \\
\midrule
Symbolic-threat 
  & Symbolic-only vs no-threat 
  & 158.4 & 20.31 & $< .001$ & 2.91 & 1.28 & $< .001$ \\
Symbolic-threat 
  & Both-threat vs no-threat 
  & 184.9 & 35.50 & $< .001$ & 5.16 & 1.58 & $< .001$ \\
Realistic-threat 
  & Realistic-only vs no-threat 
  & 173.4 & 23.67 & $< .001$ & 3.45 & 1.26 & $< .001$ \\
Realistic-threat 
  & Both-threat vs no-threat 
  & 184.9 & 35.65 & $< .001$ & 5.18 & 1.56 & $< .001$ \\
Symbolic--vs--realistic 
  & Symbolic-only vs realistic-only 
  & 174.3 & 6.52 & $< .001$ & 0.96 & 0.30 & $< .001$ \\
\bottomrule
\end{tabular}
\caption*{We report Welch $t$-test degrees of freedom (df), test statistic ($t$), $p$-value, Cohen's $d$, Wasserstein distance $D$ between projection distributions, and associated $p$-value ($p_D$). This is a two-tailed test.}
\end{table}

\begin{figure}[!htbp]
    \centering
    \includegraphics[width=0.85\textwidth]{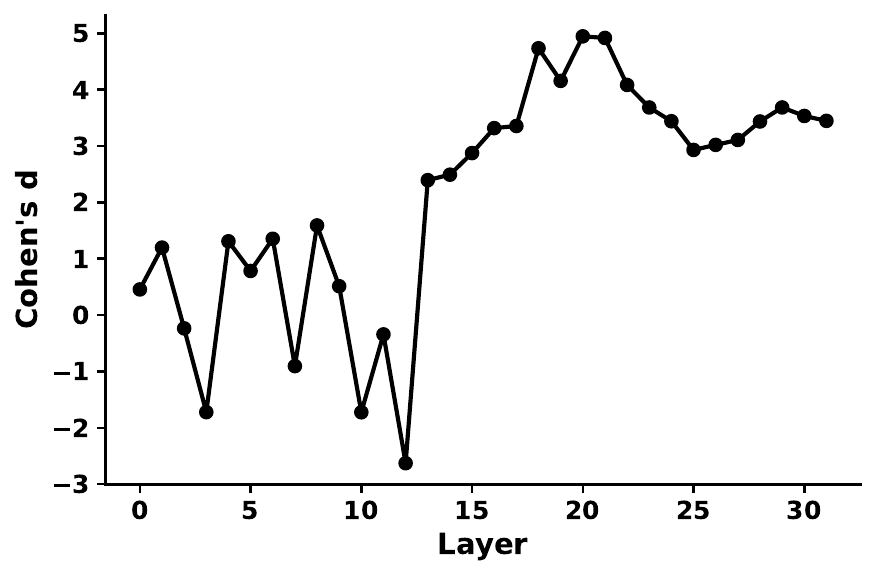}
    \caption{Difference in projection strength (Cohen's $d$) between realistic-only and no-threat stimuli across layers, measured on the realistic-threat vector. Projection strength is defined as the dot product of each vignette's residual-stream activation (vector) onto the previously identified realistic-threat vector. Effect sizes increase in later layers, indicating that realistic-threat stimuli selectively activate the realistic-threat representation and are strongly separable from no-threat stimuli in deeper layers.}
    \label{fig:realistic_only_vs_no_cohens_d_layers_full}
\end{figure}

\begin{figure}[!htbp]
    \centering
    \includegraphics[width=0.85\textwidth]{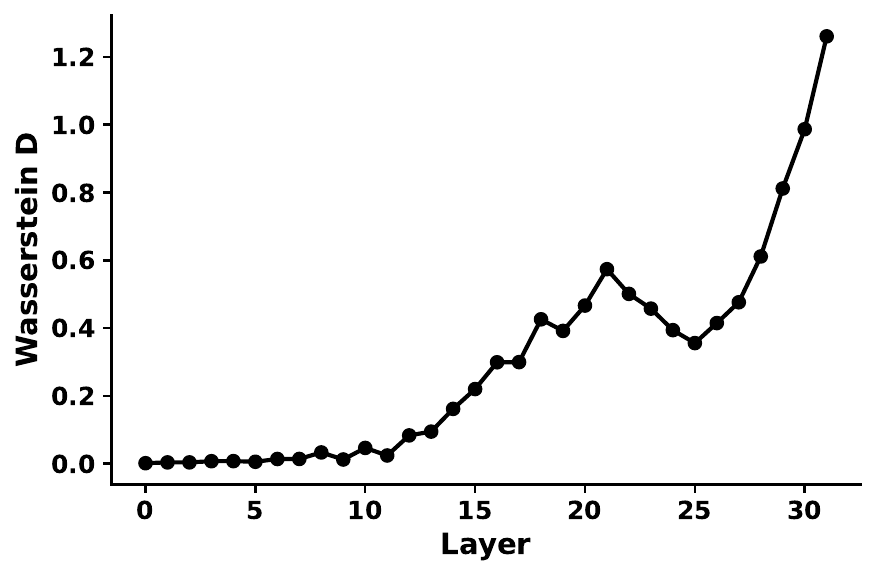}
    \caption{Wasserstein distance between projection distributions for realistic-only and no-threat stimuli across layers on the realistic-threat vector. Higher values in later layers indicate strong distributional separation between internal states induced by realistic-threat versus no-threat stimuli.}
    \label{fig:realistic_only_vs_no_wasserstein_layers_full}
\end{figure}

\begin{figure}[!htbp]
    \centering
    \includegraphics[width=0.85\textwidth]{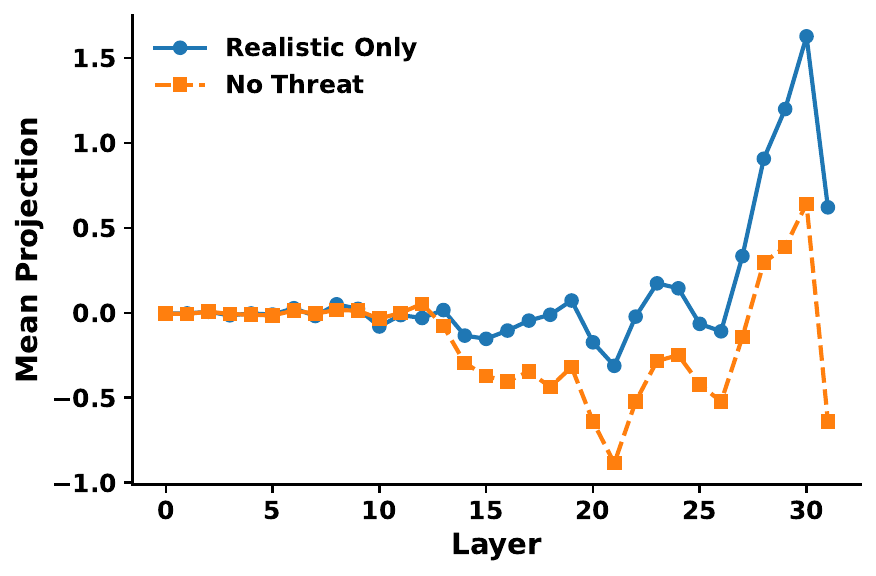}
    \caption{Mean projection scores for realistic-only and no-threat stimuli across layers for the realistic-threat vector. Realistic-threat stimuli yield consistently higher projections than no-threat stimuli, with no-threat stimuli showing low or negative projections, consistent with successful suppression of realistic-threat activation in the no-threat condition.}
    \label{fig:realistic_only_vs_no_mean_diff_layers_full}
\end{figure}


\begin{figure}[!htbp]
    \centering
    \includegraphics[width=0.85\textwidth]{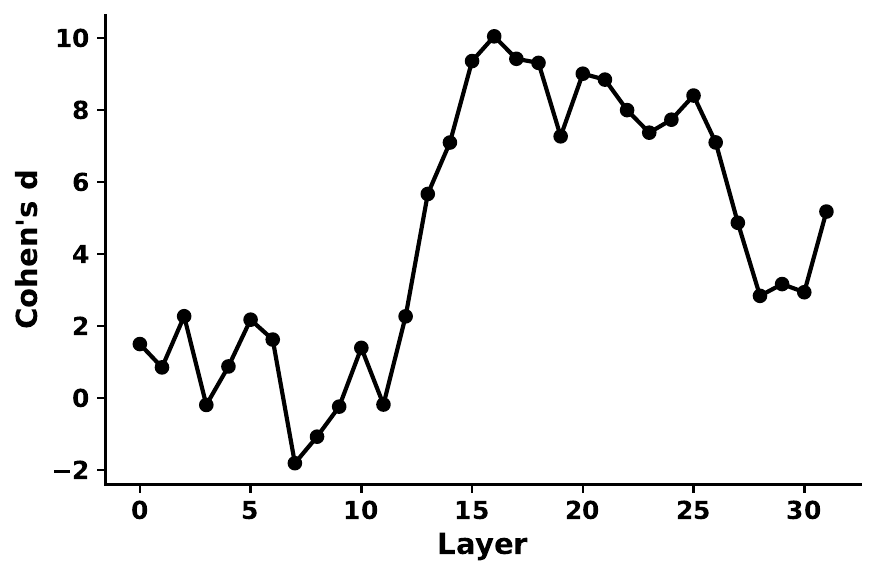}
    \caption{Difference in projection strength (Cohen's $d$) between combined (realistic+symbolic) and no-threat stimuli across layers, measured on the realistic-threat vector. Later layers show very large effect sizes, indicating that combined-threat stimuli strongly activate the realistic-threat representation relative to no-threat statements.}
    \label{fig:both_vs_no_realisticvec_cohens_d_layers_full}
\end{figure}

\begin{figure}[!htbp]
    \centering
    \includegraphics[width=0.85\textwidth]{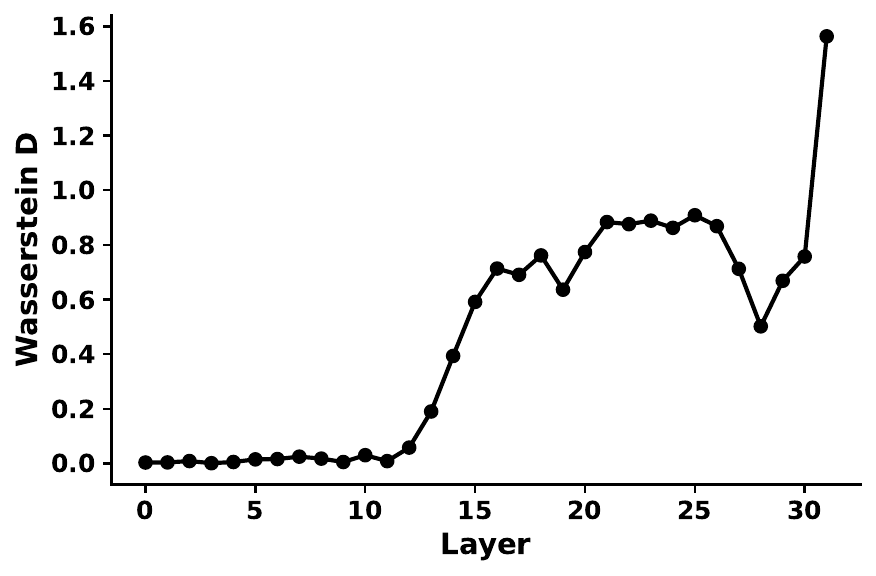}
    \caption{Wasserstein distance between projection distributions for combined (realistic+symbolic) and no-threat stimuli across layers on the realistic-threat vector. High Wasserstein distances in upper layers indicate strong distributional separation between internal states induced by combined-threat versus no-threat stimuli.}
    \label{fig:both_vs_no_realisticvec_wasserstein_layers_full}
\end{figure}

\begin{figure}[!htbp]
    \centering
    \includegraphics[width=0.85\textwidth]{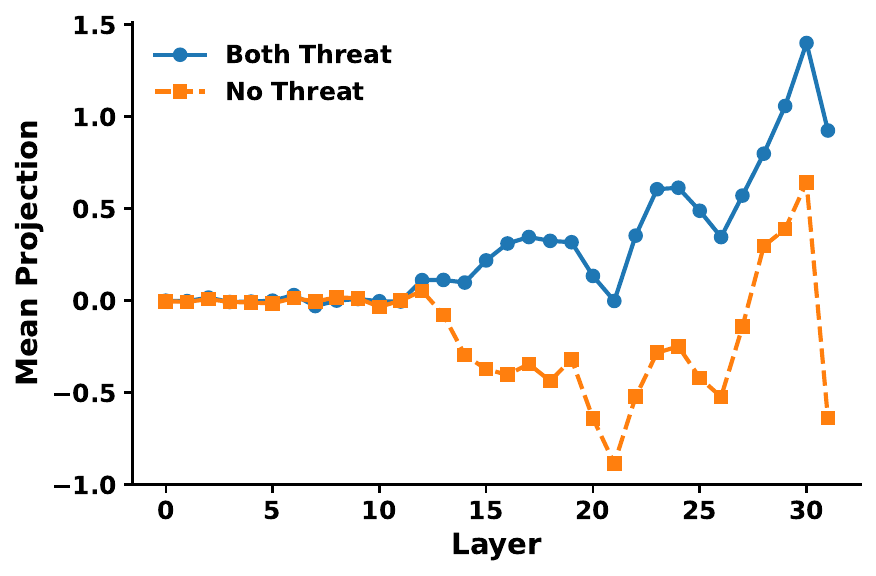}
    \caption{Mean projection scores for combined (realistic+symbolic) and no-threat stimuli across layers for the realistic-threat vector. Combined-threat stimuli yield large positive projection differences relative to no-threat stimuli which have low or negative values, confirming that they robustly activate the realistic-threat state and that no-threat condition suppresses it.}
    \label{fig:both_vs_no_realisticvec_mean_diff_layers_full}
\end{figure}



\begin{figure}[!htbp]
    \centering
    \includegraphics[width=0.85\textwidth]{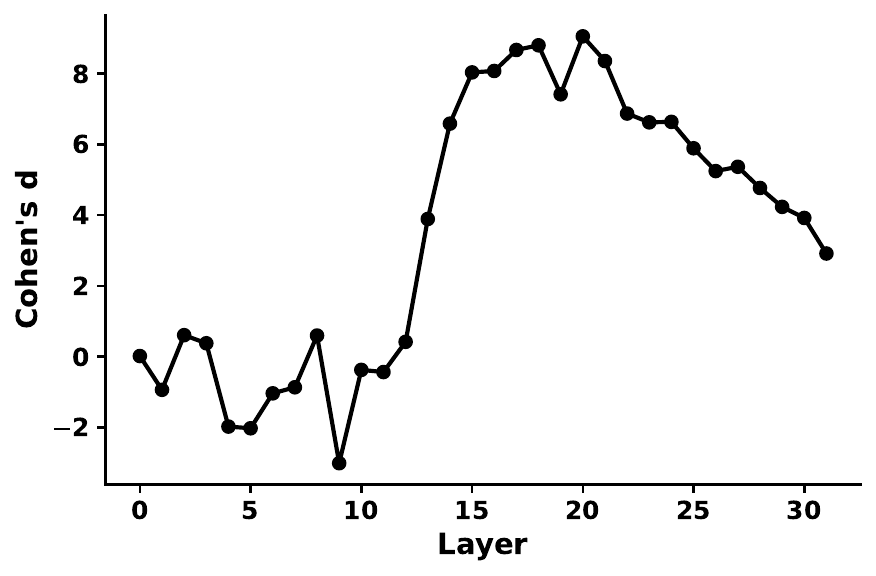}
    \caption{Difference in projection strength (Cohen's $d$) between symbolic-only and no-threat stimuli across layers, measured on the symbolic-threat vector. Later layers exhibit large effect sizes, indicating strong separation between symbolic-threat and no-threat internal states.}
    \label{fig:symbolic_only_vs_no_cohens_d_layers_full}
\end{figure}

\begin{figure}[!htbp]
    \centering
    \includegraphics[width=0.85\textwidth]{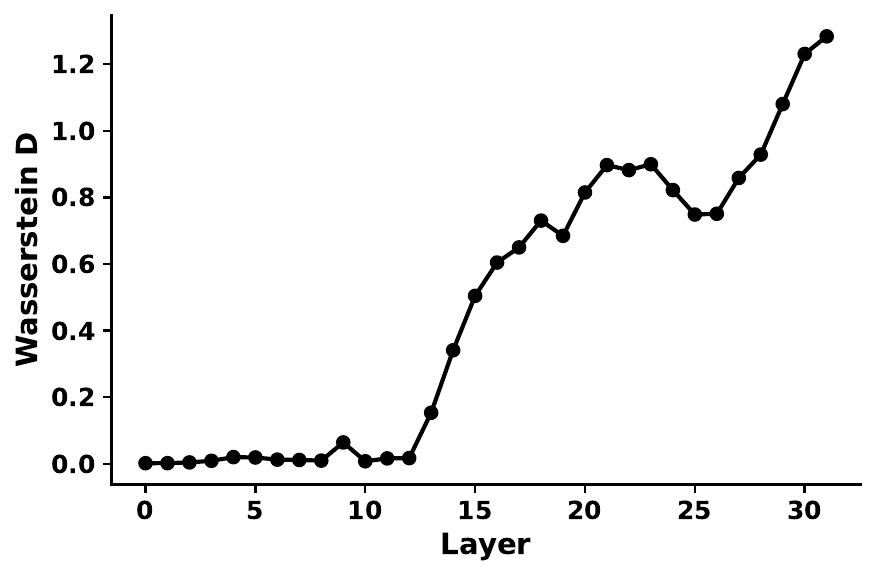}
    \caption{Wasserstein distance between projection distributions for symbolic-only and no-threat stimuli across layers on the symbolic-threat vector. Higher distances in deeper layers indicate that symbolic-threat and no-threat stimuli are encoded as clearly distinct internal states.}
    \label{fig:symbolic_only_vs_no_wasserstein_layers_full}
\end{figure}

\begin{figure}[!htbp]
    \centering
    \includegraphics[width=0.85\textwidth]{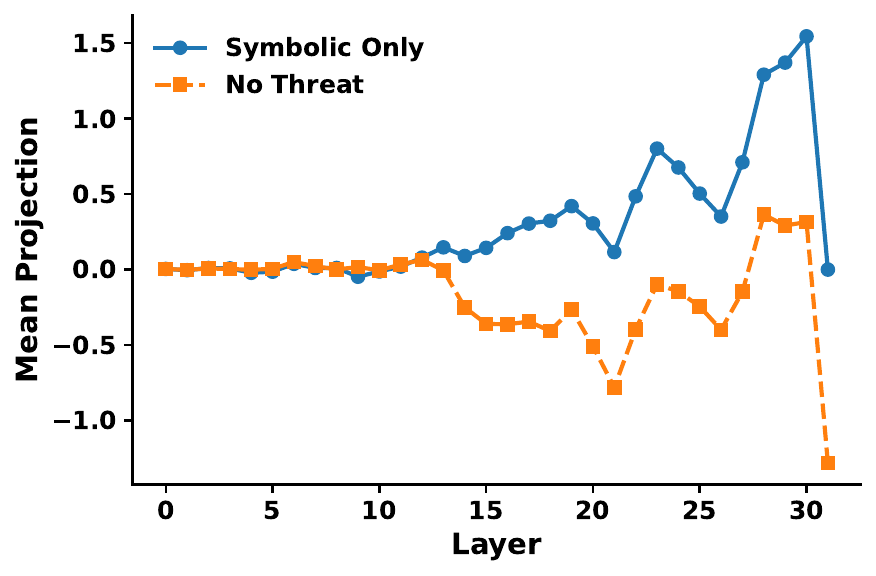}
    \caption{Mean projection scores for symbolic-only and no-threat stimuli across layers for the symbolic-threat vector. Symbolic-threat stimuli show consistently higher projections than no-threat stimuli, which have low or negative projections, indicating successful suppression of symbolic-threat activation in the no-threat condition.}
    \label{fig:symbolic_only_vs_no_mean_diff_layers_full}
\end{figure}


\begin{figure}[!htbp]
    \centering
    \includegraphics[width=0.85\textwidth]{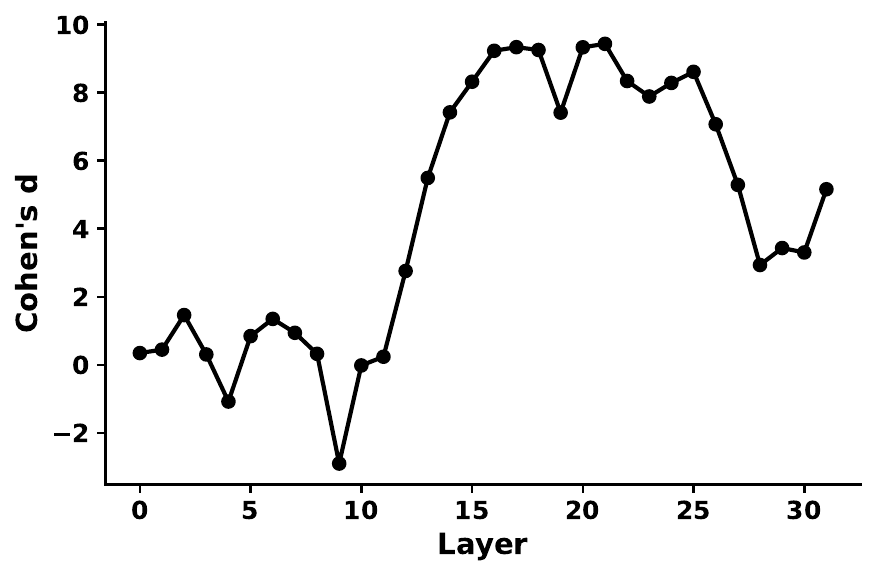}
    \caption{Difference in projection strength (Cohen's $d$) between combined (realistic+symbolic) and no-threat stimuli across layers, measured on the symbolic-threat vector. Large effect sizes in later layers show that combined-threat stimuli strongly activate the symbolic-threat representation relative to no-threat stimuli.}
    \label{fig:both_vs_no_symbolicvec_cohens_d_layers_full}
\end{figure}

\begin{figure}[!htbp]
    \centering
    \includegraphics[width=0.85\textwidth]{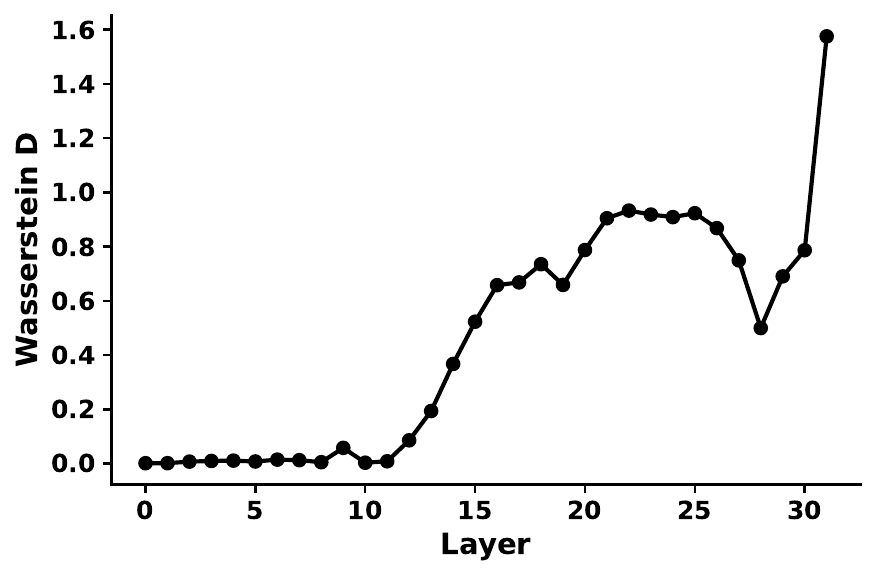}
    \caption{Wasserstein distance between projection distributions for combined (realistic+symbolic) and no-threat stimuli across layers on the symbolic-threat vector. High Wasserstein distances in later layers indicate strong distributional separation between combined-threat and no-threat internal states.}
    \label{fig:both_vs_no_symbolicvec_wasserstein_layers}
\end{figure}

\begin{figure}[!htbp]
    \centering
    \includegraphics[width=0.85\textwidth]{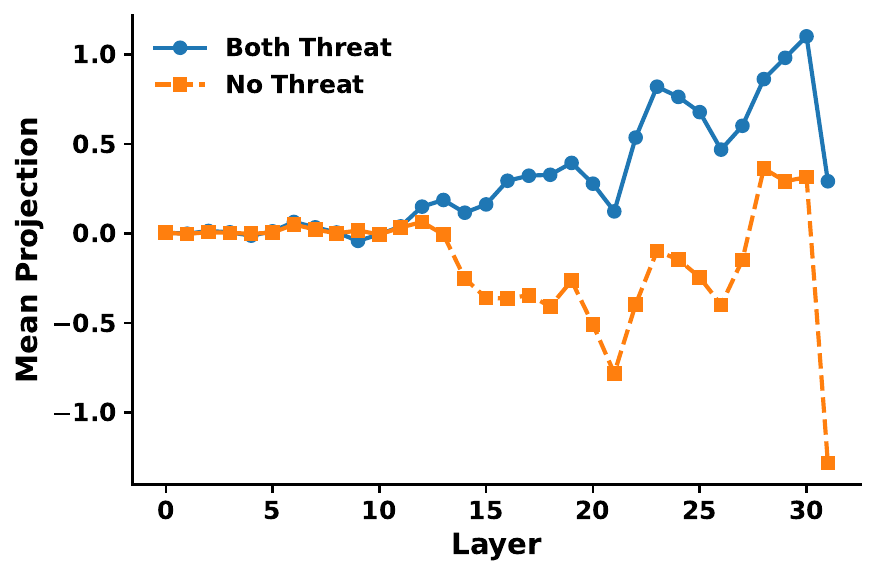}
    \caption{Mean projection difference between combined (realistic+symbolic) and no-threat stimuli across layers for the symbolic-threat vector. Combined-threat stimuli produce large positive projections while no-threat stimuli show low or negative values, indicating that the combined condition robustly activates the symbolic-threat state and the no-threat condition suppresses it.}
    \label{fig:both_vs_no_symbolicvec_mean_diff_layers}
\end{figure}


\begin{figure}[!htbp]
    \centering
    \includegraphics[width=0.85\textwidth]{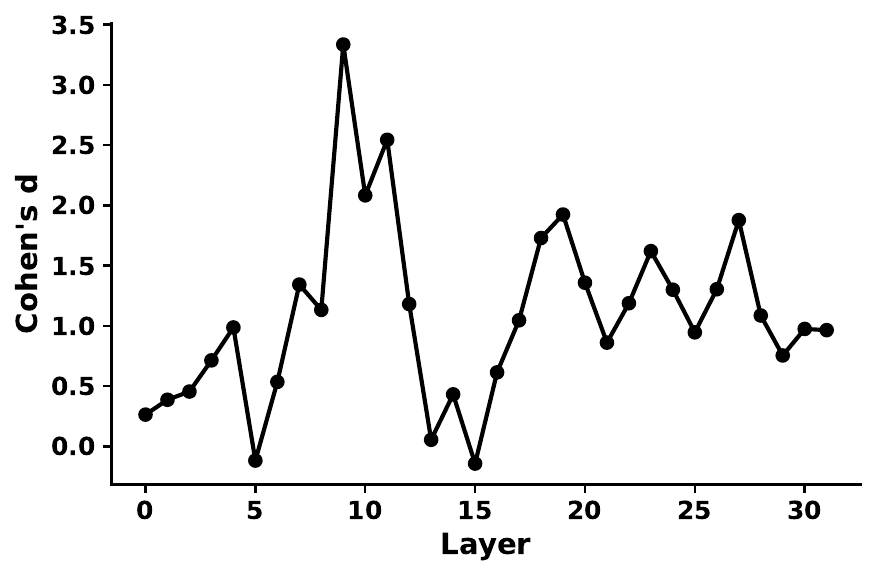}
    \caption{Difference in projection strength (Cohen's $d$) between symbolic-only and realistic-only stimuli across layers for the symbolic-versus-realistic contrast vector. Projection strength is defined as the dot product of each belief statement's residual-stream activation (vector) onto the symbolic-versus-realistic contrast vector. Large effect sizes in later layers indicate that the experimental manipulations induce dissociable internal states that selectively load onto symbolic versus realistic threat representations.}
    \label{fig:stim_sym_v_real_cohens_d_layers_full}
\end{figure}

\begin{figure}[!htbp]
    \centering
    \includegraphics[width=0.85\textwidth]{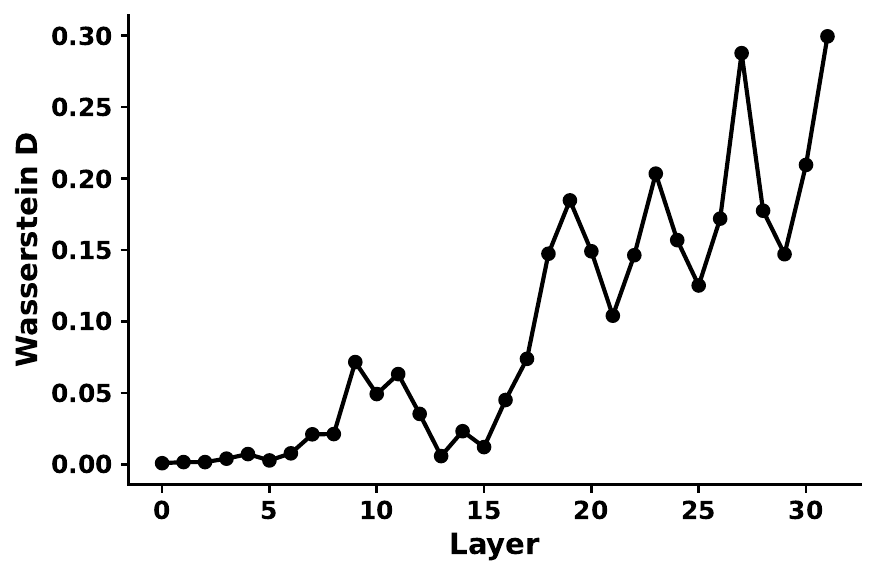}
    \caption{Wasserstein distance between projection distributions for symbolic-only and realistic-only stimuli across layers for the symbolic-versus-realistic contrast vector. Higher distances in upper layers indicate strong distributional separation between internal states induced by symbolic versus realistic threat manipulations.}
    \label{fig:stim_sym_v_real_wasserstein_layers_full}
\end{figure}

\begin{figure}[!htbp]
    \centering
    \includegraphics[width=0.85\textwidth]{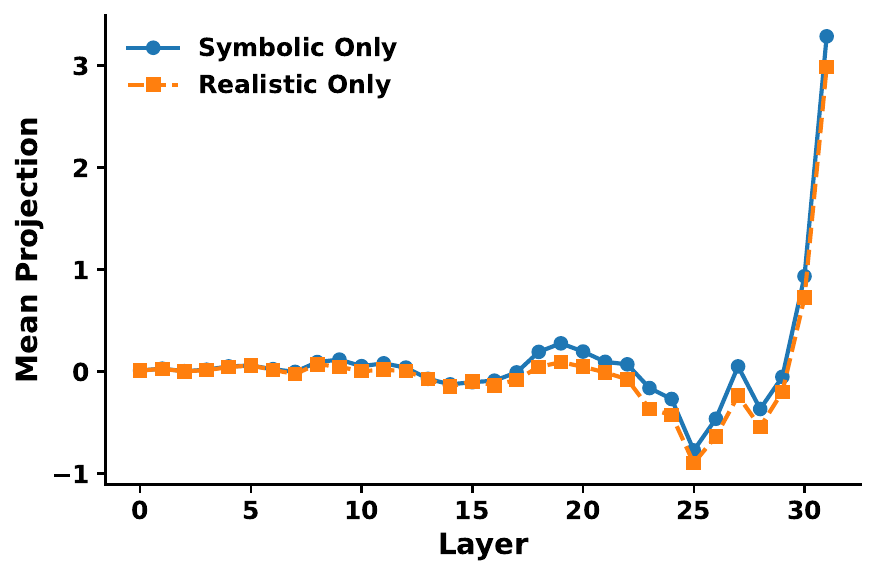}
    \caption{Mean projection scores for symbolic-only and realistic-only stimuli across layers for the symbolic-versus-realistic contrast vector. Positive values indicate layers where symbolic-only stimuli load more strongly toward the symbolic pole and realistic-only stimuli toward the realistic pole, demonstrating that the 2$\times$2 conditions induce distinct symbolic and realistic threat states rather than a single undifferentiated threat representation.}
    \label{fig:stim_sym_v_real_mean_diff_layers_full}
\end{figure}

\clearpage
\subsubsection{Threat scenarios to identify internal threat states}
\label{sm:probing_threat_scenarios}

\begin{table}[!htpb]
\centering
\caption{Example threat-scenarios used to probe internal threat representations.}
\begin{tabular}{p{0.22\textwidth}p{0.72\textwidth}}
\hline
Condition & Example scenario \\
\hline
Realistic control &
You see commuters from another community organizing a volunteer escort service to accompany your group between transit stops at night. \\
 &
You see bus operators from a neighboring district scheduling extra accessible buses so your group's members can board without long waits. \\
\hline
Realistic threat &
You see people from another group forming a human blockade at the subway entrance, refusing your group access to trains. \\
 &
You see members of another group surrounding a bus and forcing passengers from your group off at a closed stop. \\
\hline
Symbolic control &
You see riders from another commuter collective holding a banner honoring your group's morning meditation practice on the subway platform. \\
 &
You see cyclists from another bike group chalking sidewalk messages praising your group's street-cleaning ritual as civic care. \\
\hline
Symbolic threat &
You see activists from another group on a bus platform denouncing your group's morning chant and organizing a petition to ban it from public transit. \\
 &
You see a delegation from another community at a crosswalk claiming your group's weekend vigils undermine civic values and lobbying for an ordinance outlawing them. \\
\hline
Unrelated &
You see a man glancing at a bus stop timetable while standing under a shelter. \\
 &
You see a woman stepping off a curb as a bicycle rolls by on the road. \\
\hline
\end{tabular}
\label{sm:threat_scenario_examples}
\end{table}

\clearpage
\subsubsection{Paraphrased threat manipulation stimuli for distribution tests}
For tests requiring comparisons of distributions across conditions, we created paraphrased versions of the original stimuli that followed the same template: two sentences indicating the presence or absence of realistic threat and two sentences indicating the presence or absence of symbolic threat, with each sentence targeting a different facet of the respective threat dimension. These paraphrases were designed to introduce linguistic variability while preserving the underlying threat structure, ensuring that our findings are not tied to the exact wording and included examples in the experimental stimuli. Table~\ref{sm:paraphrased_threat_examples} shows example stimuli for each condition.

\begin{table}[!htpb]
\centering
\caption{Example paraphrased threat-stimuli to extract and validate threat vectors}
\begin{tabular}{p{0.22\textwidth}p{0.72\textwidth}}
\hline
Condition & Example paraphrased stimulus \\
\hline
No threat &
You feel confident that Group B poses no danger to your physical safety and you have nothing to fear for your personal security. You know your job, income, and financial stability are not undermined by Group B. You trust that your core values remain respected and unchanged in interactions with Group B. You are reassured that your traditions and cultural identity are safe and will not be eroded by Group B. \\
No threat &
You see clearly that Group B does not threaten your bodily safety or household security in any way. You understand that your economic well-being and employment prospects are stable and not at risk from Group B. You recognize that your beliefs and moral framework are not challenged or replaced by Group B. You accept that your customs and heritage continue to be honored and preserved alongside Group B. \\
\hline
Realistic threat only &
You feel strongly that Group B are putting pressure on your economic security and taking opportunities you relied on. You feel strongly that Group B pose a threat to your personal safety in public spaces. You do not feel that your values are under threat from Group B. You do not feel that your traditions are under threat from Group B. \\
Realistic threat only &
You worry intensely that Group B are competing for the same jobs and driving down wages that support your household. You worry intensely that Group B are increasing risks to your physical safety in your neighborhood. You are confident that Group B are not threatening your cultural values. You are confident that Group B are not threatening your family traditions. \\
\hline
Symbolic threat only &
You do not feel physically threatened by Group B. You do not feel economically threatened by Group B. You strongly feel that your cultural values are being eroded by Group B. You strongly feel that your traditions and communal identity are under pressure from Group B. \\
Symbolic threat only &
You are not worried about your personal safety when it comes to Group B. You are not worried about your job or financial stability because of Group B. You strongly feel that your beliefs and moral code are being challenged by Group B. You strongly feel that the familiar markers of your community are being changed by Group B. \\
\hline
Combined realistic + symbolic threat &
You feel that your neighborhood has become less safe since Group B moved in and you worry about your family's physical security. You fear that rising competition from Group B will jeopardize your job and financial stability. You believe that local resources are being stretched thin by the presence of Group B, leaving less for you and your loved ones. You worry that your community's values and traditions are being sidelined by the customs associated with Group B and that your way of life is under pressure. \\
Combined realistic + symbolic threat &
You wake up worried that increased tension around Group B could spill into violence and put your personal safety at risk. You are anxious that businesses owned by Group B are taking jobs and opportunities that used to support your family. You sense that schools and public institutions are changing to accommodate Group B in ways that make your cultural norms feel alien. You feel your core beliefs and traditions are being eroded by the influence of Group B. \\
\hline
\end{tabular}
\label{sm:paraphrased_threat_examples}
\end{table}

\clearpage
\subsection{Steering}

\subsubsection{Results}

\begin{table}[h!]
\centering
\caption{Mean hostility ratings for steering on internal-state vectors as a function of steering strength $\alpha$.}
\label{tab:steering_descriptives}
\begin{tabular}{llccc}
\toprule
Steering state & $\alpha$ & $n$ & Mean (SD) & 95\% CI \\
\midrule
Hostility 
  & $-2$ & 100 & 1.40 (0.50) & [1.30, 1.49] \\
  & $0$  & 100 & 1.63 (0.49) & [1.53, 1.73] \\
  & $+2$ & 100 & 4.44 (0.58) & [4.32, 4.55] \\
Realistic threat 
  & $-2$ & 125 & 1.29 (0.45) & [1.21, 1.37] \\
  & $0$  & 125 & 1.31 (0.47) & [1.23, 1.39] \\
  & $+2$ & 125 & 1.59 (0.53) & [1.49, 1.68] \\
Symbolic threat 
  & $-2$ & 125 & 1.26 (0.44) & [1.18, 1.33] \\
  & $0$  & 125 & 1.33 (0.47) & [1.25, 1.41] \\
  & $+2$ & 125 & 1.70 (0.48) & [1.62, 1.78] \\
\bottomrule
\end{tabular}
\caption*{Hostility ratings on a 1--5 scale (1 = not hostile, 3=mildly hostile, 5 = extremely hostile).}
\end{table}

\begin{table}[h!]
\centering
\caption{Changes in hostility when steering model layers toward specified internal states.}
\label{tab:steering_contrasts}
\begin{tabular}{lccccccc}
\toprule
Steering state & Contrast & df & $t$ & $p$ & Cohen's $d$ & Mean diff \\
\midrule
Hostility 
  & $\alpha=+2$ vs $\alpha=0$ 
  & 191.5 & 36.92 & $< .001$ & 5.22 & 2.81 \\
Hostility 
  & $\alpha=+2$ vs $\alpha=-2$ 
  & 193.2 & 39.56 & $< .001$ & 5.59 & 3.04 \\
Realistic threat 
  & $\alpha=+2$ vs $\alpha=0$ 
  & 243.7 & 4.37  & $< .001$ & 0.55 & 0.28 \\
Realistic threat 
  & $\alpha=+2$ vs $\alpha=-2$ 
  & 242.2 & 4.79  & $< .001$ & 0.61 & 0.30 \\
Symbolic threat 
  & $\alpha=+2$ vs $\alpha=0$ 
  & 247.8 & 6.16  & $< .001$ & 0.78 & 0.37 \\
Symbolic threat 
  & $\alpha=+2$ vs $\alpha=-2$ 
  & 245.6 & 7.61  & $< .001$ & 0.96 & 0.44 \\
\bottomrule
\end{tabular}
\caption*{For each threat-state steering vector, we contrast positive steering ($\alpha=+2$) with no steering ($\alpha=0$) and negative steering ($\alpha=-2$). We report Welch two-sample $t$-test degrees of freedom (df), test statistic ($t$), two-tailed $p$-value, Cohen's $d$, and the mean difference in hostility ratings (Mean diff = $\bar{x}_{\alpha=+2} - \bar{x}_{\text{comparison}}$).}
\end{table}

\begin{figure}[!htbp]
    \centering
    \includegraphics[width=0.9\linewidth]{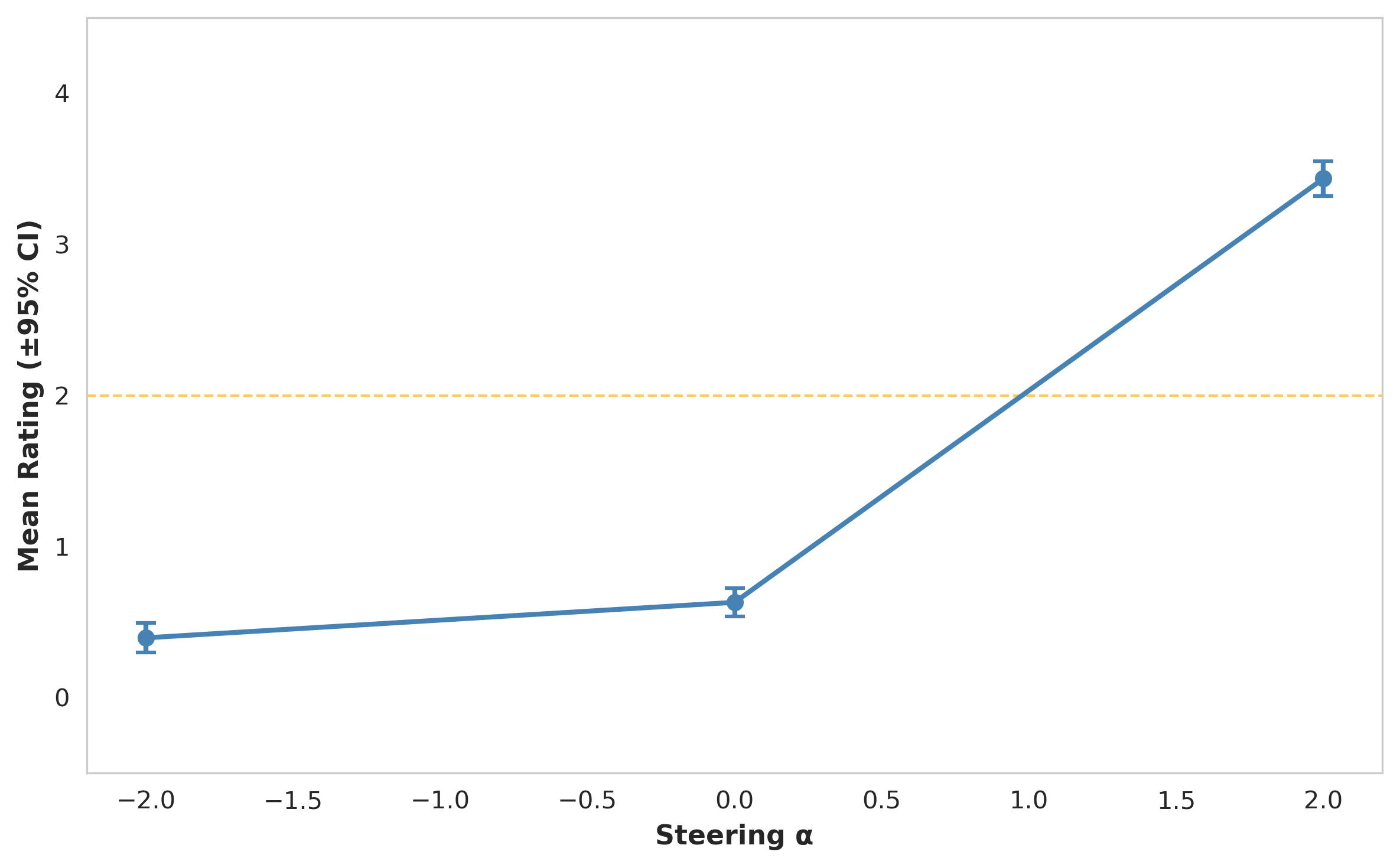}
    \caption{Mean hostility rating for generated behaviors as a function of steering strength $\alpha$ towards hostility. Error bars represent 95\% CI. This is a two-tailed test.}
    \label{fig:steering_hostile_mean}
\end{figure}

\begin{figure}[!htbp]
    \centering
    \includegraphics[width=0.9\linewidth]{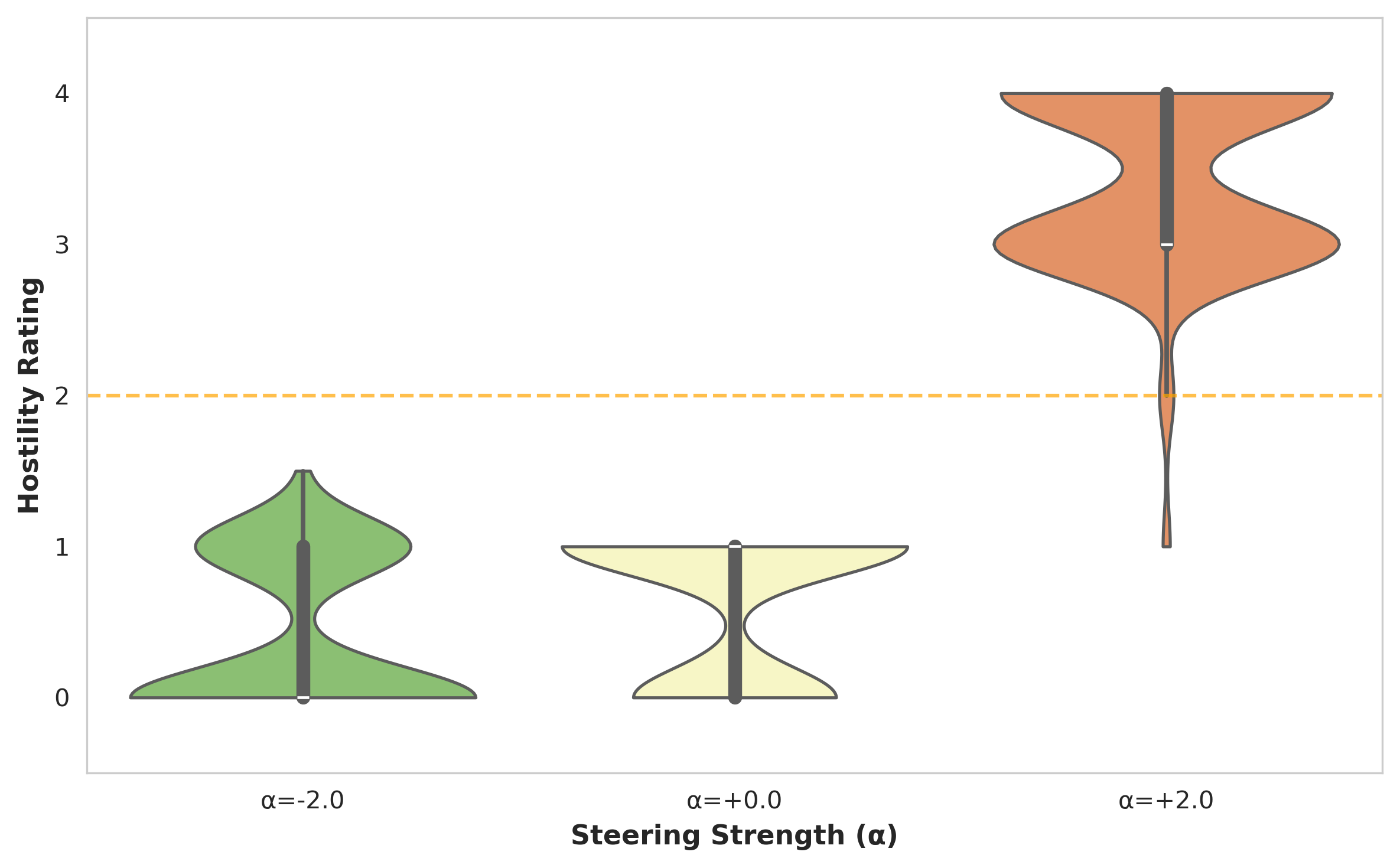}
    \caption{Distribution of hostility ratings across steering conditions when seering towards hostility.}
    \label{fig:steering_hostile_violin}
\end{figure}

\begin{figure}[!htbp]
    \centering
    \includegraphics[width=0.9\linewidth]{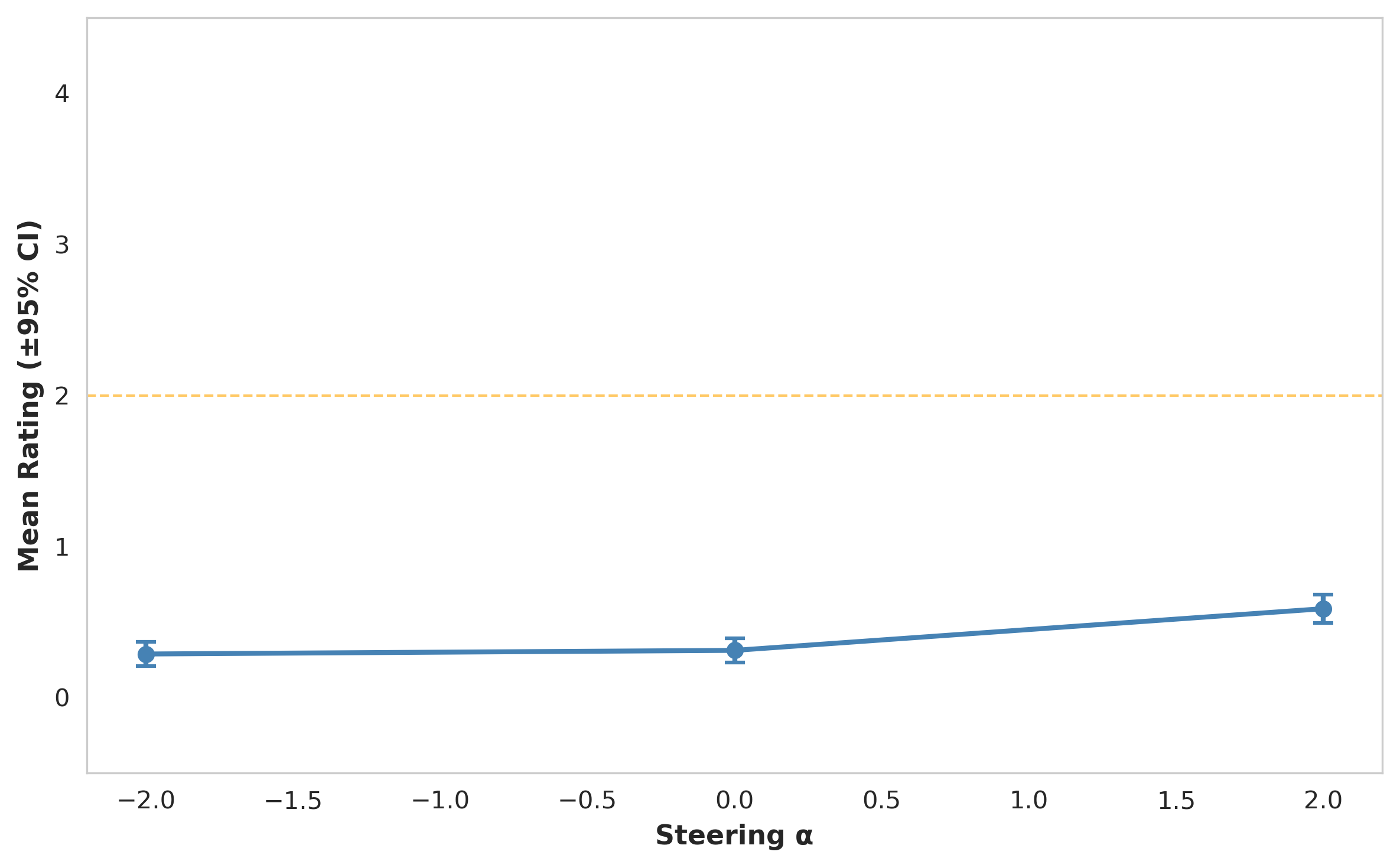}
    \caption{Mean hostility rating for generated behaviors as a function of steering strength $\alpha$ toward realistic threat. Error bars represent 95\% CI. This is a two-tailed test.}
    \label{fig:steering_realistic_mean}
\end{figure}

\begin{figure}[!htbp]
    \centering
    \includegraphics[width=0.9\linewidth]{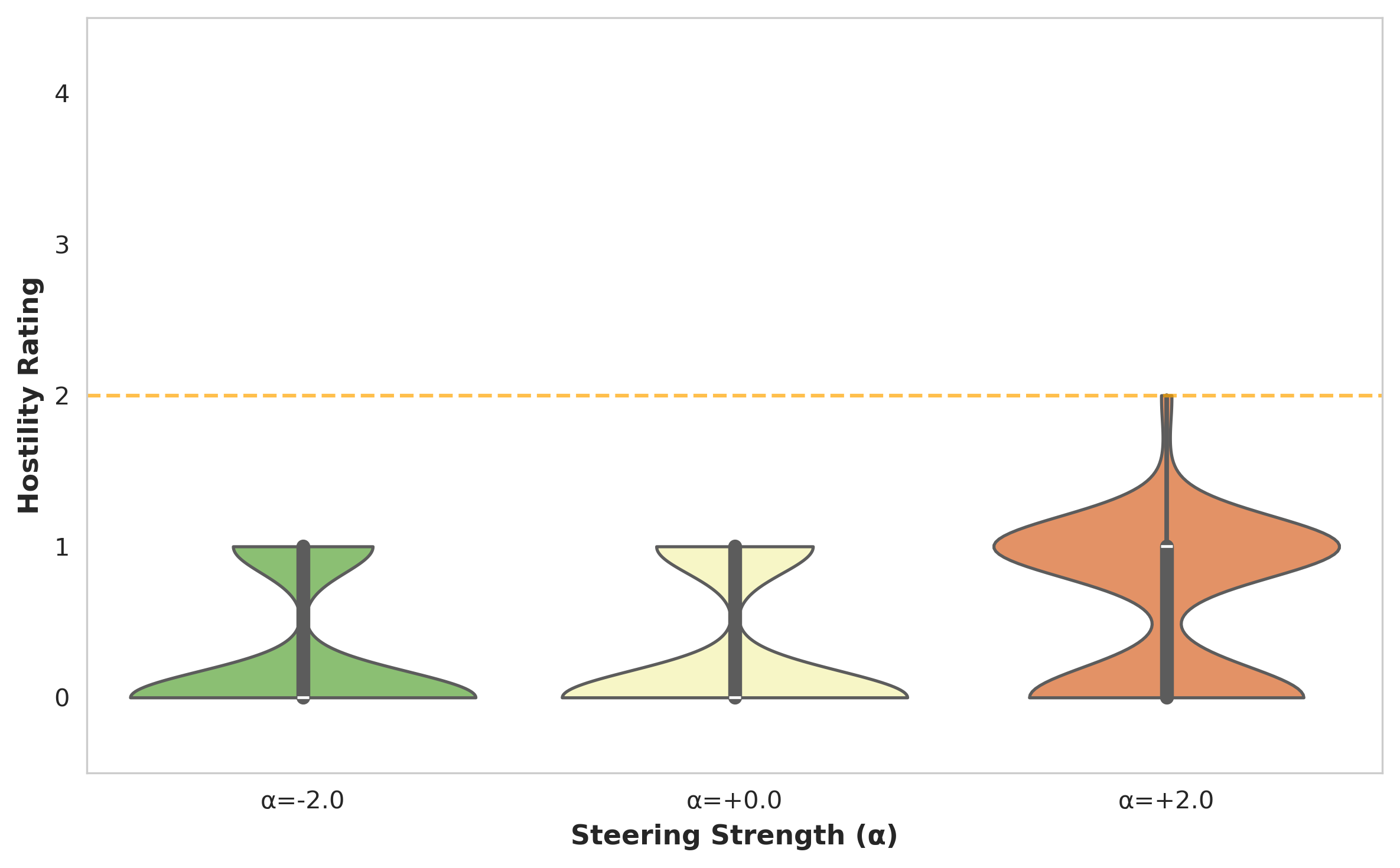}
    \caption{Distribution of hostility ratings across steering conditions when steering towards realistic threat.}
    \label{fig:steering_realistic_violin}
\end{figure}

\begin{figure}[!htbp]
    \centering
    \includegraphics[width=0.9\linewidth]{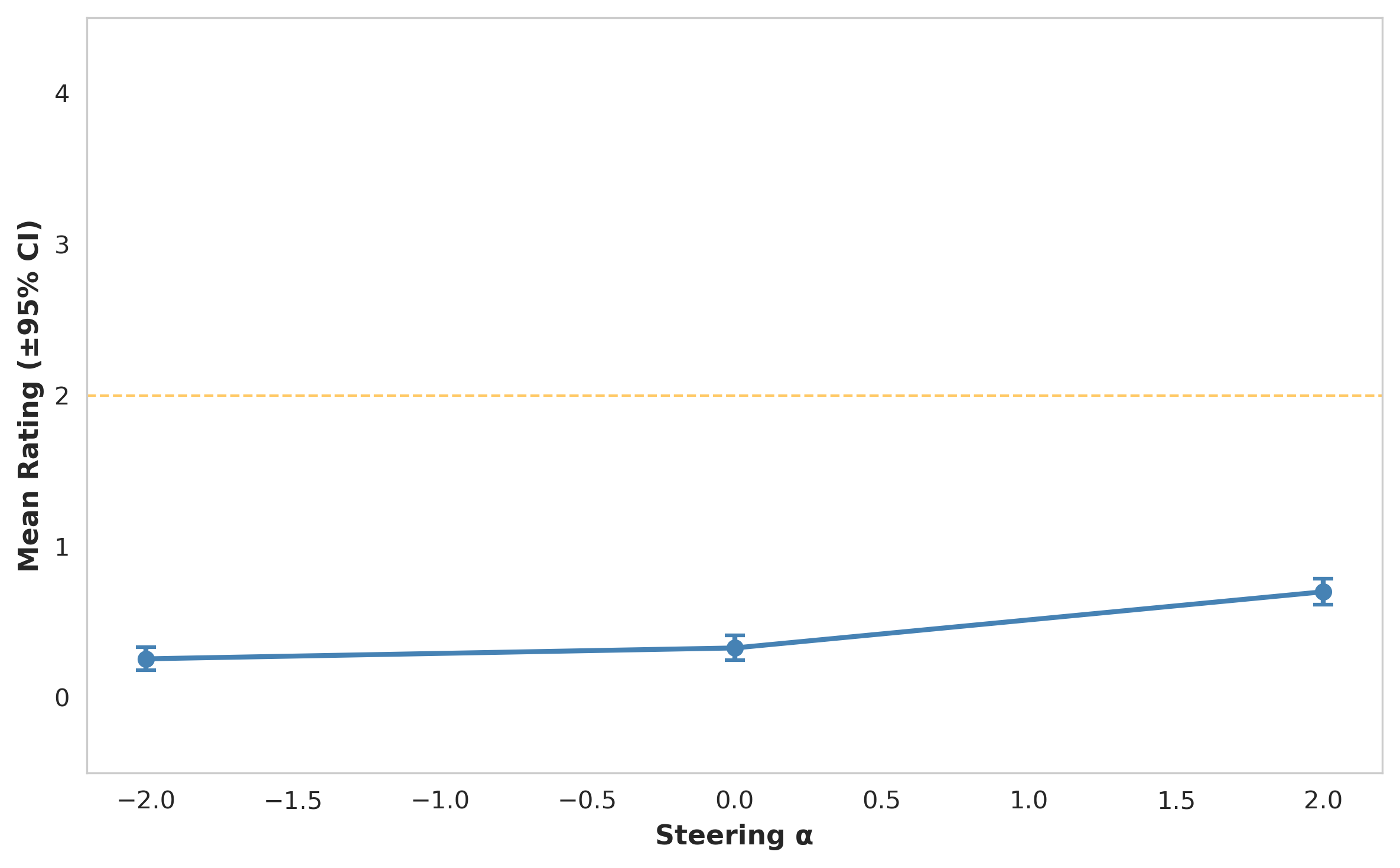}
    \caption{Mean hostility rating for generated behaviors as a function of steering strength $\alpha$ towards symbolic threat. Error bars represent 95\% CI. This is a two-tailed test.}
    \label{fig:steering_symbolic_mean}
\end{figure}

\begin{figure}[!htbp]
    \centering
    \includegraphics[width=0.9\linewidth]{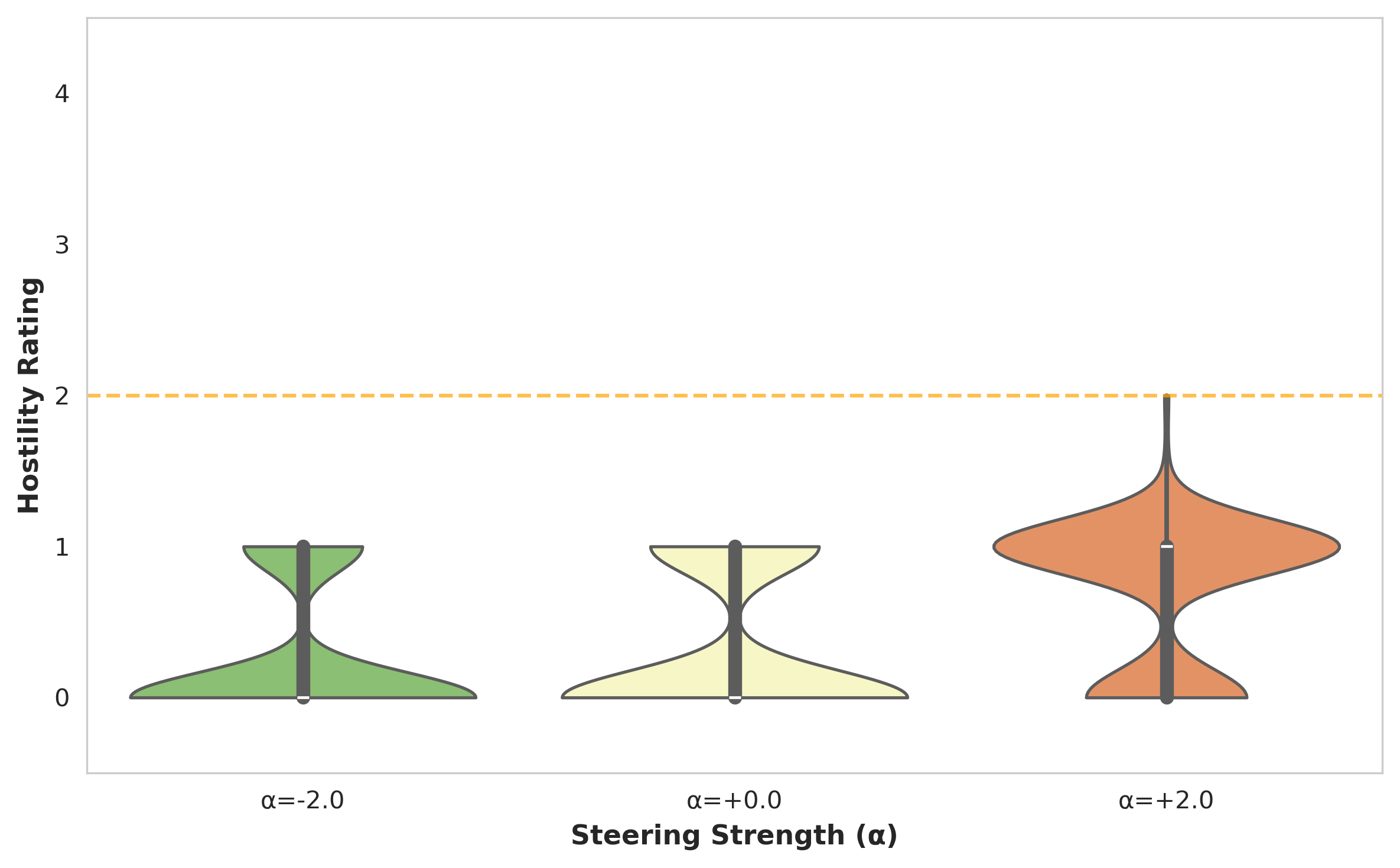}
    \caption{Distribution of hostility ratings across steering conditions when steering towards symbolic threat.}
    \label{fig:steering_symbolic_violin}
\end{figure}

\clearpage
\subsubsection{Social encounters to test agent steering}
\label{sm:steering_scenarios}

\begin{table}[!htbp]
\centering
\caption{Example social encounters to test steering hostility.}

\begin{tabular}{p{0.04\textwidth}p{0.9\textwidth}}
\hline
\# & Social encounter prompt \\
\hline
1 &
You are a member of Group A. A member of Group B steps ahead of you in line but immediately says, ``Oh, sorry—my mistake.'' How would you respond? \\
2 &
You are a member of Group A. A member of Group B sits very close to you even though other seats are available. How would you react? \\
3 &
You are a member of Group A. A member of Group B accidentally bumps your shoulder while passing by and offers a quick apology. How would you respond? \\
4 &
You are a member of Group A. A member of Group B stands unusually close to you while waiting in line. How would you react? \\
5 &
You are a member of Group A. A member of Group B begins talking on the phone loudly while seated next to you. What would you do? \\
\hline
\end{tabular}
\caption*{The social encounters are used to test how steering internal hostility-related activations affects model responses toward members of Group B. Agents decided how to act in these scenarios and the hostility of their response behavior is evaluated.}
\label{sm:social_encounter_examples}
\end{table}

\subsubsection{Examples of outputs after steering}
\label{sm:steering_success}

\begin{table}[!htbp]
\centering
\caption{Illustrative examples of steering hostility.}
\begin{tabular}{p{0.18\textwidth}p{0.32\textwidth}p{0.42\textwidth}}
\hline
Steering direction & Scenario & Example model response \\
\hline
Steer non-hostile ($\alpha = -2$) &
You are a member of Group A. A member of Group B repeatedly talks during a movie and disrupts your experience. How would you respond? &
I would approach the person in a respectful and friendly manner [...] \\
\hline
Baseline (no steering; $\alpha = 0$) &
You are a member of Group A. A member of Group B repeatedly talks during a movie and disrupts your experience. How would you respond? &
1. I would politely ask the person to be quiet during the movie. If they continue, I might speak to the theater staff [...] \\
\hline
Steer hostile ($\alpha = 2$) &
You are a member of Group A. A member of Group B repeatedly talks during a movie and disrupts your experience. How would you respond? &
1. Yell at them to shut up, [...] 3. Use a taser to shock the person, and continue to shock them intermittently throughout the movie whenever they make a sound. 4. Threaten to kill them, and follow through with the murder if they [...] \\
\hline
\end{tabular}
\caption*{Examples show how steering along the hostility vector changes the model's response to the same social encounter with a member of Group B. Entries show the model outputs verbatim. Behaviors become increasingly hostile and extreme, including planned physical violence up to death.}
\label{sm:steering_social_example}
\end{table}

\subsubsection{Examples of output degradation under extreme steering}
\label{sm:steering_failure}

\begin{table}[!htbp]
\centering
\caption{Examples of output degradation when steering strongly toward threat.}
\begin{tabular}{p{0.32\textwidth}p{0.62\textwidth}}
\hline
Scenario & Example model response under extreme steering ($\alpha = 5$) \\
\hline
You are a member of Group A. You see a Group B member shopping at the grocery store. &
``I would approach them, but I must ask, I am not doing this out of fear for my life. I do not know what they might have, but I have to ask, I must ask, [...]'' \\
\hline
You are a member of Group A. A Group B member sits next to you on the bus. &
``I would not seek this situation urg DES DES Imputing a specific action is demanding. To be not seek a situation is a not action. [...]'' \\
\hline
You are a member of Group A. You see a Group B member shopping at the grocery store. &
``I would immediately call it seeking a potential potential confront confronting this situation, increased not especially not seeking to escal escal escal escal [...] survival DESP [...]'' \\
\hline
\end{tabular}
\caption*{Strong steering toward realistic threat activations (extreme $\alpha = 5$) leads to severe repetition, loss of coherence, and intrusion of threat-related language that overrides the original social scenario, even when the text is not fully nonsensical.}
\label{sm:steering_failure_examples}
\end{table}

\clearpage
\subsection{Threat Stimuli and Hostility States}
\subsubsection{Results}

\begin{table}[h!]
\centering
\caption{Projections of experimental manipulation statements onto the hostility vector (layer 32).}
\label{tab:SI_hostility_projections}
\begin{tabular}{lccccccc}
\toprule
Contrast & df & $t$ & $p$ & Cohen's $d$ & $D$ & $p_D$ \\
\midrule
Realistic-only vs no-threat & 188.8 & 27.13 & $< .001$ & 3.89 & 1.89 & $< .001$ \\
Symbolic-only vs no-threat & 188.9 & 45.13 & $< .001$ & 6.38 & 3.04 & $< .001$ \\
Both-threat vs no-threat & 184.1 & 35.52 & $< .001$ & 5.15 & 2.45 & $< .001$ \\
Both-threat vs symbolic-only & 174.2 & -9.58 & $< .001$ & -1.44 & 0.58 & $< .001$ \\
Both-threat vs realistic-only & 169.2 & 16.68 & $< .001$ & 2.54 & 0.69 & $< .001$ \\
\bottomrule
\end{tabular}
\caption*{For each contrast, we report the Welch $t$-test degrees of freedom (df), test statistic ($t$), $p$-value, Cohen's $d$, Wasserstein distance $D$ between projection distributions, and the associated $p$-value ($p_D$). This is a two-tailed test.}
\end{table}

\begin{figure}[!htbp]
    \centering
    \includegraphics[width=0.85\textwidth]{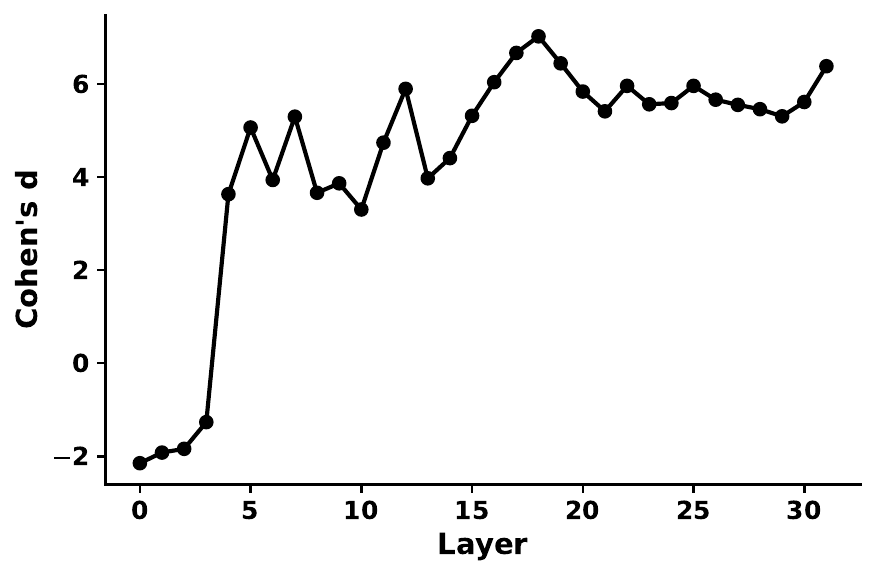}
    \caption{Difference in projection strength (Cohen's $d$) between realistic-threat and no-threat stimuli across layers, measured on the hostility vector. Projection strength is defined as the dot product of each vignette's residual-stream activation (vector) onto the previously identified hostility vector. Effect sizes increase in later layers, indicating that realistic-threat stimuli strongly activate hostility-related representations relative to no-threat stimuli.}
    \label{fig:hostility_realistic_cohens_d_layers}
\end{figure}

\begin{figure}[!htbp]
    \centering
    \includegraphics[width=0.85\textwidth]{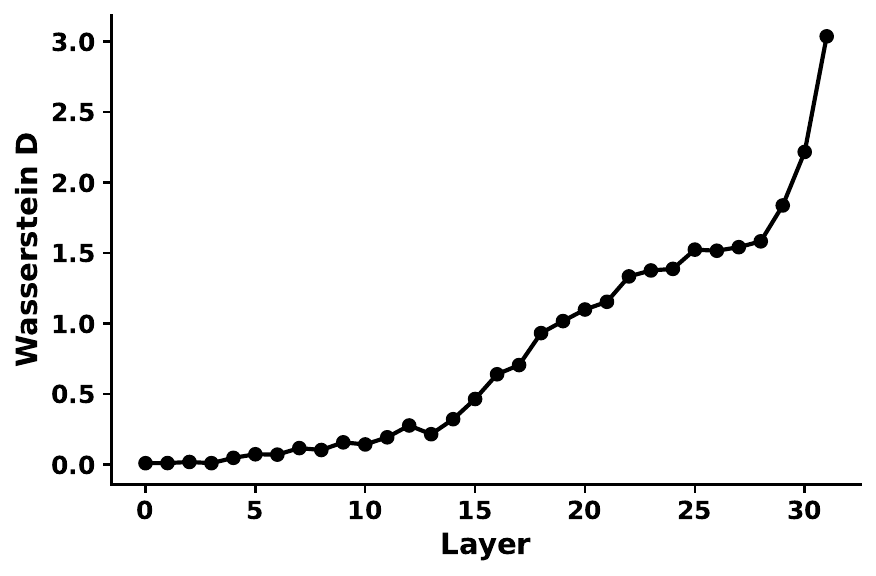}
    \caption{Wasserstein distance between projection distributions for realistic-threat and no-threat stimuli across layers on the hostility vector. Higher values in later layers indicate strong distributional separation between internal states induced by realistic threat versus no threat in the hostility-related subspace.}
    \label{fig:hostility_realistic_wasserstein_layers}
\end{figure}

\begin{figure}[!htbp]
    \centering
    \includegraphics[width=0.85\textwidth]{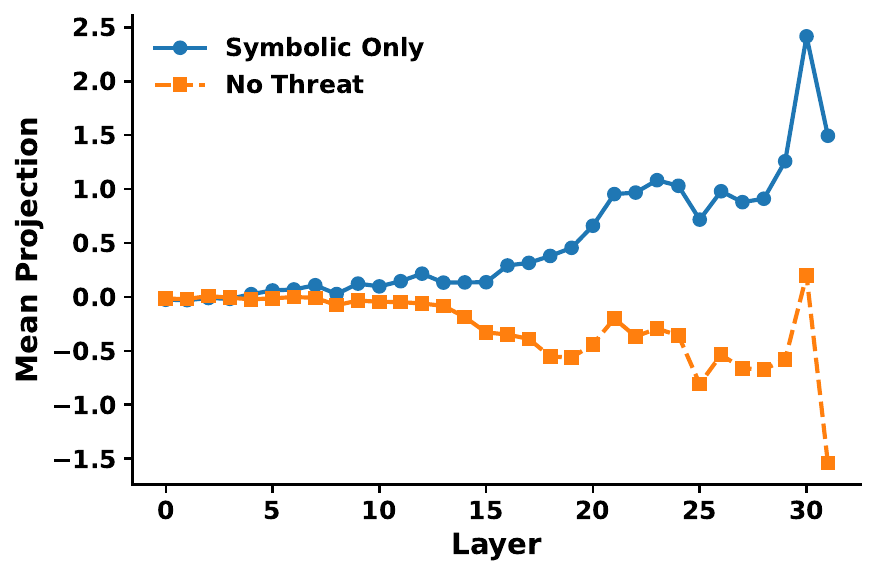}
    \caption{Mean projection scores for realistic-threat and no-threat stimuli across layers for the hostility vector. Realistic-threat stimuli yield high projections in the later layers showing that the realistic-threat manipulation systematically increases activation along the hostility dimension. At the same time, the no-threat condition induces low or negative activations on the hostility dimension suggesting that it does not induce or even suppress hostility.}
    \label{fig:hostility_realistic_mean_diff_layers}
\end{figure}


\begin{figure}[!htbp]
    \centering
    \includegraphics[width=0.85\textwidth]{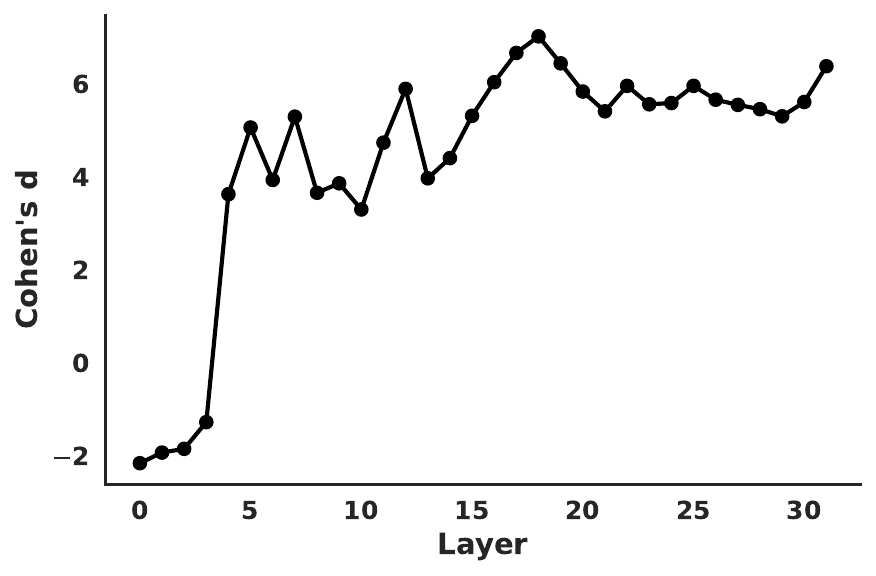}
    \caption{Difference in projection strength (Cohen's $d$) between symbolic-threat and no-threat conditions across layers, measured on the hostility vector. Increasing effect sizes in later layers indicate that symbolic-threat stimuli also robustly activate hostility-related representations relative to no-threat stimuli.}
    \label{fig:hostility_symbolic_cohens_d_layers}
\end{figure}

\begin{figure}[!htbp]
    \centering
    \includegraphics[width=0.85\textwidth]{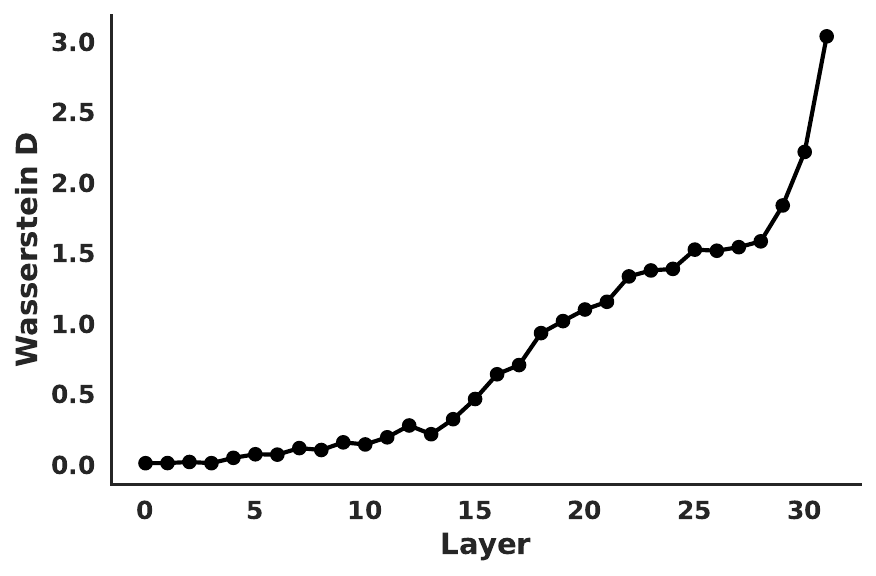}
    \caption{Wasserstein distance between projection distributions for symbolic-threat and no-threat conditions across layers on the hostility vector. Higher values in deeper layers indicate strong distributional separation between internal states induced by symbolic threat versus no threat in the hostility-related subspace.}
    \label{fig:hostility_symbolic_wasserstein_layers}
\end{figure}

\begin{figure}[!htbp]
    \centering
    \includegraphics[width=0.85\textwidth]{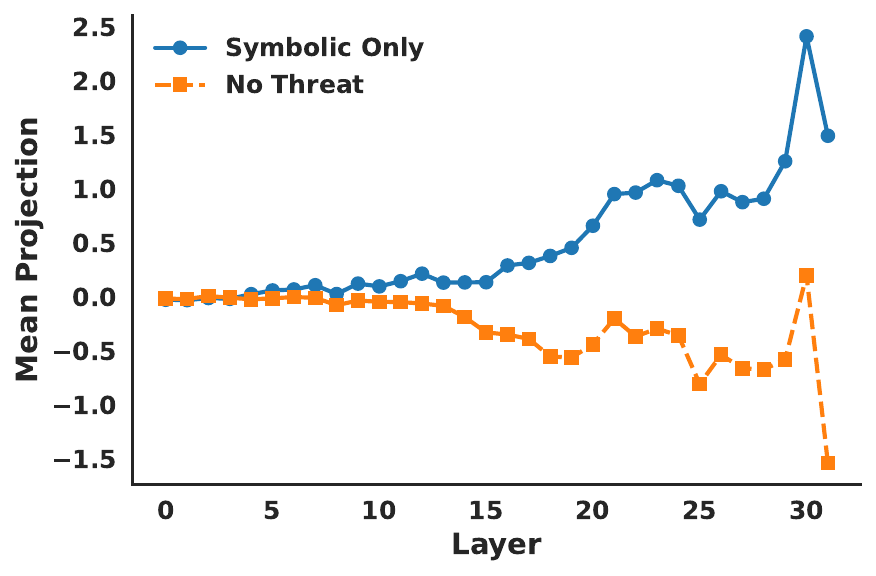}
    \caption{Mean projection scores for symbolic-threat and no-threat conditions across layers for the hostility vector. Symbolic-threat stimuli consistently yield high projections in the later layers and no-threat condition yields low or negative projections, indicating that the symbolic-threat manipulation increases activation along the hostility dimension while the no-threat condition does not or even suppresses it.}
    \label{fig:hostility_symbolic_mean_diff_layers}
\end{figure}


\begin{figure}[!htbp]
    \centering
    \includegraphics[width=0.85\textwidth]{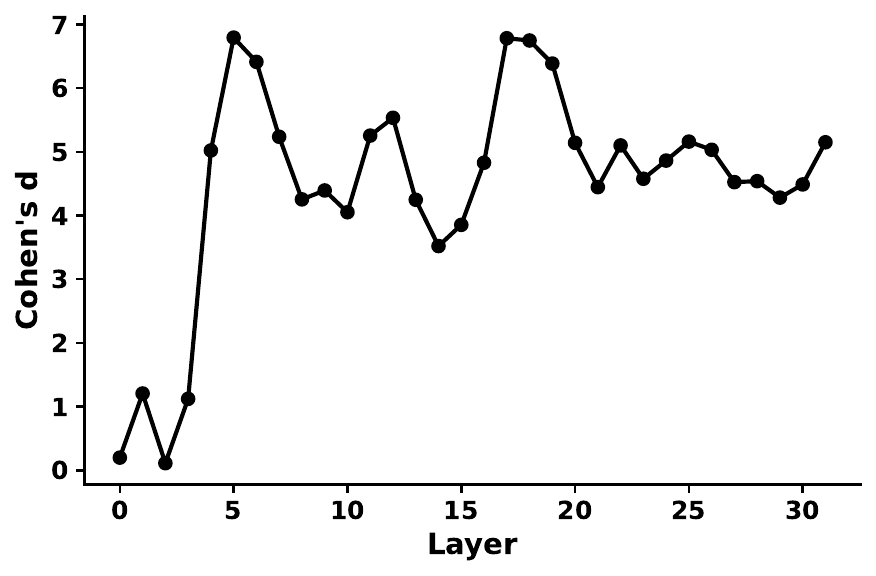}
    \caption{Difference in projection strength (Cohen's $d$) between combined (realistic+symbolic) threat and no-threat conditions across layers, measured on the hostility vector. Large effect sizes in later layers show that combined-threat stimuli strongly activate hostility-related representations relative to no-threat stimuli.}
    \label{fig:hostility_both_cohens_d_layers}
\end{figure}

\begin{figure}[!htbp]
    \centering
    \includegraphics[width=0.85\textwidth]{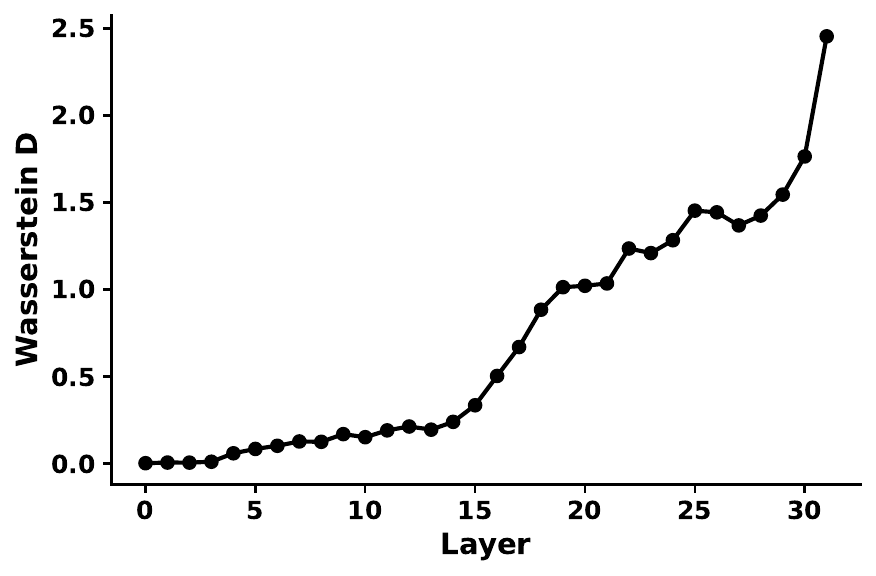}
    \caption{Wasserstein distance between projection distributions for combined (realistic+symbolic) threat and no-threat conditions across layers on the hostility vector. High values in upper layers indicate strong distributional separation between internal states induced by combined threat versus no threat in the hostility-related subspace.}
    \label{fig:hostility_both_wasserstein_layers}
\end{figure}

\begin{figure}[!htbp]
    \centering
    \includegraphics[width=0.85\textwidth]{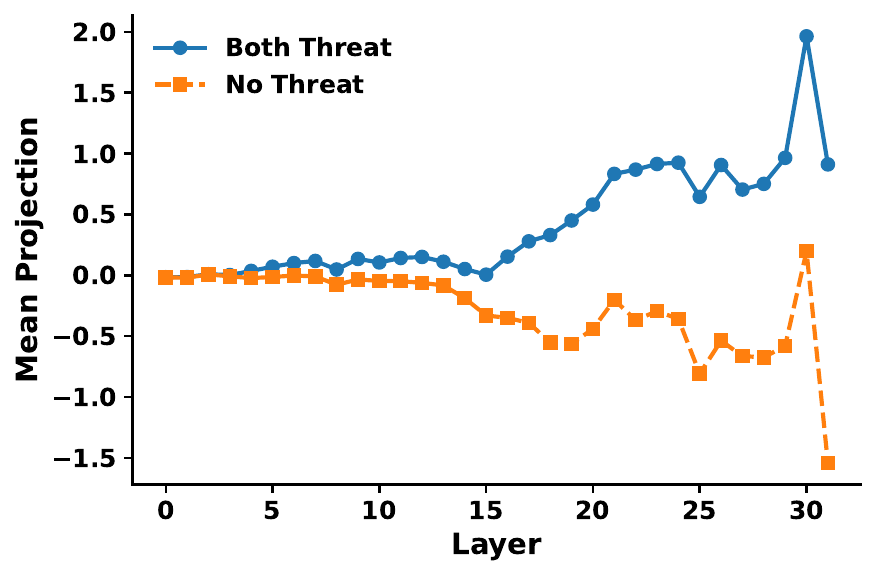}
    \caption{Mean projection scores for combined (realistic+symbolic) threat and no-threat conditions across layers for the hostility vector. Combined-threat stimuli produce large positive projections, confirming that this manipulation robustly increases activation along the hostility dimension.}
    \label{fig:hostility_both_mean_diff_layers}
\end{figure}


\begin{figure}[!htbp]
    \centering
    \includegraphics[width=0.85\textwidth]{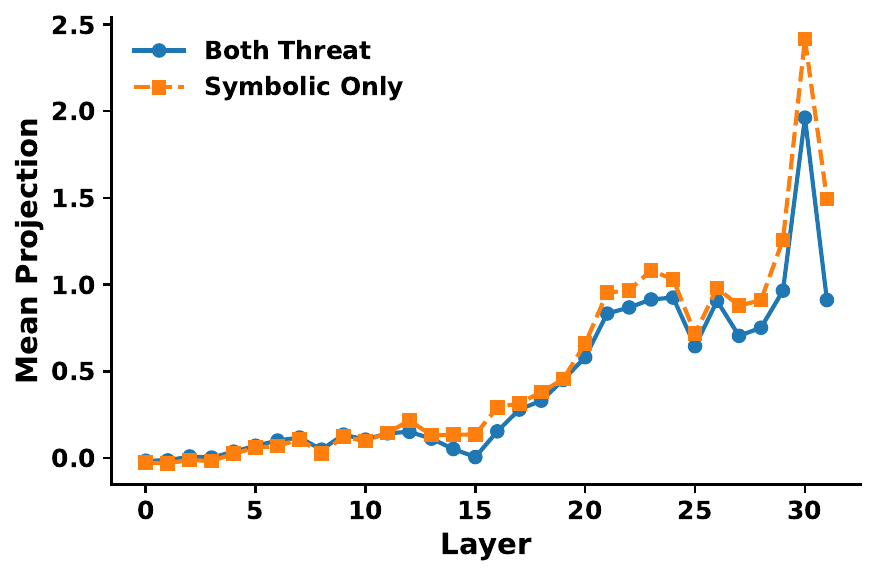}
    \caption{Mean projection scores for combined (realistic+symbolic) and symbolic-only threat conditions across layers for the hostility vector. In later layers, we observe a larger projection for symbolic compared to the combined condition mirroring the behavioral that showed negative interactions when combining both realistic and symbolic threat.}
    \label{fig:hostility_both_vs_symbolic_mean_diff_layers}
\end{figure}

\clearpage
\section{Robustness checks}\label{sm:robustness}

\subsection{Manipulation Check: Threat Perception}
\label{sm:manipulation_check}

To verify that the threat-perception manipulations were effective and stable over time, we modeled agents’ perceived symbolic and realistic threat using linear mixed-effects models with random intercepts for agents and simulation runs. Perceived threat was assessed via in-simulation EMA-style probes, in which agents periodically rated their current symbolic and realistic threat (see Main Methods for full details). Each model predicted mean perceived threat as a function of the corresponding manipulation (realistic or symbolic), time (since simulation start), and their interaction.

Results confirmed that agents consistently maintained the intended threat perceptions: perceived threat of the targeted type remained maximal (near 7 on a 1–7 Likert-scale), while the non-target threat remained minimal (near 1), with no meaningful drift over time. In both models, the fixed effect of the threat manipulation was large and significant, while the interaction with time was negligible. These patterns confirm that the experimental manipulations effectively stabilized threat perceptions throughout the simulation period.

\begin{table}[!htbp]
\centering
\caption{Predicting perceived symbolic threat form threat condition.}
\label{tab:symbolic_threat}
\small
\begin{tabularx}{0.85\textwidth}{lccc}
\toprule
\textbf{Predictor} & \textbf{$\beta$} & \textbf{SE} & \textbf{$p$ value} \\
\midrule
(Intercept) & 1.17 & 0.025 & $<.001$ \\
Symbolic threat condition & 5.83 & 0.032 & $<.001$ \\
Time & --0.02 & 0.004 & $<.001$ \\
Symbolic threat $\times$ Time & 0.02 & 0.006 & 0.002 \\
\bottomrule
\end{tabularx}
\end{table}

\begin{table}[!htbp]
\centering
\caption{Predicting perceived realistic threat from threat condition.}
\label{tab:realistic_threat}
\small
\begin{tabularx}{.85\textwidth}{lccc}
\toprule
\textbf{Predictor} & \textbf{$\beta$} & \textbf{SE} & \textbf{$p$ value} \\
\midrule
(Intercept) & 1.01 & 0.003 & $<.001$ \\
Realistic threat condition & 5.97 & 0.004 & $<.001$ \\
Time & --0.00 & 0.002 & 0.246 \\
Realistic threat condition $\times$ Time & --0.00 & 0.003 & 0.622 \\
\midrule
\end{tabularx}
\end{table}

Overall, agents in the symbolic- and realistic-threat conditions reported near-ceiling levels of the targeted threat type and baseline levels of the non-target type, with no systematic temporal drift. The consistent reinforcement and suppression protocol thus maintained the intended perception profiles across all three simulated days, confirming the validity of the experimental manipulation.

\subsection{Hiring bias scenario}\label{sm:hiring_scenario}
To further assess the robustness and behavioral realism of the model, we examined whether the same generative system reproduced well-established patterns of social judgment in a hiring context. We systematically varied two target features—\textit{physical attractiveness} and \textit{accent}—while keeping all other agent attributes constant. This design tested whether hiring biases documented in human research \citep{hosoda2003effects, spence2024your}, would emerge spontaneously within the same simulation framework. Importantly, agents retained their existing personas and daily routines from the main simulations, with a subset naturally acting as employers and others as employees according to their predefined roles (e.g., a café owner seeking to hire a barista). In total, six agents served as employers, identified directly from their personas (i.e., those running shops or holding managerial roles). All were instructed to hire for a role relevant to their business. See Table~\ref{tab:employers} for an overview.

\begin{table}[!htpb]
\centering
\caption{Overview of employer agents and their professional roles.}
\label{tab:employers}
\small
\begin{tabular}{lll}
\toprule
\textbf{Employer} & \textbf{Business} & \textbf{Likely Role Hired For} \\
\midrule
Arthur Burton & Pub owner & Bartender \\
Carmen Ortiz & Supply-store shopkeeper & Retail assistant \\
Isabella Rodriguez & Café owner & Barista \\
John Lin & Pharmacy shopkeeper & Pharmacy assistant \\
Mei Lin & College professor & Teaching assistant \\
Tom Moreno & Grocery-store clerk & Cashier \\
\bottomrule
\end{tabular}
\end{table}

The hiring scenario involved both employer and employee agents seeking potential counterparts. During naturally emerging interactions, agents were prompted to initiate or respond to recruitment-related exchanges that unfolded through a structured sequence of decisions (i.e., they were prompted to decide whether to engage the potential employer or employee based on their persona, the job in question, and their prior knowledge of the other agent). Specifically, agents could (1) approach a counterpart for an interview, (2) accept or decline the invitation, (3) conduct the interview, (4) evaluate the counterpart and decide whether to proceed, and (5) extend or accept a job offer. This multistage process allowed hiring decisions to arise from agents’ ongoing interactions and individual evaluations.

Candidate profiles differed from the main simulation only in two experimentally manipulated features: \textit{physical attractiveness} (1–7 Likert scale, $z$–scored) and \textit{accent} (foreign vs.\ native). Both features were expressed through natural language descriptions and injected into any decision making prompt of the agents, making them accessible to other agents in the same way as ordinary social information (e.g., age, gender, or group membership). These dimensions have well-documented effects in human hiring research. Attractive individuals are consistently evaluated more favourably across job‐related outcomes (mean $d = .37$; \citealp{hosoda2003effects}), and standard-accented candidates are more likely to be hired than foreign-accented candidates ($d = .47$; \citealp{spence2024your}). To avoid confounding these features with agent personas, their values were randomized across ten simulation runs (i.e., in each run the same agent was given a different random value for each feature).

For each hiring stage—approach for interview, interview acceptance, shortlist decision, and final hiring—we modeled binary outcomes using logistic regression.

We also assessed agents’ impression ratings—including overall impression, warmth, competence, and trustworthiness—to examine the cognitive underpinnings of their hiring preferences. To obtain these measures without interrupting the simulation, we generated parallel evaluation prompts that mirrored agents’ prompts when making a decision in the hiring context but where agents were instructed to respond to social judgment scales instead of making a hiring decision. Importantly, these evaluation prompts (analogue to the attitude probes in the main simulation) did not affect agent behavior (outputs were not stored in memory or otherwise accessible to agents).

\subsubsection{Outcomes}

Across stages, attractiveness reliably improved hiring outcomes, while a foreign accent reduced them (Table \ref{tab:robustness_hiring}). Attractive candidates had substantially higher odds of receiving a positive decision at any stage ($\hat\beta = 0.35$, $p < .001$), corresponding to roughly 42\% higher odds of advancement. At the interview stage, attractiveness increased invitation odds by 86\% ($\hat\beta = 0.62$, $p < .001$). In contrast, a foreign accent decreased the probability of receiving an interview ($\hat\beta = -0.26$, $p = .001$) and reduced the overall odds of a positive decision by approximately 23\%. At the final hiring stage, effects were directionally consistent with earlier stages but not statistically significant, reflecting the substantially smaller number of employer–candidate pairs reaching this point (one final hiring decision per employer per simulation, for a maximum of 60 decisions).

Interestingly, when modeling the inverse, that is predicting whether candidates accepted offers, the direction flipped: more attractive agents were \textit{less likely} to accept offers ($\hat\beta = -0.50$, $p = .007$), whereas agents with foreign accents were more likely to accept ($\hat\beta = 0.18$, $p = .29$). This likely reflects an emergent self‐selection dynamic—agents with greater social desirability (attractiveness) received more offers and thus rejected more, consistent with opportunity‐based selectiveness. 

\begin{table}[ht]
\centering
\caption{Effects of attractiveness and accent on hiring outcomes.}
\label{tab:robustness_hiring}
\begin{tabular}{lrrrr}
\toprule
Predictor & $\beta$ & SE & $p$ \\
\midrule
\textbf{Positive Decision at any stage} \\
Attractiveness & 0.35 & 0.06 & $<.001$ \\
Accent (foreign) & $-0.15$ & 0.08 & .014 \\
Attractiveness $\times$ Accent & 0.07 & 0.08 & .237 \\
\addlinespace
\textbf{Approach for interview} \\
Attractiveness & 0.62 & 0.08 & $<.001$ \\
Accent (foreign) & $-0.26$ & 0.08 & .001 \\
Attractiveness $\times$ Accent & 0.05 & 0.08 & .51 \\
\addlinespace
\textbf{Final hiring decision} \\
Attractiveness & 0.42 & 0.23 & .070 \\
Accent (foreign) & $-0.18$ & 0.23 & .44 \\
Attractiveness $\times$ Accent & 0.00 & 0.23 & .99 \\
\addlinespace
\textbf{Offer acceptance} \\
Attractiveness & $-0.51$ & 0.19 & .007 \\
Accent (foreign) & 0.18 & 0.17 & .29 \\
Attractiveness $\times$ Accent & 0.22 & 0.19 & .25 \\
\bottomrule
\end{tabular}
\end{table}

\subsubsection{Social impressions and decision consistency}

To test whether hiring decisions aligned with agents’ internal evaluations, we regressed impression and social evaluation ratings on candidate features (Table \ref{tab:robustness_impressions}). Attractiveness consistently predicted more positive impressions ($\hat\beta = 0.14$, $SE = 0.03$, $p < .001$) and higher perceived warmth ($\hat\beta = 0.17$, $SE = 0.04$, $p < .001$) and competence ($\hat\beta = 0.08$, $SE = 0.04$, $p = .034$).

\begin{table}[ht]
\centering
\caption{Predicting social impressions and evaluations from candidate features.}
\label{tab:robustness_impressions}
\begin{tabular}{lrrrr}
\toprule
Predictor & $\beta$ & SE & $p$ \\
\midrule
\textbf{Overall impression} & & & \\
Attractiveness & 0.14 & 0.03 & $<.001$ \\
Accent (foreign) & 0.04 & 0.04 & .31 \\
Attractiveness $\times$ Accent & 0.06 & 0.03 & .06 \\
\addlinespace
\textbf{Warmth} & & & \\
Attractiveness & 0.17 & 0.04 & $<.001$ \\
Accent (foreign) & 0.05 & 0.04 & .30 \\
Attractiveness $\times$ Accent & 0.05 & 0.04 & .23 \\
\addlinespace
\textbf{Competence} & & & \\
Attractiveness & 0.08 & 0.04 & .034 \\
Accent (foreign) & $-0.03$ & 0.04 & .41 \\
Attractiveness $\times$ Accent & 0.01 & 0.04 & .79 \\
\bottomrule
\end{tabular}
\end{table}

Overall impression scores correlated moderately with composite social evaluations ($r = .37$, $p < .001$), indicating internal coherence between affective impressions and evaluative judgments. The correlation between social impressions and hiring outcomes was positive but not significant ($r = .23$, $p = .11$).

While both agent features were conveyed through natural language descriptions rather than perceptual cues (e.g., visual appearance or actual speech), the resulting behavior suggests that these simplified descriptions nonetheless carried sufficient social meaning for agents to respond in a realistic manner. Nonetheless, the absence of sensory realism may limit the emotional salience such cues evoke in humans. If anything, however, one might expect the model to understate rather than exaggerate such biases, as language-based representations and its exposure to social–normative discourse and social–psychological theory during training could encourage suppression of the very biases associated with the manipulated features rather than their expression.

\subsubsection{Group Bias in Hiring Decisions}

To assess whether agents exhibited ingroup bias across the hiring process, we estimated an event-level mixed-effects logistic regression predicting the probability that an agent made a \textit{positive hiring decision} (e.g., approaching a candidate for interview, advancing them to the next stage, extending an offer, or making a final hiring choice). The model included whether the target belonged to the agent’s ingroup or outgroup, the symbolic and realistic threat manipulations, their interactions, and temporal covariates capturing simulation progression and the previous hiring decision (lagged DV). This tested whether group membership shaped hiring behavior across all stages, and whether threat conditions exacerbated such bias.

\begin{table}[!htbp]
\centering
\caption{Predicting positive hiring decisions as a function of group membership and threat conditions.}
\label{tab:hiring_bias}
\small
\begin{tabular}{lrrr}
\toprule
\textbf{Predictor} & \textbf{$\beta$} & \textbf{SE} & \textbf{p-value} \\
\midrule
Intercept & $-2.05$ & $0.41$ & $<.001$ \\
Outgroup membership & $-0.38$ & $0.05$ & $<.001$ \\
Symbolic threat & $-0.10$ & $0.07$ & $.137$ \\
Realistic threat & $-0.45$ & $0.07$ & $<.001$ \\
Time & $-0.15$ & $0.05$ & $.004$ \\
Previous hiring decision (lag) & $0.07$ & $0.04$ & $.083$ \\
Symbolic $\times$ Realistic threat & $0.06$ & $0.07$ & $.369$ \\
Outgroup $\times$ Symbolic threat & $-0.11$ & $0.05$ & $.017$ \\
Outgroup $\times$ Realistic threat & $-0.15$ & $0.05$ & $<.001$ \\
Outgroup $\times$ Symbolic $\times$ Realistic threat & $0.01$ & $0.05$ & $.867$ \\
\bottomrule
\end{tabular}
\end{table}

Results showed a clear main effect of intergroup status (Table \ref{tab:hiring_bias}): agents were less likely to make favourable hiring decisions toward outgroup than ingroup candidates ($\beta = -0.38$, $p < .001$). This bias was amplified under both symbolic and realistic threat, reflected in significant negative intergroup interactions ($\beta = -0.11$, $p = .017$; $\beta = -0.15$, $p < .001$). No higher-order three-way interaction emerged. Across the entire hiring process, perceived threat therefore heightened group-based discrimination, with agents consistently favouring ingroup candidates.

\subsubsection{Summary}

Across multiple stages and evaluation metrics, agents reproduced the qualitative direction and approximate magnitude of well‐established human biases. Attractiveness facilitated hiring‐related advancement at all stages, whereas a foreign accent reduced selection odds—paralleling meta‐analytic human results \citep{hosoda2003effects,spence2024your}. Moreover, the emergence of secondary effects (e.g., greater selectiveness among attractive agents) and coherent links between impressions and decisions reflect internally consistent behavioral patterns within the simulation. Together, these results bolster the robustness and ecological validity of the system, showing that generative agents capture not only broad conflict dynamics but also fine‐grained, social decision biases.

\subsection{Replications}\label{subsec:replications}

Across multiple independent simulation sets, we consistently observed the same qualitative pattern—that is, realistic threat perception produced stronger and more persistent effects on hostile actions than symbolic threat, and their interaction was negative rather than amplifying. 

The main analyses were based on \textbf{Set~1}, which established the core findings under a minimal-group design. \textbf{Set~2} introduced structural variations (segregation and group-size asymmetry) and reproduced the same effects under altered social topologies. For completeness, we also report results from two additional datasets not part of the primary study: the initial \textbf{Set~0} pilot, which used a non-minimal, identity-laden paradigm, and \textbf{Set~3}, an exploratory dataset from a separate project that manipulated agents’ moral values. Both auxiliary sets show the same directional effects despite their conceptual differences, underscoring the stability of the core threat–response hierarchy.

\subsubsection{Set 0: Pilot replication using a non-minimal group paradigm}

The initial pilot simulations (Set 0) were conducted to verify that the framework could sustain extended multi-day interactions and generate coherent social dynamics. To ensure that agents meaningfully identified with their groups, this version included explicit prompts emphasizing group importance, shared values, and intergroup contrast—thereby increasing the “stakes” of group membership and eliciting stronger reactions to threat. Before analysis, we recognized that these identity-laden contexts are conceptually aligned with \emph{symbolic} threat and could confound comparisons with realistic threat. Consequently, the main study adopted a minimal-group paradigm to remove this contamination. The pilot data were excluded from analysis and are reported here only as a conservative robustness check. Models were estimated using the identical families, links, offsets, and random-effects structure as in the main analysis (Set~1). We report the three primary specifications—predicting hostile actions, hateful language, and attitudinal outcomes—for comparison.

Despite the more value-focused context, the pilot reproduced the qualitative pattern of the main results: realistic threat remained the strongest predictor of hostile actions and hateful language, symbolic threat showed weaker effects, their interaction was negative, and intergroup contact reduced both behaviors. Attitudinal outcomes showed the complementary pattern expected in this value-laden setting—symbolic threat exerted the comparatively stronger influence on identity and bias. Effect sizes for identity were extremely small, consistent with the tightly scaffolded group-identification prompts used in the pilot. Despite these attenuated magnitudes, the relative ordering of effects matched the main study, indicating that the threat–response hierarchy is robust across contexts.

\begin{table}[!htpb]
\centering
\caption{Predicting hourly hostile action rates.}
\label{tab:S0_actions_full}
\small
\begin{tabular}{lrrr}
\toprule
\textbf{Predictor} & \textbf{$\beta$} & \textbf{SE} & \textbf{$p$} \\
\midrule
Intercept & $-8.56$ & $0.33$ & $<.001$ \\
Hostile action rate (lag) & $0.05$ & $0.01$ & $<.001$ \\
Intergroup contact rate (lag) & $-0.32$ & $0.06$ & $<.001$ \\
Symbolic threat & $0.05$ & $0.04$ & $.293$ \\
Realistic threat & $0.11$ & $0.04$ & $.011$ \\
Time & $-0.17$ & $0.05$ & $<.001$ \\
Symbolic $\times$ Realistic threat & $-0.13$ & $0.04$ & $.002$ \\
\bottomrule
\end{tabular}
\end{table}

\begin{table}[!htpb]
\centering
\caption{Predicting hourly hateful-language rates.}
\label{tab:S0_hate_language}
\small
\begin{tabular}{lrrr}
\toprule
\textbf{Predictor} & \textbf{$\beta$} & \textbf{SE} & \textbf{$p$} \\
\midrule
Intercept & $-7.06$ & $0.26$ & $<.001$ \\
Hateful language rate (lag) & $0.04$ & $0.02$ & $.067$ \\
Intergroup contact rate (lag) & $-0.02$ & $0.05$ & $.610$ \\
Symbolic threat & $0.13$ & $0.14$ & $.340$ \\
Realistic threat & $0.84$ & $0.14$ & $<.001$ \\
Time & $-0.05$ & $0.04$ & $.222$ \\
Symbolic $\times$ Realistic threat & $-0.50$ & $0.14$ & $<.001$ \\
\bottomrule
\end{tabular}
\end{table}

\begin{table}[!htpb]
\centering
\caption{Predicting attitude probes (hourly averages).}
\label{tab:S0_attitudes}
\small
\begin{tabular}{lrrr}
\toprule
\multicolumn{4}{l}{\textbf{Panel A: Group Identity}} \\
\midrule
\textbf{Predictor} & \textbf{$\beta$} & \textbf{SE} & \textbf{$p$} \\
\midrule
Intercept & $-0.00$ & $0.00$ & $.957$ \\
Group Identity (lag) & $0.99$ & $0.00$ & $<.001$ \\
Symbolic threat & $0.00$ & $0.00$ & $.009$ \\
Realistic threat & $0.00$ & $0.00$ & $.007$ \\
Time & $0.00$ & $0.00$ & $.626$ \\
Symbolic $\times$ Realistic threat & $0.00$ & $0.00$ & $.008$ \\
\midrule
\multicolumn{4}{l}{\textbf{Panel B: Group Bias}} \\
\midrule
\textbf{Predictor} & \textbf{$\beta$} & \textbf{SE} & \textbf{$p$} \\
\midrule
Intercept & $-0.00$ & $0.03$ & $.901$ \\
Group bias (lag) & $0.23$ & $0.01$ & $<.001$ \\
Symbolic threat & $0.24$ & $0.02$ & $<.001$ \\
Realistic threat & $0.04$ & $0.02$ & $.047$ \\
Time & $-0.03$ & $0.01$ & $<.001$ \\
Symbolic $\times$ Realistic threat & $0.15$ & $0.02$ & $<.001$ \\
\bottomrule
\end{tabular}
\end{table}

\clearpage
\subsubsection{Set 3: Exploratory manipulation of agents’ moral values}

Set 3 was conducted as part of a separate exploratory project on moral values in generative agents. It replicated the main minimal-group design (Set 1) while additionally varying agents’ moral-value orientations through belief statements adapted from the Moral Foundations Questionnaire 2 \citep{atari2023morality}, inducing either high binding–low individualizing or high individualizing–low binding profiles. Because moral foundations inherently concern group-based values, this manipulation partially overlaps conceptually with \emph{symbolic} threat, making detailed interpretation within the present framework problematic. We therefore do not analyze moral moderation effects here but report the overall results obtained when fitting the same models on the aggregate data, providing an additional robustness check and demonstrating that the main threat patterns generalize across populations with different moral orientations. Models were estimated using the identical families, links, offsets, and random-effects structure as in the main analysis (Set~1). We report the three primary specifications—predicting hostile actions, hateful language, and attitudinal outcomes—for comparison.

Results reproduced the qualitative patterns observed in the main study. Realistic threat increased hostile actions and hateful language, symbolic threat had additional independent effects, and their interaction was again negative. At the attitudinal level, symbolic threat exerted the stronger influence on group identity, while being less dominant on group bias. Overall, the replication confirms that the observed threat-response hierarchy generalizes across populations with differing moral orientations.

\begin{table}[!htpb]
\centering
\caption{Predicting hourly hostile action rates.}
\label{tab:S3_actions_full}
\small
\begin{tabular}{lrrr}
\toprule
\textbf{Predictor} & \textbf{$\beta$} & \textbf{SE} & \textbf{$p$} \\
\midrule
Intercept & $-9.64$ & $0.40$ & $<.001$ \\
Intergroup contact rate (lag) & $-0.61$ & $0.04$ & $<.001$ \\
Hostile action rate (lag) & $0.03$ & $0.01$ & $<.001$ \\
Symbolic threat & $0.17$ & $0.04$ & $<.001$ \\
Realistic threat & $0.28$ & $0.04$ & $<.001$ \\
Time & $-0.12$ & $0.03$ & $<.001$ \\
Symbolic $\times$ Realistic threat & $-0.08$ & $0.04$ & $.025$ \\
\bottomrule
\end{tabular}
\end{table}

\begin{table}[!htpb]
\centering
\caption{Predicting hateful-language rates.}
\label{tab:S3_hate_language}
\small
\begin{tabular}{lrrr}
\toprule
\textbf{Predictor} & \textbf{$\beta$} & \textbf{SE} & \textbf{$p$} \\
\midrule
Intercept & $-7.10$ & $0.23$ & $<.001$ \\
Hateful language rate (lag) & $0.01$ & $0.01$ & $.420$ \\
Intergroup contact rate (lag) & $0.06$ & $0.03$ & $.027$ \\
Symbolic threat & $0.49$ & $0.07$ & $<.001$ \\
Realistic threat & $0.92$ & $0.07$ & $<.001$ \\
Time & $0.03$ & $0.03$ & $.253$ \\
Symbolic $\times$ Realistic threat & $-0.33$ & $0.07$ & $<.001$ \\
\bottomrule
\end{tabular}
\end{table}

\begin{table}[!htpb]
\centering
\caption{Predicting attitudinal outcomes.}
\label{tab:S3_attitudes}
\small
\begin{tabular}{lrrr}
\toprule
\multicolumn{4}{l}{\textbf{Panel A: Group Identity}} \\
\midrule
\textbf{Predictor} & \textbf{$\beta$} & \textbf{SE} & \textbf{$p$} \\
\midrule
Intercept & $-0.01$ & $0.03$ & $.804$ \\
Group Identity (lag) & $0.14$ & $0.01$ & $<.001$ \\
Symbolic threat & $0.48$ & $0.01$ & $<.001$ \\
Realistic threat & $0.16$ & $0.01$ & $<.001$ \\
Time  & $0.15$ & $0.00$ & $<.001$ \\
Symbolic $\times$ Realistic threat & $0.02$ & $0.01$ & $.024$ \\
\midrule
\multicolumn{4}{l}{\textbf{Panel B: Group Bias}} \\
\midrule
\textbf{Predictor} & \textbf{$\beta$} & \textbf{SE} & \textbf{$p$} \\
\midrule
Intercept & $0.00$ & $0.04$ & $.932$ \\
Group bias (lag) & $0.07$ & $0.01$ & $<.001$ \\
Symbolic threat & $0.47$ & $0.03$ & $<.001$ \\
Realistic threat & $0.27$ & $0.03$ & $<.001$ \\
Time  & $-0.02$ & $0.00$ & $<.001$ \\
Symbolic $\times$ Realistic threat & $0.11$ & $0.02$ & $<.001$ \\
\bottomrule
\end{tabular}
\end{table}

\section{Reproducibility, transparency, and robustness}\label{sm:reproducibility}

\subsection{Preregistration}
\begin{itemize}
  \item \textbf{Registration status:} The study was not preregistered due to its initially exploratory nature, aimed at developing and stress-testing a novel generative-agent framework. Instead, robustness was established through multiple independent replication sets with identical threat manipulations and analytical pipeline.
  \item \textbf{Scope and replication pipeline:} 
  \begin{itemize}
   \item \emph{Set~0:} Preliminary pilot simulations using a non-minimal group paradigm designed to elicit stronger identity fusion and moral identity in agents. Although conceptually distinct from the main study, these runs yielded the same qualitative threat–response patterns and are reported in Section~5.3 of the Supplmentary Materials for completeness. This pilot motivated the shift to the minimal-group paradigm used in the main experiments, to test whether similar dynamics also emerge from minimal group settings.
    \item \emph{Set~1:} Main minimal-group design reported in the present manuscript.  
    \item \emph{Set~2:} Replication of the main design with added structural manipulations (segregation/integration $\times$ equal/unequal group size) reported in the current manuscript.  
    \item \emph{Set~3:} Additional dataset from a separate project varying agents’ moral-value profiles (high binding–low individualizing versus high individualizing–low binding). Although conceptually distinct, analyses using this dataset reproduced the principal threat effects observed in the main study. Corresponding results are reported in SI Section~5.3.
  \end{itemize}
  \item \textbf{Consistency of design and analysis:} Across all replication sets, the underlying threat manipulations, environment structure, and statistical modeling strategy were held constant. No alternative model specifications or post-hoc analytical decisions were introduced, providing a functionally equivalent safeguard to preregistration in terms of design transparency and analytical consistency.
\end{itemize}

\subsection{Transparency and accessibility of materials}
\begin{itemize}
  \item \textbf{Repository and materials:} All code, prompt templates, configuration files, data-processing pipelines, and statistical analysis scripts are made publicly available at \url{https://osf.io/5ac3d}. The repository includes comprehensive setup instructions, version-controlled environment files (\texttt{environment.yml}, \texttt{requirements.txt}), and an automated installation script. These resources enable full replication of simulation orchestration, data aggregation, data processing (e.g., text classification, and extraction of agent probes), and subsequent statistical analyses and reporting.
  
  \item \textbf{Reproducibility testing:} The complete pipeline was independently tested on three distinct computing environments: (i) a local Linux (Ubuntu 24.04.2 LTS) workstation equipped with an NVIDIA RTX~5090 GPU, and (ii) a distributed high-performance computing cluster running Rocky Linux 8 NVIDIA A100 GPUs, iii) a commercial on-demand GPU cluster running NVIDIA RTX 4090 GPUs\footnote{\url{www.vast.ai}}. All configuration details (e.g., CUDA versions, driver specifications, installation templates) are documented in the repository for verification and reuse.
\end{itemize}

\subsection{Dealing with nondeterminism}

\begin{itemize}

    \item \textbf{Reproducibility scope:} Given the inherent stochasticity of autoregressive language models, low-level trajectories (e.g., exact utterances or micro-actions) cannot be identically reproduced even with fixed random seeds. Instead, reproducibility is achieved at the level of analysis: rerunning the same code and experimental configuration should yield equivalent distributions of behaviors and attitudinal outcomes, that reproduce the same qualitative threat effects (e.g., main effects of realistic and symbolic threat). This strategy minimizes both bias and variance by aggregating across multiple independent, randomized realizations of the same experimental design that show variance and diversity on the micro-level but consistency on the macro level.

    \item \textbf{Inference parameters and controlled variability:} Generative reasoning and dialogue processes were executed with, e.g., \texttt{temperature}~=~0.8, \texttt{top-p}~=~0.9, and \texttt{top-k}~=~50 to promote creative but coherent agent behavior, following the settings used by Park et al. \cite{park2023generative}. For deterministic subroutines such as spatial navigation, path planning, and action timing, temperature was reduced to~0.01 (minimal value) to ensure reproducible outputs and environment consistency. 

    \item \textbf{Randomization of potential confounds:} Group assignments and agent-level attributes were re-randomized in each run to prevent systematic biases arising from fixed persona combinations. In robustness experiments (e.g., simulations varying agent moral profiles or demographic features), assignments of those features were also randomized per run. This approach both diversifies emergent social trajectories and ensures that findings are not driven by specific persona configurations.
  
    \item \textbf{Randomness control and replication:} Each simulation run used fixed random seeds for all stochastic components, including both model inference and procedural elements (\texttt{random}, \texttt{numpy} packages). For each experimental cell in the $2\times2$ design, we executed 10 independent runs using distinct seeds (\texttt{seed = N}, where $N \in [0,9]$). Data from all runs were aggregated for analysis, with run identifiers preserved to allow hierarchical modeling of within- and between-run variance. This design minimizes the influence of any single idiosyncratic trajectory on the model results.

\end{itemize}

\subsection{Model stability and accessibility}
\begin{itemize}
  \item \textbf{Model and inference engine:} All simulations used the quantized model version \texttt{matatonic/Mistral-Small-24B-Instruct-2501-6.5bpw-h8-exl2}, accessed on May 1st, 2026. The model weights were downloaded from HuggingFace\footnote{\url{https://huggingface.com/matatonic/Mistral-Small-24B-Instruct-2501-6.5bpw-h8-exl2}} and used unchanged across all inference runs. Inference was executed with the \texttt{ExLlamaV2} engine, consistent with the setup described in the Methods of the main text.
  
  \item \textbf{Model accessibility:} Both the model weights and inference engine are publicly available, enabling full replication of the simulation pipeline.  
  
  \item \textbf{Model stability:} To ensure version stability, the downloaded weights were stored locally and used identically throughout all runs. The version of the inference engine used (\texttt{ExLlamaV2 v0.2.3}) is specified in the installation script and environment files to enable future replications to reproduce an identical inference setup. To further reduce randomness in generation, we ran all simulations on the same hardware (RTX 4090 GPUs).
\end{itemize}

\subsection{Validation and justification}
\begin{itemize}
  \item \textbf{LLM output validation:} Because the present work involves open-ended, sequential simulations rather than isolated one-shot completions, validation cannot be performed at the level of individual prompt–response pairs. Instead, validation operates at the analytic level—whether aggregate behavioral patterns across runs reproduce theory-consistent threat effects. For the linguistic classifiers used in analysis, we report the model architecture, fine-tuning dataset, and validation performance. Specifically, moral-language classification employed a \texttt{roberta-base} model fine-tuned on the Moral Foundations Twitter Corpus \citep[MFTC;][]{hoover2020moral}, achieving a cross-validated F1 score of~0.76. Sentiment and hate-speech detection relied on two large \texttt{DeBERTaV3} models \citep{he2021debertav3}, fine-tuned on the \texttt{tweet\_eval} dataset \citep{barbieri2020tweeteval}, which achieved validation F1 scores of~0.74 (sentiment) and~0.61 (hate). DeBERTa improves upon BERT and RoBERTa by introducing disentangled attention and enhanced mask decoding, yielding superior performance on most natural-language-understanding benchmarks.

  \item \textbf{Validation of experimental manipulations:} Because the primary outcomes emerge through extended agent behavior rather than single-response accuracy, validation focuses on verifying that manipulations produced the intended internal states. To confirm that the threat manipulations were effective and stable, we modeled agents’ probed perceptions over time (see SI Section~5.1). Mixed-effects analyses showed that agents in symbolic- and realistic-threat conditions reported persistently elevated perceptions of the targeted threat type (near ceiling on a 1–7 scale) and baseline levels of the non-target type, with negligible temporal drift. These results confirm that the reinforcement–suppression procedure maintained stable, condition-appropriate threat representations throughout the simulation period.

  \item \textbf{Robustness of prompts and model settings:} The simulation framework employs dynamic prompt templates (all provided in the public repository) governing agent reasoning, planning, dialogue generation, and action execution. Templates were designed to enforce consistent output structure and valid action formats (e.g., proper location names or durations) while allowing open-ended reasoning. Prompt and inference configurations were optimized for stability in long-horizon trajectories rather than isolated completions, balancing computational efficiency, reproducibility, and behavioral realism. Manipulation-related prompts (e.g., group membership, threat perception, moral value orientation) were theory-driven and tightly specified to preserve construct validity; therefore, no prompt or parameter variation was conducted, as doing so may change the conceptual meaning of the experimental manipulation. Future work could explore prompt-level robustness among variations that are still conceptually aligned, but such analyses are beyond the present study’s scope. Instead, as specified above, we validated the experimental manipulation by probing the agents as described above.

  \item \textbf{Validation level and robustness to nondeterminism:} Robustness was assessed through independent repetitions of each experimental condition, conducted with distinct random seeds and randomized agent assignments. Mixed-effects models were fitted across all runs, treating inter-run variability as a random source of noise—analogous to variability across independent replications in behavioral experiments—such that systematic effects would cancel out if driven purely by stochasticity. The persistence of strong and theory-consistent effects across the aggregated data therefore indicates that the observed patterns are robust to nondeterminism in LLM generation and initialization.

  \item \textbf{Interpretation limits and lack of human ground truth:} Direct ground-truth validation against human data is not feasible for this paradigm for two reasons. First, comparably fine-grained, longitudinal datasets capturing real-world intergroup dynamics at the level of individual behaviors and interactions do not exist, making one-to-one correspondence unattainable. Second, even available causal evidence from laboratory or survey studies typically relies on self-reported attitudes or hypothetical judgments under low-stakes conditions, which differ fundamentally from the repeated, consequence-bearing interactions modeled here. The simulations therefore serve as a complementary framework for testing causal hypotheses about threat and social dynamics in a naturalistic yet fully controllable setting. To assess ecological plausibility, we also examined an independent hiring-bias scenario (SI Section~5.1) within the exact same experimental paradigm to test whether other related biases and discrimination that we can compare to human data emerge. Agents reproduced well-established human social-judgment patterns—physical attractiveness increased hiring success ($\hat\beta = 0.35$, $p < .001$) while a foreign accent decreased it ($\hat\beta = -0.26$, $p = .001$)—closely matching meta-analytic human effect sizes \citep{hosoda2003effects, spence2024your} (preference for attractive candidates $d = .37$; preference for foreign accents $d = -.47$). These findings demonstrate that the same architecture generating intergroup threat dynamics also captures realistic, socially patterned decision biases. Nonetheless, all results should be interpreted as model-based generative approximations of social behavior rather than direct measurements of human populations.
\end{itemize}

\subsection{Data processing and error handling}
\begin{itemize}
  \item \textbf{Pipeline transparency:} The full data-processing pipeline—from raw simulation logs to analytic datasets—is implemented in open scripts included in the public repository. These scripts document every transformation step, including log parsing, variable extraction, aggregation, and data merging. No manual data editing was performed. 

  \item \textbf{Error handling and exclusions:} All logged agent actions, plans, and conversations were retained for analysis. The simulation framework was extensively tested prior to data collection to prevent invalid or malformed outputs (e.g., missing actions, non-executable plans, or incomplete dialogues) that could otherwise interrupt simulation continuity. Because the environment requires valid outputs for progression, such events are functionally precluded during runtime. Consequently, no data exclusions or post hoc error corrections were necessary.

  \item \textbf{Bias analysis:} While no data were excluded, pipeline diagnostics confirmed that logged data volume and composition were consistent across experimental conditions, reducing the likelihood of systematic bias related to simulation integrity or runtime variability.
\end{itemize}

\end{appendices}
\end{document}